%% file: main.tex
\documentclass{article} % For LaTeX2e
\usepackage{iclr2027_conference,times}
\iclrfinalcopy

\input{math_commands.tex}

\usepackage[nopatch=eqnum]{microtype}
\usepackage{url}
\usepackage{booktabs}
\usepackage{siunitx}
\usepackage{float}
\usepackage{pgfplots}
\usepgfplotslibrary{groupplots}
\pgfplotsset{compat=1.18}
\usepackage{tikz}
\usepackage{twemojis}
\usepackage{hyperref}
\usepackage{fontawesome5}
\usetikzlibrary{arrows.meta,calc,positioning}
\input{pooldino_style.tex}

\title{Pooling Representation Autoencoders for Efficient Diffusion}

\author{Ramón Calvo-González \\
University of Geneva \\
\texttt{ramon.calvogonzalez@unige.ch}
\And
Youssef Saied \\
University of Geneva \\
\texttt{youssef.saied@unige.ch}
\And
François Fleuret \\
University of Geneva and Meta
}

\begin{document}

\maketitle
\lhead{Preprint}

\begin{abstract}
Representation autoencoders (RAEs) generate images from pretrained visual features, but their dense token grids make generative modeling expensive. Motivated by local feature correlations, we introduce PoolDINO, a learned affine pooling operator that merges neighboring tokens. Training the pooling operator jointly with the RGB decoder preserves the standard two-stage RAE procedure without a separate feature autoencoder. On ImageNet-256, $4\times$ token compression retains comparable generation quality under internal guidance, while $16\times$ compression trades some quality for greater efficiency. At a fixed budget of 100 sampling steps, latent-sampling throughput increases by $3.7\times$ and $9.0\times$, respectively, relative to the unpooled baseline. Classification and dense prediction evaluations show that comparable guided generation quality can coexist with weaker performance on other tasks.
\end{abstract}

\begin{center}
\small
\href{https://github.com/noctrog/pooldino}{\faGithub\enspace Code}
\qquad
\href{https://pooldino.ramoncalvo.com}{\faGlobe\enspace Project page}
\qquad
% Official logo: https://huggingface.co/brand (hf-logo-monochrome.png).
\href{https://huggingface.co/noctrog/pooldino}{\raisebox{-0.18em}{\includegraphics[height=1.15em]{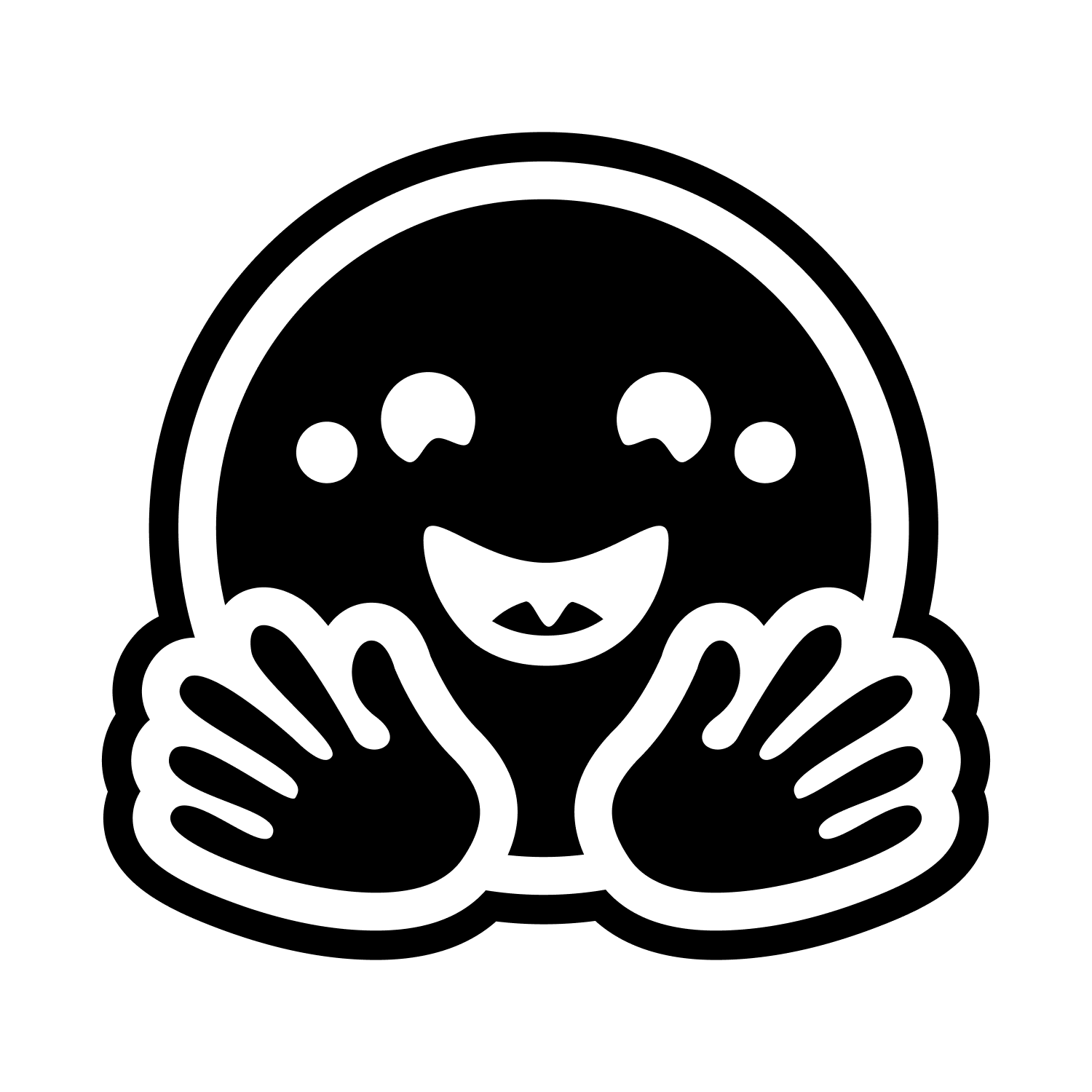}}\enspace Hugging Face}
\end{center}

\input{figures/quality_throughput.tex}

\section{Introduction}
% Keep the opening paragraph together across page breaks.
\begin{samepage}
% INTRO CORE POINT 1 -- Why SSL features matter for generation: pretrained
% visual structure improves training and sample quality; RAEs model these
% features directly as generative latents.
Pretrained self-supervised visual features have become a powerful foundation for image generation. REPA shows that using these features as alignment targets substantially accelerates diffusion training and improves sample quality~\citep{ICLR2025_d9e42b4d}. Representation autoencoders (RAEs) use them directly as the generative latent space within a two-stage framework: first training an image decoder on frozen vision-encoder features, and then training a flow-matching model to generate these features~\citep{ICLR2026_3c4141c1}. Although earlier work identified optimization difficulties in high-dimensional latent spaces~\citep{Yao_2025_CVPR}, RAEs successfully enable efficient generation directly from these features, with RAEv2 yielding further gains~\citep{singh2026improvedbaselinesrepresentationautoencoders}.
\par
\end{samepage}

% INTRO CORE POINT 2 -- Plausibility first, payoff second: high feature dimension
% and local correlation motivate compression, with iREPA as supporting context.
% If generation quality is retained, fewer tokens reduce generative compute.

However, the spatial resolution of these representations translates into long token sequences that are computationally expensive to model with Transformers~\citep{NIPS2017_3f5ee243}. The iREPA analysis demonstrates that nearby tokens in these feature maps are strongly correlated~\citep{ICLR2026_3929a778}. Because these tokens are also high-dimensional, we hypothesize that a single pooled token possesses enough capacity to summarize its local neighborhood, thereby reducing both the token count and the associated memory and compute requirements of the generative Transformer.

% INTRO CORE POINT 3 -- Method and rationale: learn pooling jointly with the RGB
% decoder so reconstruction determines what to retain; then train the generator
% on compressed latents, preserving the frozen encoder and two-stage procedure.
Building on RAEv2, we introduce PoolDINO, a learned affine operator that pools neighboring encoder patches into a single token. Pooling reduces the sequence length while retaining an explicit coarse two-dimensional grid. We train the pooling operator jointly with the RGB decoder while keeping the vision encoder frozen. The decoder's perceptual and adversarial objectives favor perceptually faithful reconstruction rather than exact recovery of every image detail~\citep{ICLR2026_3c4141c1}. Learning a pooling operator under these objectives allows reconstruction requirements to guide compression. We then freeze the pooling operator and decoder and train the generator on the compressed representations. By learning compression jointly with RGB reconstruction, PoolDINO preserves the two-stage RAEv2 training pipeline.

% INTRO CORE POINT 4 -- Results and limits: 4x compression retains comparable
% guided quality with faster sampling; 16x trades quality for more speed.
% Classification weakens, and RGB-trained pooling does not consistently beat
% average pooling on segmentation or depth estimation.
By varying the compression rate, we establish a controlled trade-off between generation quality and sampling throughput (Figure~\ref{fig:quality-throughput}). At $4\times$ compression, we retain generation quality comparable to the uncompressed reference, while more aggressive scales trade fidelity for significant efficiency gains. We also examine the downstream transferability of the pooled representations through operator analysis, image classification, semantic segmentation, and monocular depth estimation. Classification accuracy decreases despite comparable guided generation quality at $4\times$, and pooling learned for RGB reconstruction does not consistently outperform average pooling on the dense prediction tasks.

% INTRO CORE POINT 5 -- Contributions: the pooling method, the generation-speed
% trade-off, and retained-information analysis; end each point with section refs.
Our contributions are:
\begin{itemize}
    \item \textbf{Method:} We introduce PoolDINO, a local affine pooling operator trained jointly with an RGB decoder to compress frozen vision-encoder representations (Section~\ref{sec:method}).
    \item \textbf{Generation:} We characterize the trade-off between token compression, guided generation quality, and sampling throughput on ImageNet-256 (Section~\ref{sec:generation}; Appendix~\ref{app:throughput}).
    \item \textbf{Analysis:} We assess the semantic content using operator analysis, classification, semantic segmentation, and monocular depth estimation (Section~\ref{sec:semantics}; Appendix~\ref{app:operator-analysis}).
\end{itemize}

\section{Related work}

\textbf{Pretrained representations for generation.} REPA aligns a diffusion model's hidden states with pretrained visual features to accelerate training and improve generation quality~\citep{ICLR2025_d9e42b4d}. iREPA identifies spatial self-similarity in these features as a stronger predictor of alignment effectiveness than classification accuracy~\citep{ICLR2026_3929a778}. Pretrained features can also be used to guide the latent space itself: VA-VAE aligns variational autoencoder (VAE; \citealt{kingma2014autoencoding}) latents with vision-encoder features during tokenizer training~\citep{Yao_2025_CVPR}, while REPA-E uses representation alignment to jointly tune the VAE and diffusion model~\citep{Leng_2025_ICCV}.

RAEs use frozen vision-encoder features directly as generative latents, training an image decoder to reconstruct RGB images from them~\citep{ICLR2026_3c4141c1}. RAEv2 combines features from multiple encoder layers to improve reconstruction and introduces further refinements to generator training~\citep{singh2026improvedbaselinesrepresentationautoencoders}.

\newpage
\textbf{Compressing self-supervised features.} FAE reduces the channel dimension of RAE latents using a single attention layer and a linear projection~\citep{gaoOneLayerEnough2025}. However, this requires an additional training stage to learn the feature autoencoder. Moreover, because FAE strictly maintains the original spatial token count, it fails to alleviate the computational bottleneck in the generation stage.

FlatDINO addresses spatial redundancy with a Transformer autoencoder that compresses dense DINOv2 features into a shorter token sequence~\citep{calvogonzalez2026laminatingrepresentationautoencodersefficient}. Like FAE, it introduces an extra training stage for feature compression. Furthermore, despite its complex architecture, most of its latent tokens learn to compress fixed spatial chunks independent of the image content. Under the evaluated sampling configurations, PoolDINO achieves a stronger quality--throughput trade-off than FlatDINO (Figure~\ref{fig:quality-throughput-50steps}).

Motivated by strong correlations between neighboring features, PoolDINO employs explicit spatial pooling, compressing each local window with a simple affine map trained jointly with the RGB decoder. This directly optimizes compression for image reconstruction, avoiding the separate feature autoencoders required by FAE and FlatDINO while preserving the two-stage RAE training pipeline.

\section{Preliminaries}

\subsection{Flow matching}

Given samples from an unknown data distribution $p_{\mathrm{data}}$, the goal of generative modeling is to produce new samples that follow the same distribution. Flow matching approaches this task by learning a time-dependent velocity field that transforms samples from a simple distribution, typically Gaussian noise, into data~\citep{lipman2023flow}.

To construct training examples, we independently sample $x\sim p_{\mathrm{data}}$ and $\epsilon\sim\mathcal N(0,I)$, and interpolate between them:
\begin{align}
    x_t = (1-t)\epsilon+tx, \qquad t\in[0,1].
\end{align}
Here, $t=0$ corresponds to noise and $t=1$ to clean data. For each sampled pair, this path has constant velocity $x-\epsilon$.

A model $v_\theta(x_t,t)$ learns to predict this velocity by minimizing
\begin{align}
    \mathcal L_{\mathrm{FM}}
    =
    \mathbb E_{x,\epsilon,t}
    \left[
        \left\|v_\theta(x_t,t)-(x-\epsilon)\right\|_2^2
    \right].
\end{align}
Although each training target depends on a particular data--noise pair, the optimal predictor averages the velocities compatible with the observed $x_t$. This yields a velocity field that transports the corresponding distributions, without requiring numerical integration during training.

To generate a sample, we initialize $x_0\sim\mathcal N(0,I)$ and numerically integrate $\mathrm d x_t/\mathrm dt=v_\theta(x_t,t)$ from $t=0$ to $t=1$. The same formulation applies to latent representations in place of images. The model can also predict the clean sample $f_\theta(x_t,t)$, with the velocity obtained analytically as $v_\theta(x_t,t)=(f_\theta(x_t,t)-x_t)/(1-t)$ for $t<1$.

\subsection{Representation autoencoders}

The RAE framework uses a pretrained vision encoder $E$ to map an image $x \in \mathbb{R}^{H\times W\times 3}$ to a latent representation $z \in \mathbb{R}^{h\times w\times D}$, where $h, w$ define the spatial grid size and $D$ is the token dimensionality. An image decoder $D_\theta$ is then trained to recover $x$ directly from $z$.

While the original RAE extracts $z$ from the final block $K$ of the encoder, $z=E_K(x)$, RAEv2 defines the latent space as an average over a set of intermediate layers $\mathcal{L}$ to balance the fine-grained spatial structure of earlier layers with the high-level semantic context of deeper layers:
\begin{align*}
    z &= \frac{1}{|\mathcal{L}|} \sum_{\ell \in \mathcal{L}} \operatorname{LN}(E_\ell(x)),
\end{align*}
where $E_\ell(x)$ denotes the activations after block $\ell$, and $\operatorname{LN}$ is a parameter-free layer normalization. Following RAEv2, we use a DINOv3-L/16 encoder~\citep{simeoni2026dinov3} with $\mathcal L=\{11,13,15,17,19,21,23\}$.

\section{Method}
\label{sec:method}

\subsection{Image decoding}

% can we quantify if information is lost?
% can we quantify which information is used by the decoder?
% why is generating RAE latents easier than pixel space if the dimensionalities are the same?

Two properties of continuous semantic latents make them well suited to spatial compression. First, nearby vision-encoder features exhibit strong local correlation~\citep{ICLR2026_3929a778}. Second, pooling preserves the high-dimensional channel capacity $D$ of the original tokens, providing each pooled token with sufficient bandwidth to summarize the features of several constituent patches. Because the number of scalar elements in the uncompressed latent field $z$ can be comparable to that in the raw image $x$, we hypothesize that a learned spatial pooling operator can heavily compress this token grid while preserving the essential semantic structure required for high-quality image decoding.

Given the RAEv2 latent $z \in \mathbb{R}^{h \times w \times D}$, we partition it into non-overlapping windows of size $p_y \times p_x$ and map each window to a single token:
\begin{align}
\hat{z}_{i,j}
=
\mathcal{P}\!\left(
z[i p_y:(i+1)p_y,\; j p_x:(j+1)p_x,\; :]
\right).
\end{align}
Here, the colon denotes array slicing, and $\mathcal{P}: \mathbb{R}^{p_y \times p_x \times D} \rightarrow \mathbb{R}^{D}$ is applied independently to each window.
Assuming that $h$ and $w$ are divisible by $p_y$ and $p_x$, respectively, this gives the pooled latent $\hat{z} \in \mathbb{R}^{\frac{h}{p_y} \times \frac{w}{p_x} \times D}$.

We use learned affine pooling: $\mathcal{P}$ is a shared affine map with weight $W \in \mathbb{R}^{D \times mD}$ and bias $b\in\mathbb R^D$, where $m=p_xp_y$, applied to the concatenated features of each window. This is equivalent to a convolution with kernel size and stride both equal to $(p_y,p_x)$. Before RGB decoding, we repeat each pooled token over its original window. We compare this learned pooling against fixed average and max pooling.

A ViT-XL decoder then learns to reconstruct $x$ from the repeated pooled field. Repetition restores the original grid, keeping the decoder architecture and input sequence length fixed across pooling geometries to control for decoder compute.

\begin{figure}[t]
\centering
\resizebox{0.94\textwidth}{!}{\input{figures/pooled_decoder_training.tex}}
\caption{Training the pooled image decoder. A frozen DINOv3-L/16 produces a dense patch field. The learned projection $\mathcal P$ maps each non-overlapping pooling window to one token, which is repeated over its original window before ViT-XL reconstructs the image. Shades distinguish source patches; uniform colors within the repeated blocks denote identical token copies. A $4\times4$ field with $2\times2$ pooling windows is shown for clarity.}
\label{fig:pooled-decoder-training}
\end{figure}

\subsection{Image generation}

With the tokenizer and RGB decoder frozen, we train one class-conditional diffusion Transformer (DiT$^{\mathrm{DH}}$-XL; \citealt{Peebles_2023_ICCV,singh2026improvedbaselinesrepresentationautoencoders}) for each pooling geometry. The generator operates directly on the corresponding pooled latent grid,
\begin{align}
    z = E(x), \qquad
    \hat z = \mathcal P(z), \qquad
    \bar z = \frac{\hat z-\mu_{\hat z}}{\sigma_{\hat z}},
\end{align}
where $\mu_{\hat z}$ and $\sigma_{\hat z}$ are channel-wise statistics. We use the RAEv2 architecture and transport objective, adapting the spatial input grid to each pooling geometry. Unless otherwise specified, each model is trained for 80 epochs with a global batch size of $1{,}024$, using a hybrid Muon--AdamW optimizer~\citep{jordan2024muon,loshchilov2019decoupled} and a base learning rate of $2\times10^{-4}$. The learning rate is held constant through epoch 25 and decayed linearly to $2\times10^{-5}$ by epoch 50. We drop the class condition with probability $0.1$ to enable classifier-free guidance. Full optimization details are given in Table~\ref{tab:opt-hparams}. We also use the internal guidance (IG; \citealt{zhou2026guiding}) mechanism, which trains an intermediate generator layer to predict the clean latent. During sampling, extrapolating from this weaker intermediate prediction to the full-depth prediction provides a strong guidance signal without requiring a separate unconditional evaluation.

The training objective contains the usual flow-matching loss and two auxiliary losses applied to the hidden states after the eighth DiT block:
\begin{align}
    \mathcal L
    =
    \mathcal L_{\mathrm{FM}}
    +
    \mathcal L_{\mathrm{IG}}
    +
    0.5\,\mathcal L_{\mathrm{rec}}^{E}.
\end{align}
Here $\mathcal L_{\mathrm{FM}}$ is the flow-matching loss of the full generator (using an $x$-prediction objective; \citealt{Li_2026_CVPR}), and $\mathcal L_{\mathrm{IG}}$ applies the same objective to the intermediate IG prediction. A second auxiliary head reconstructs the original dense encoder patch field $z=E(x)$ using mean-squared error ($\mathcal L_{\mathrm{rec}}^{E}$). Both auxiliary heads are trained jointly for every model. Appendix~\ref{app:training-details} gives the full training diagram.

During sampling, let $\bar z_{\mathrm{full}}$ denote the full-depth prediction and $\bar z_{\mathrm{int}}$ the intermediate prediction. IG extrapolates between them:
\begin{align}
    \bar z_{\mathrm{IG}}
    =
    \bar z_{\mathrm{int}}
    +
    s_{\mathrm{IG}}
    \left(
        \bar z_{\mathrm{full}}-\bar z_{\mathrm{int}}
    \right),
\end{align}
where $s_{\mathrm{IG}}=1$ recovers the original prediction. We evaluate IG both alone (used in Figure~\ref{fig:quality-throughput}) and combined with standard classifier-free guidance (CFG; \citealt{ho2021neuripsw-classifierfree}), whose scale is $s_{\mathrm{CFG}}$. Scales are selected separately for each model. Appendix~\ref{app:guidance-ablations} gives full ablation results, including a comparison between IG and guidance derived from the encoder-reconstruction head.

After sampling, we undo the latent normalization, repeat each pooled token over its original spatial window, and apply the frozen RGB decoder:
\begin{align}
    x_{\mathrm{gen}}
    =
    D_\theta\!\left(
        \operatorname{repeat}
        \left(
            \sigma_{\hat z}\bar z_{\mathrm{sample}}+\mu_{\hat z}
        \right)
    \right).
\end{align}
Repetition restores the decoder's $16\times16$ input grid but introduces no additional information; we do this to strictly control for decoder FLOPs across all pooling geometries. Appendix~\ref{app:training-details} lists the latent grids, architecture, and optimization settings.

\section{Results}

We evaluate reconstruction and class-conditional generation on ImageNet-1K~\citep{ILSVRC15} at $256\times256$ resolution.

\subsection{Reconstruction}
\label{sec:reconstruction}

Learning the pooling operator preserves reconstruction quality more effectively than parameter-free pooling (Table~\ref{tab:pooled-decoder-rfid}). At $4\times$ compression, learned pooling reaches a reconstruction Fr\'echet Inception Distance (rFID; \citealt{NIPS2017_8a1d6947}) of 0.36, close to the unpooled reference at 0.32 and lower than the average pooling control at 0.65. At $16\times$, its rFID rises to 0.41, compared with 1.81 for average pooling and 1.72 for max pooling. PSNR and LPIPS also favor learned pooling over both parameter-free alternatives.

\begin{table}[!htbp]
\centering
\caption{
  Reconstruction on the ImageNet-1K $256\times256$ validation set, grouped by pooling method. All RGB decoders process 256 tokens after repetition.
}
\label{tab:pooled-decoder-rfid}
\small
\setlength{\tabcolsep}{3.5pt}
\begin{tabular}{lrrrrr}
  \toprule
  Spatial pooling & Tokens & rFID $\downarrow$ & sFID $\downarrow$ & PSNR $\uparrow$ & LPIPS $\downarrow$ \\
  \midrule
  RAEv2 baseline          & 256 & 0.32 & 2.29 & 22.74 & 0.15 \\
  \midrule
  Learned $2{\times}2$ &  64 & 0.36 & 2.61 & 22.25 & 0.17 \\
  Learned $2{\times}4$ &  32 & 0.39 & 2.76 & 21.88 & 0.18 \\
  Learned $4{\times}2$ &  32 & 0.39 & 2.82 & 21.85 & 0.18 \\
  Learned $4{\times}4$ &  16 & 0.41 & 2.99 & 21.45 & 0.19 \\
  \midrule
  Average $2{\times}2$ &  64 & 0.65 & 4.97 & 19.58 & 0.25 \\
  Average $4{\times}4$ &  16 & 1.81 & 8.65 & 16.64 & 0.37 \\
  \midrule
  Max $2{\times}2$     &  64 & 0.71 & 5.17 & 19.17 & 0.26 \\
  Max $4{\times}4$     &  16 & 1.72 & 8.31 & 16.22 & 0.38 \\
  \bottomrule
\end{tabular}
\end{table}

\newpage
\subsection{Image generation}
\label{sec:generation}

\input{figures/curated_compression_samples.tex}

\begin{table}[!htbp]
\caption{ImageNet-256 generation at 100 Euler steps. Models train for 80 epochs. We report generation FID and Inception Score (IS; \citealt{salimans2016improved}). Guided columns report the lowest-FID settings from our guidance scale sweeps; the corresponding scales are in Table~\ref{tab:selected-guidance-scales}. Best and second-best metrics in each column are bold and underlined.}
\label{tab:main-generation-results}
\centering
\small
\setlength{\tabcolsep}{4.5pt}
\begin{tabular}{@{}lrrr|rr|rr@{}}
\toprule
& & \multicolumn{2}{c|}{Unguided} & \multicolumn{2}{c|}{IG} & \multicolumn{2}{c}{IG + CFG} \\
\cmidrule(lr){3-4}\cmidrule(lr){5-6}\cmidrule(lr){7-8}
Spatial pooling & Tokens & FID $\downarrow$ & IS $\uparrow$ & FID $\downarrow$ & IS $\uparrow$ & FID $\downarrow$ & IS $\uparrow$ \\
\midrule
RAEv2 baseline & 256 & \textbf{1.53} & \underline{226.63} & \textbf{1.08} & \underline{262.00} & \underline{1.07} & 267.15 \\
\midrule
Learned $2{\times}2$ & 64 & 3.00 & 184.49 & \underline{1.09} & 250.36 & \textbf{1.07} & 268.50 \\
Learned $2{\times}4$ & 32 & 4.91 & 161.42 & 1.19 & 249.58 & 1.16 & 273.06 \\
Learned $4{\times}2$ & 32 & 4.97 & 159.67 & 1.21 & 249.96 & 1.19 & \underline{273.98} \\
Learned $4{\times}4$ & 16 & 8.16 & 132.67 & 1.44 & 247.07 & 1.35 & \textbf{282.29} \\
\midrule
Average $2{\times}2$ & 64 & \underline{1.70} & \textbf{232.55} & 1.29 & \textbf{268.08} & -- & -- \\
\bottomrule
\end{tabular}
\end{table}

Table~\ref{tab:main-generation-results} shows that $4\times$ compression retains comparable image generation quality with IG alone (1.09 FID versus the uncompressed baseline's 1.08). While $16\times$ compression trades some fidelity for greater efficiency by reaching 1.44 FID, adding CFG further improves the lowest observed FID across all models. This combined guidance yields the largest improvements at $16\times$ compression, though it requires an additional unconditional forward pass per sampling step. Appendix~\ref{app:guidance-ablations} gives the full comparison with CFG alone and encoder-reconstruction guidance, including the selected scales and complete sweeps.

\paragraph{Average pooling baseline.}
We compare the learned affine mapping against a simpler pooling strategy by training a $2\times2$ average-pooled generator for 80 epochs. While it achieves stronger unguided generation (1.70 FID) than its learned counterpart (3.00 FID), it responds less effectively to IG (1.29 FID) and fails to match the uncompressed baseline.

\paragraph{Extended training.}
The lower per-update cost of compressed generators allows longer training within the compute budget of the unpooled baseline. We train the $2\times2$ and $4\times4$ generators for 180 and 300 epochs, respectively (Appendix~\ref{app:training-budget}). Table~\ref{tab:extended-training-generation} shows that extended training improves generation quality at both compression rates. Figure~\ref{fig:curated-training} compares samples from the 80-epoch and extended-training checkpoints, which are also included in Figure~\ref{fig:quality-throughput}. Further training to approximately match the unpooled baseline's compute budget does not improve the best observed FID (Appendix~\ref{app:training-duration}).

\paragraph{Fewer sampling steps.}
Reducing sampling from 100 to 50 steps maintains comparable generation quality while providing a further twofold increase in latent-sampling throughput. Appendix~\ref{app:sampling-steps} provides the full comparison.

\begin{table}[!htbp]
\caption{Effect of extended training, evaluated with 100 Euler steps at the guidance settings in Table~\ref{tab:selected-guidance-scales}. The unpooled 80-epoch model is included as a reference; bold marks the stronger result within each pooling window. Appendix~\ref{app:training-duration} gives longer runs and finer IG sweeps.}
\label{tab:extended-training-generation}
\centering
\small
\setlength{\tabcolsep}{5pt}
\begin{tabular}{@{}lr|rr|rr@{}}
\toprule
& & \multicolumn{2}{c|}{Unguided} & \multicolumn{2}{c}{IG} \\
\cmidrule(lr){3-4}\cmidrule(lr){5-6}
Pooling & Epochs & FID $\downarrow$ & IS $\uparrow$ & FID $\downarrow$ & IS $\uparrow$ \\
\midrule
$1\times1$ (unpooled) & 80 & 1.53 & 226.63 & 1.08 & 262.00 \\
\midrule
$2\times2$ & 80 & 3.00 & 184.49 & 1.09 & 250.36 \\
$2\times2$ & 180 & \textbf{2.78} & \textbf{192.37} & \textbf{1.05} & \textbf{272.00} \\
\midrule
$4\times4$ & 80 & 8.16 & 132.67 & 1.44 & 247.07 \\
$4\times4$ & 300 & \textbf{7.19} & \textbf{146.13} & \textbf{1.29} & \textbf{258.04} \\
\bottomrule
\end{tabular}
\end{table}

\input{figures/curated_training_samples.tex}

Figure~\ref{fig:curated-compression} shows generated samples across compression levels. Appendix~\ref{app:generated-samples} provides randomly selected samples from the unpooled reference, the compressed models, and both extended-training checkpoints.

\subsection{Classification and dense prediction}
\label{sec:semantics}
\label{sec:dense-prediction}

Beyond reconstruction and generation, we ask what information remains accessible in the compressed representations. We evaluate both global semantic information through image classification and localized information through semantic segmentation and monocular depth estimation.

\textbf{Image classification and operator analysis.} We evaluate global semantic content on ImageNet classification by spatially averaging the pooled field $\hat z$ into one feature vector per image. Features from learned pooling yield lower accuracy than all three simpler baselines at every compression rate on both linear probing and k-NN evaluations (Table~\ref{tab:dense-transfer}; Appendix~\ref{app:classification}). To understand this difference, we examine the operator's subspace: the learned operator retains a different subspace from average pooling and local PCA. Its alignment with both decreases as compression increases, and it preserves less total variance than PCA at the same output dimension (Appendix~\ref{app:operator-analysis}). These measurements characterize the learned representation, but do not establish which preserved directions account for the generation results.

\textbf{Dense prediction.} Because spatial pooling reduces resolution, we test whether the compressed token grid still preserves the localized semantic and geometric details required for pixel-level tasks. We evaluate this on ADE20K~\citep{Zhou_2017_CVPR} semantic segmentation and NYUv2~\citep{Silberman_2012} monocular depth estimation by freezing the encoder and pooling operator and training a randomly initialized ViT-XL task model for each representation (Table~\ref{tab:dense-transfer}). We report mean intersection over union (mIoU) for segmentation and absolute relative error (AbsRel) for depth. At $4\times$ compression, average pooling performs better on both tasks. At $16\times$, learned pooling gives higher segmentation mIoU, while average pooling retains lower depth error. Thus, the comparable guided generation quality achieved at $4\times$ coexists with weaker transfer to dense prediction. These results do not establish a consistent transfer advantage for pooling learned through RGB reconstruction. Future work should test whether learning $\mathcal P$ separately for each downstream task yields more suitable representations. Appendix~\ref{app:dense-prediction} gives the complete metrics and results with RGB-decoder initialization.

\begin{table}[!htbp]
\caption{Classification and dense prediction from frozen representations. Dense task models train from scratch. Bold marks the stronger pooling rule at each compression rate.}
\label{tab:dense-transfer}
\centering
\small
\setlength{\tabcolsep}{6pt}
\begin{tabular}{@{}lrrrr@{}}
\toprule
& & ImageNet-1K & ADE20K & NYUv2 \\
\cmidrule(lr){3-3}\cmidrule(lr){4-4}\cmidrule(lr){5-5}
Spatial pooling & Window & Linear top-1 (\%) $\uparrow$ & mIoU (\%) $\uparrow$ & AbsRel $\downarrow$ \\
\midrule
Unpooled & $1\times1$ & 85.31 & 46.86 & 0.0817 \\
\midrule
Learned & $2\times2$ & 83.47 & 39.38 & 0.0895 \\
Average & $2\times2$ & \textbf{85.33} & \textbf{42.59} & \textbf{0.0869} \\
\midrule
Learned & $4\times4$ & 80.80 & \textbf{38.61} & 0.1022 \\
Average & $4\times4$ & \textbf{85.33} & 36.76 & \textbf{0.0997} \\
\bottomrule
\end{tabular}
\end{table}

\section{Conclusion}

We introduced PoolDINO, a learned spatial pooling operator that compresses representation autoencoder latents for efficient generative modeling. By training a local affine mapping jointly with an RGB decoder, PoolDINO avoids the complex feature autoencoders required by prior methods and fully preserves the standard two-stage RAE training pipeline. 

Our evaluations demonstrate a trade-off between computational cost and generation quality. At $4\times$ compression, PoolDINO substantially increases sampling throughput while retaining a guided FID highly competitive with the uncompressed baseline. More aggressive $16\times$ compression enables substantial efficiency gains, trading visual fidelity for even higher throughput.

Finally, we evaluated transfer to classification and dense prediction. Learned pooling yields lower k-NN and linear-probe accuracy than average pooling, local PCA, and random projection, and does not consistently outperform average pooling on dense prediction tasks. These results show that strong guided generation quality does not guarantee equally strong downstream performance. Future work should investigate pooling objectives tailored to downstream tasks or jointly optimized for generation and transfer.

\paragraph{Limitations.}
Our generation experiments are restricted to ImageNet-256 and a single autoencoder and generator architecture. Although we explore extended generator training, our pooling-and-decoder training budget remains fixed; whether longer first-stage training can improve performance at higher compression rates remains open. Finally, competitive generation quality relies heavily on guidance, as unguided performance noticeably degrades under compression.

\bibliography{references}
\bibliographystyle{iclr2027_conference}

\clearpage

\appendix

\section{Training and model details}
\label{app:training-details}

We retain the two-stage RAE procedure. The first stage learns the local affine pooling operator and RGB decoder with the vision encoder frozen; the second freezes these components and trains the class-conditional generator. Figure~\ref{fig:pooled-generator-training} shows the full generator training graph. Tables~\ref{tab:latent-settings}--\ref{tab:opt-hparams} give the latent grids and model settings, and Table~\ref{tab:flops-estimate} reports component-wise forward costs.

\paragraph{RGB reconstruction objective.}
Following RAEv2, we train the pooling operator and image decoder with
\begin{equation}
    \mathcal L_{\mathrm{RGB}}
    =\mathcal L_1+\mathcal L_{\mathrm{LPIPS}}
    +0.75\,a\,\mathcal L_{\mathrm{adv}},
\end{equation}
where $\mathcal L_1$ is mean absolute pixel error, $\mathcal L_{\mathrm{LPIPS}}$ is the perceptual reconstruction loss, and $\mathcal L_{\mathrm{adv}}$ is the negative mean discriminator score on reconstructed images. The adaptive weight $a$ is the ratio of the gradient norms of $\mathcal L_1+\mathcal L_{\mathrm{LPIPS}}$ and $\mathcal L_{\mathrm{adv}}$ at the decoder's final projection, with $10^{-6}$ added to the denominator and the ratio clipped to $[0,10^4]$. Pixel and perceptual losses are active from the start; discriminator training begins after six epochs and the decoder's adversarial term after eight. During training, we add Gaussian noise to the pooled tokens before repetition, with a standard deviation sampled independently for each image from $\mathcal U(0,0.8)$. Evaluation uses clean tokens and EMA parameters.

\begin{figure}[htbp]
\centering
\resizebox{0.98\textwidth}{!}{\input{figures/pooled_generator_training.tex}}
\caption{Training the pooled generator. The frozen encoder and pooling operator define the normalized clean latent $\bar z$. The generator predicts $\bar z_{\mathrm{full}}$ from $z_t=(1-t)\epsilon+t\bar z$, conditioned on time $t$ and class $y$. From block-8 hidden states $h^{(8)}$, the internal-guidance head $H_\theta$ predicts the same clean pooled target, while the encoder-reconstruction head $R_\theta$ predicts the dense features $z$.}
\label{fig:pooled-generator-training}
\end{figure}

As illustrated in Figure~\ref{fig:pooled-generator-training}, let $R_\theta$ denote the learned encoder-reconstruction head. At each pooled spatial position $(i,j)$, it maps the hidden state $h^{(8)}_{i,j}\in\mathbb R^{d_h}$ to
\begin{align}
    R_\theta:\mathbb R^{d_h}\rightarrow\mathbb R^{p_y p_x D},
    \qquad
    h^{(8)}_{i,j}\mapsto R_\theta\!\left(h^{(8)}_{i,j}\right).
\end{align}
This vector is reshaped into $p_y p_x$ separate $D$-dimensional predictions, one for every position in the original pooling window from which $\hat z_{i,j}$ was formed.

\begin{table}[htbp]
\caption{Latent grids. Compression is relative to the 256-token control.}
\label{tab:latent-settings}
\centering
\small
\begin{tabular}{lcccc}
\toprule
Pooling window & Latent grid & Tokens & Compression & $\kappa$ \\
\midrule
$1\times1$ (control) & $16\times16$ & 256 & $1\times$ & 8 \\
$2\times2$ & $8\times8$ & 64 & $4\times$ & 4 \\
$2\times4$ & $8\times4$ & 32 & $8\times$ & $\sqrt{8}$ \\
$4\times2$ & $4\times8$ & 32 & $8\times$ & $\sqrt{8}$ \\
$4\times4$ & $4\times4$ & 16 & $16\times$ & 2 \\
\bottomrule
\end{tabular}
\end{table}

\begin{table}[htbp]
\caption{Model architecture hyperparameters.}
\label{tab:arch-hparams}
\centering
\small
\begin{tabular}{lccc}
\toprule
Hyperparameter & Image decoder & Flow encoder & Flow decoder \\
\midrule
Transformer blocks & 28 & 28 & 2 \\
Hidden size & 1,152 & 1,440 & 2,048 \\
MLP width & 4,096 & 3,840 & 5,461 \\
Attention heads & 16 & 20 & 16 \\
Time tokens & -- & 4 & -- \\
Class tokens & -- & 8 & -- \\
\bottomrule
\end{tabular}
\end{table}

\begin{table}[htbp]
\caption{Optimization hyperparameters for both training stages.}
\label{tab:opt-hparams}
\centering
\small
\begin{tabular}{p{0.23\textwidth}p{0.27\textwidth}p{0.38\textwidth}}
\toprule
Hyperparameter & Decoder Stage & Flow-Model Stage \\
\midrule
Training epochs & 16 & 80 \\
Training steps & 40,032 & 100,080 \\
Global batch size & 512 & 1,024 \\
Optimizer & AdamW & Muon (AdamW fallback) \\
Base learning rate & $2\times10^{-4}$ & $2\times10^{-4}$ \\
Learning rate schedule & Cosine decay to $2\times10^{-5}$ & Constant through epoch 25; linear decay to $2\times10^{-5}$ at epoch 50 \\
Warm-up epochs & 1 & -- \\
Weight decay & 0 & 0 \\
Optimizer momentum & $(\beta_1=0.9, \beta_2=0.95)$ & Muon 0.95, AdamW $(0.9, 0.95)$ \\
Gradient clipping & None & 1.0 \\
EMA decay & 0.9978 & 0.9995 \\
\bottomrule
\end{tabular}
\end{table}

\begin{table}[htbp]
\caption{Estimated forward FLOPs per image, counting a multiply-add as two operations. Component columns report one evaluation. ``Internal head'' is the additional cost of returning the early clean-latent prediction. ``Representation path'' includes the dense-representation projection and pooling operation. Unguided generation counts 100 base generator evaluations and one image-decoder evaluation, excluding auxiliary heads and additional CFG evaluations.}
\label{tab:flops-estimate}
\centering
\small
\setlength{\tabcolsep}{3.2pt}
\begin{tabular}{lrrrrrr}
\toprule
Window & Tokens & Generator & Internal head & Repr. path & Decoder & \shortstack{Unguided\\generation} \\
\midrule
$1\times1$ & 256 & 473.182G & 2.887G & 0.756G & 224.216G & 47.542T \\
$2\times2$ & 64 & 128.653G & 0.722G & 1.292G & 224.216G & 13.090T \\
$2\times4$ & 32 & 72.531G & 0.361G & 1.292G & 224.216G & 7.477T \\
$4\times4$ & 16 & 44.612G & 0.180G & 1.292G & 224.216G & 4.685T \\
\bottomrule
\end{tabular}
\end{table}

\subsection{Extended-training compute budget}
\label{app:training-budget}

We compare the cost of extended generator training with the 80-epoch unpooled baseline. The per-update estimate includes frozen encoder and pooling operations, generator forward and backward passes, auxiliary losses, optimizer updates, and EMA updates. We count a multiply-add as two FLOPs and exclude first-stage decoder training and evaluation. With a global batch size of 1,024, each epoch contains 1,251 updates. The unpooled, $2\times2$, and $4\times4$ models require approximately 1,662.451, 599.874, and 340.648 TFLOPs per update, respectively. For reference, matching the baseline's total training cost would require
\begin{equation}
    e_{2\times2}^{\mathrm{match}}=80\,\frac{1662.451}{599.874}\simeq221.71,
    \qquad
    e_{4\times4}^{\mathrm{match}}=80\,\frac{1662.451}{340.648}\simeq390.42
\end{equation}
epochs. The $2\times2$ and $4\times4$ runs shown in the main text use 180 and 300 epochs, respectively. Their estimated costs are 135.08 and 127.85 EFLOPs, compared with 166.38 EFLOPs for the unpooled baseline: approximately 81\% and 77\% of its training budget. We additionally train to 222 and 390 epochs, respectively, using an estimated 166.60 and 166.20 EFLOPs (100.13\% and 99.89\% of the baseline budget). Appendix~\ref{app:training-duration} reports these approximately compute-matched endpoints.

\clearpage

\input{figures/training_duration.tex}

\clearpage

\section{Throughput measurement}
\label{app:throughput}

Figure~\ref{fig:quality-throughput} shows latent-sampling throughput for our models with IG alone. Figure~\ref{fig:quality-throughput-guidance} adds denoiser-step throughput and the CFG+IG results under the same measurement conventions.

\input{figures/quality_throughput_guidance.tex}

Figure~\ref{fig:quality-throughput-guidance} reports the throughput of individual guided denoiser steps (top) and complete latent-sampling trajectories (bottom). Our models follow the architecture detailed in Table~\ref{tab:arch-hparams}, varying only the spatial input grid. For all measurements, we execute model computation in BF16 and return full-precision model outputs.

Means and sample standard deviations are computed across 30 runs of 20 denoiser steps or one complete sampling trajectory, measured after compilation and a warmup phase. PoolDINO trajectory timings use CUDA events with device synchronization after at least five warmup trajectories and five seconds of warmup. Latent samples/s in the bottom panel counts completed latent samples, not end-to-end RGB outputs.

The throughput for our models is measured under two guidance settings. \textit{IG alone} uses a conditional model evaluation and its block-8 internal head. \textit{CFG+IG} adds an unconditional evaluation when CFG is active.

For the 100-step benchmarks, full trajectories use the same pooling-dependent time shifts as the quality evaluations: $8$, $4$, $\sqrt{8}$, and $2$ for $1\times1$, $2\times2$, $2\times4$, and $4\times4$, respectively. IG is active on $[0,0.9]$ and CFG on $(0.3,1)$ in noise-to-data time.

For context, Figure~\ref{fig:quality-throughput} includes the original guided XL implementations of RAE~\citep{ICLR2026_3c4141c1}, LightningDiT~\citep{Yao_2025_CVPR}, E2E-VAE + REPA from REPA-E~\citep{Leng_2025_ICCV}, MAETok~\citep{chen2025maskedautoencoderseffectivetokenizers}, SoftVQ-VAE~\citep{chen2025softvqvaeefficient1dimensionalcontinuous}, and REPA~\citep{ICLR2025_d9e42b4d}. MAR-H~\citep{NEURIPS2024_66e22646} appears only in the full-sampling panel, since its autoregressive iterations are not comparable denoiser steps. We use MAETok-B-128, SoftVQ-B-64, and SoftVQ-L-32 with SiT-XL, and the released 4M-iteration SiT-XL/2 for REPA. RAE employs a separate small autoguidance model; the other systems use their released CFG recipes. Literature FIDs are published results, not re-evaluations under benchmark precision, and training budgets, sampling recipes, and class-balance protocols differ. The REPA-E point is the 800-epoch E2E-VAE + REPA system, not the 80-epoch jointly trained variant.

RAE and LightningDiT use shifted Euler sampling with 49 and 249 denoiser evaluations, respectively. REPA-E, MAETok, SoftVQ-VAE, and REPA use 250-evaluation Euler--Maruyama sampling. MAR-H uses 256 autoregressive iterations, each with 100 DDPM steps; its internal Transformer decoder and diffusion-loss network are included. Full trajectories respect each method's guidance windows. These measured sampling rates are not obtained by dividing isolated step throughput by the number of evaluations.

The literature baselines use PyTorch/Inductor, while our models use JAX/XLA, so the comparison includes software-stack differences. All use cuDNN attention. The PoolDINO trajectory benchmark uses cuBLAS. 

Figure~\ref{fig:inception-throughput} gives the comparison using Inception Score, retaining the 100-step settings for our models.

\input{figures/inception_throughput.tex}

\input{figures/sampling_steps_results.tex}
\clearpage
\subsection{Batch-size and hardware sweeps}

Figures~\ref{fig:inference-speed} and~\ref{fig:training-speed} compare the same generator architecture across pooling grids on H100 NVL and A100 GPUs. The forward--backward benchmark measures the base generator and its main prediction loss; it excludes optimizer updates and auxiliary objectives.

\input{figures/speed.tex}

\input{figures/train_speed.tex}

\clearpage

\section{Image classification details}
\label{app:classification}

We freeze the encoder and pooling operator. For both ImageNet-1K splits, we apply deterministic resizing and center cropping to $256\times256$, followed by ImageNet normalization, without random augmentation. We spatially average each pooled field $\hat z$ into a 1,024-dimensional vector and normalize it to unit $\ell_2$ norm. The uncompressed reference uses the spatial mean of $z$. All pooling methods use the same encoder features and classifier input dimension.

\textbf{k-NN.} Following the weighted voting procedure of DINO~\citep{Caron_2021_ICCV}, we retrieve the $k$ training features with highest cosine similarity to each validation feature. Each neighbor votes for its class with weight $\exp(s/\tau)$, where $s$ is its cosine similarity and $\tau=0.07$. The class with the largest total weight is predicted. We report the best validation top-1 accuracy over $k\in\{10,20,100\}$.

\textbf{Linear probe.} We train a single affine classifier with cross-entropy on the cached training features, keeping the representation fixed. We use LARS for 90 epoch-equivalents with batch size 16,384, zero weight decay, and gradient-norm clipping at 10. Each update samples a batch of distinct training examples. The learning rate increases linearly from zero during the first 10\% of updates and then decays to zero with a cosine schedule. We sweep peak learning rates $\{1.6,3.2,6.4,12.8\}$ and report the highest validation top-1 accuracy among the final classifiers.

\textbf{Pooling baselines.} Table~\ref{tab:semantic-probes} compares learned pooling with average pooling, training-fitted local PCA, and random projection. For a window of $m$ patches, PCA centers the flattened $mD$-dimensional features with the training mean and projects onto the leading $D=1024$ components, without whitening. The components are fitted by randomized PCA on training blocks and held fixed for both classification splits. Random projections use the same training mean: we draw a standard Gaussian matrix in $\mathbb R^{mD\times D}$, orthonormalize its columns by QR decomposition, and use its transpose as the projection. Each projection is shared across all spatial windows. We report the mean and standard deviation across projection draws, selecting the best probe setting separately for each draw. These probes evaluate classification from spatially averaged features; the dense-task evaluations retain the token grid.

\begin{table}[t]
\caption{ImageNet-1K k-NN and linear-probe top-1 accuracy (\%). The uncompressed RAEv2 representation $z$ is included as a reference. Best and second-best methods at each compression rate are bold and underlined.}
\label{tab:semantic-probes}
\centering
\small
\setlength{\tabcolsep}{3.5pt}
\resizebox{\textwidth}{!}{%
\begin{tabular}{lrrrrrrrr}
\toprule
& \multicolumn{4}{c}{\textit{k-NN}} & \multicolumn{4}{c}{\textit{Linear probe}} \\
\cmidrule(lr){2-5}\cmidrule(lr){6-9}
Window & Learned & Average & PCA & Random & Learned & Average & PCA & Random \\
\midrule
Uncompressed $z$ & \multicolumn{4}{c}{76.03} & \multicolumn{4}{c}{85.31} \\
\midrule
$2\times2$ & 72.16 & \underline{76.00} & \textbf{76.67} & $75.78 \pm 0.06$ & 83.47 & \textbf{85.33} & 84.35 & $\underline{84.63 \pm 0.06}$ \\
$2\times4$ & 68.50 & \underline{76.00} & \textbf{76.58} & $75.77 \pm 0.03$ & 82.35 & \textbf{85.33} & 84.10 & $\underline{84.50 \pm 0.04}$ \\
$4\times2$ & 68.17 & \underline{76.00} & \textbf{76.59} & $75.68 \pm 0.05$ & 82.35 & \textbf{85.34} & 84.12 & $\underline{84.43 \pm 0.07}$ \\
$4\times4$ & 63.22 & \underline{76.00} & \textbf{76.39} & $75.68 \pm 0.04$ & 80.80 & \textbf{85.33} & 83.77 & $\underline{84.21 \pm 0.04}$ \\
\bottomrule
\end{tabular}
}
\end{table}

\clearpage

\section{Dense prediction details}
\label{app:dense-prediction}

We report complete validation metrics for task models trained from scratch and for models initialized from the corresponding RGB decoder. The tables select the checkpoint with the best primary validation metric. The scratch controls separate information retained by the representation from transfer provided by RGB-decoder initialization. For $C$ semantic classes and $N$ valid depth pixels, the primary metrics are
\begin{equation}
\operatorname{mIoU}=\frac{1}{C}\sum_{c=1}^{C}\frac{\mathrm{TP}_c}{\mathrm{TP}_c+\mathrm{FP}_c+\mathrm{FN}_c},
\qquad
\operatorname{AbsRel}=\frac{1}{N}\sum_{j=1}^{N}\frac{|\hat d_j-d_j|}{d_j},
\end{equation}
where $d_j$ and $\hat d_j$ are the ground-truth and predicted depths. The threshold accuracy $\delta_i$ is the fraction of pixels satisfying
\begin{equation}
\max\!\left(\frac{\hat d_j}{d_j},\frac{d_j}{\hat d_j}\right)<1.25^i,
\qquad i\in\{1,2,3\}.
\end{equation}

\paragraph{Shared training setup.}
We freeze the encoder and EMA pooling operator and repeat the pooled tokens to the original $16\times16$ grid. A ViT-XL task decoder with a newly initialized output head predicts segmentation logits or log depth. RGB initialization copies the corresponding image decoder's Transformer parameters, not its RGB output head. Both initialization settings use AdamW with batch size 32, $(\beta_1,\beta_2)=(0.9,0.999)$, weight decay 0.05, and gradient-norm clipping at 3. The learning rate warms up linearly from zero to $2\times10^{-4}$ over two epochs and then follows a cosine decay to $10^{-6}$. We train for 80 epochs on ADE20K and 50 on NYUv2. Images use ImageNet normalization and are resized to $256\times256$ for the frozen encoder.

\paragraph{ADE20K protocol.}
We use the official 150-class training and validation splits. Training applies random scaling in $[0.5,2]$, $512\times512$ crops, horizontal flips, and color jitter. Crops are retried up to ten times to avoid a single class occupying more than 75\% of valid pixels. Validation preserves aspect ratio, resizes the longest side to 512 pixels, and pads to a square. The decoder's logits are bilinearly resized to the $512\times512$ target grid. Training uses pixelwise cross-entropy, excluding unlabeled and padded pixels; these pixels are also excluded from evaluation. We select the checkpoint with the highest validation mIoU.

\paragraph{NYUv2 protocol.}
We retain $480\times640$ depth targets in meters and augment training images with paired horizontal flips and color jitter. The decoder predicts log depth, which is converted to metric depth before bilinear resizing to the target resolution. On valid training pixels, the loss is $\sqrt{\langle e^2\rangle-0.5\langle e\rangle^2+10^{-6}}$, where $e=\log\hat d-\log d$ and brackets denote the mean over valid pixels; no gradient-matching term is used. Valid depths lie in $[0.1,10]$ meters. Evaluation additionally restricts pixels to rows $[45,471)$ and columns $[41,601)$ in zero-based coordinates, clips predictions to the same depth range, and uses no scale alignment. Metrics aggregate over valid pixels, and we select the checkpoint with the lowest validation AbsRel.

\paragraph{Training from scratch.}
We initialize the ViT-XL task model randomly while keeping the frozen representations and task-specific training recipes fixed.

\begin{table}[htbp]
\caption{ADE20K semantic segmentation with task models trained from scratch. Accuracy metrics are percentages.}
\label{tab:ade20k-from-scratch}
\centering
\small
\setlength{\tabcolsep}{3.5pt}
\begin{tabular}{lrrrrr}
\toprule
Spatial pooling & Best epoch & Val. loss $\downarrow$ & mIoU $\uparrow$ & Mean acc. $\uparrow$ & Pixel acc. $\uparrow$ \\
\midrule
Uncompressed & 75 & 0.671 & 46.860 & 58.060 & 81.260 \\
Learned $2\times2$ & 75 & 0.807 & 39.380 & 50.100 & 77.480 \\
Average $2\times2$ & 65 & 0.735 & 42.591 & 53.477 & 78.905 \\
Learned $4\times4$ & 80 & 0.863 & 38.610 & 48.850 & 77.780 \\
Average $4\times4$ & 75 & 0.866 & 36.758 & 46.969 & 75.715 \\
\bottomrule
\end{tabular}
\end{table}

\begin{table}[htbp]
\caption{NYUv2 monocular depth estimation with task models trained from scratch. The $\delta$ metrics are percentages.}
\label{tab:nyuv2-from-scratch}
\centering
\normalsize
\setlength{\tabcolsep}{3.5pt}
\begin{tabular}{@{}lrrrrrrrr@{}}
\toprule
\shortstack[l]{Spatial\\pooling} & \shortstack{Best\\epoch} & AbsRel $\downarrow$ & RMSE $\downarrow$ & \shortstack{Log\\RMSE $\downarrow$} & SiLog $\downarrow$ & $\delta_1$ $\uparrow$ & $\delta_2$ $\uparrow$ & $\delta_3$ $\uparrow$ \\
\midrule
Uncompressed & 12 & 0.082 & 0.409 & 0.122 & 0.122 & 94.170 & 99.030 & 99.780 \\
Learned $2\times2$ & 21 & 0.090 & 0.455 & 0.134 & 0.134 & 92.780 & 98.700 & 99.720 \\
Average $2\times2$ & 14 & 0.087 & 0.434 & 0.131 & 0.131 & 93.200 & 98.720 & 99.730 \\
Learned $4\times4$ & 27 & 0.102 & 0.490 & 0.148 & 0.148 & 90.440 & 98.260 & 99.610 \\
Average $4\times4$ & 15 & 0.100 & 0.486 & 0.149 & 0.149 & 90.580 & 98.060 & 99.560 \\
\bottomrule
\end{tabular}
\end{table}

\paragraph{RGB-decoder initialization.}
For the learned tokenizers, we initialize the ViT-XL task model from the corresponding RGB decoder and otherwise retain the same downstream protocol.

\begin{table}[htbp]
\caption{ADE20K semantic segmentation with task models initialized from the corresponding RGB decoder. Accuracy metrics are percentages.}
\label{tab:ade20k-rgbinit}
\centering
\small
\setlength{\tabcolsep}{3.5pt}
\resizebox{\textwidth}{!}{%
\begin{tabular}{lrrrrrrr}
\toprule
Spatial pooling & Tokens & Compression & Best epoch & Val. loss $\downarrow$ & mIoU $\uparrow$ & Mean acc. $\uparrow$ & Pixel acc. $\uparrow$ \\
\midrule
Uncompressed          & 256 & $1\times$  & 80 & 0.694 & 47.810 & 58.850 & 81.770 \\
Learned $2\times2$ & 64  & $4\times$  & 60 & 0.709 & 45.380 & 56.420 & 81.010 \\
Learned $4\times4$ & 16  & $16\times$ & 80 & 0.788 & 42.500 & 53.240 & 79.960 \\
\bottomrule
\end{tabular}%
}
\end{table}

\begin{table}[H]
\caption{NYUv2 monocular depth estimation with task models initialized from the corresponding RGB decoder. The $\delta$ metrics are percentages.}
\label{tab:nyuv2-rgbinit}
\centering
\normalsize
\setlength{\tabcolsep}{3.5pt}
\begin{tabular}{@{}lrrrrrrrr@{}}
\toprule
\shortstack[l]{Spatial\\pooling} & \shortstack{Best\\epoch} & AbsRel $\downarrow$ & RMSE $\downarrow$ & \shortstack{Log\\RMSE $\downarrow$} & SiLog $\downarrow$ & $\delta_1$ $\uparrow$ & $\delta_2$ $\uparrow$ & $\delta_3$ $\uparrow$ \\
\midrule
Uncompressed & 14 & 0.081 & 0.411 & 0.123 & 0.123 & 94.330 & 98.980 & 99.750 \\
Learned $2\times2$ & 27 & 0.089 & 0.456 & 0.133 & 0.133 & 92.800 & 98.720 & 99.700 \\
Learned $4\times4$ & 24 & 0.101 & 0.495 & 0.147 & 0.147 & 90.610 & 98.220 & 99.610 \\
\bottomrule
\end{tabular}
\end{table}

\section{Guidance ablations}
\label{app:guidance-ablations}

We compare CFG, internal guidance, and guidance from the encoder-reconstruction head. All evaluations use 50,000 ImageNet-256 samples and 100 Euler steps. The sweeps below use the 80-epoch models (Table~\ref{tab:guidance-sweep-best}); the longer-trained $2\times2$ and $4\times4$ checkpoints are included separately in the summary tables (Tables~\ref{tab:full-generation-results} and~\ref{tab:selected-guidance-scales}).

\subsection{Full generation comparison}

\begin{table}[H]
\caption{ImageNet-256 generation at 100 Euler steps. Models train for 80 epochs, except the extended-training $2\times2$ (180 epochs) and $4\times4$ (300 epochs) variants. Guided columns use the settings in Table~\ref{tab:selected-guidance-scales}; Appendix~\ref{app:training-duration} separately reports longer runs and finer IG sweeps. -- denotes an unevaluated setting. Best and second-best metrics in each column are bold and underlined.}
\label{tab:full-generation-results}
\centering
\small
\setlength{\tabcolsep}{4pt}
\begin{tabular}{@{}lrr|rr|rr|rr@{}}
\toprule
& \multicolumn{2}{c|}{Unguided} &
\multicolumn{2}{c|}{CFG} &
\multicolumn{2}{c|}{IG} &
\multicolumn{2}{c}{CFG + IG} \\
\cmidrule(lr){2-3}\cmidrule(lr){4-5}\cmidrule(lr){6-7}\cmidrule(lr){8-9}
Variant & FID $\downarrow$ & IS $\uparrow$ & FID $\downarrow$ & IS $\uparrow$ & FID $\downarrow$ & IS $\uparrow$ & FID $\downarrow$ & IS $\uparrow$ \\
\midrule
$1\times1$ & \textbf{1.53} & \textbf{226.63} & \textbf{1.45} & 237.80 & \underline{1.08} & \underline{262.00} & 1.07 & 267.15 \\
$2\times2$ & 3.00 & 184.49 & \underline{1.99} & 228.10 & 1.09 & 250.36 & \underline{1.07} & 268.50 \\
$2\times4$ & 4.91 & 161.42 & 2.33 & \underline{238.80} & 1.19 & 249.58 & 1.16 & 273.06 \\
$4\times2$ & 4.97 & 159.67 & 2.41 & 237.63 & 1.21 & 249.96 & 1.19 & \underline{273.98} \\
$4\times4$ & 8.16 & 132.67 & 2.87 & \textbf{249.72} & 1.44 & 247.07 & 1.35 & \textbf{282.29} \\
\midrule
$2\times2$ (180 epochs) & \underline{2.78} & \underline{192.37} & -- & -- & \textbf{1.05} & \textbf{272.00} & \textbf{1.05} & 272.80 \\
$4\times4$ (300 epochs) & 7.19 & 146.13 & -- & -- & 1.29 & 258.04 & -- & -- \\
\bottomrule
\end{tabular}
\end{table}

For the 80-epoch models, while IG alone improves generation across all compression rates, combining it with CFG yields the lowest overall FID, with the largest additional reduction occurring at $16\times$ compression. However, this combined advantage appears to diminish with longer training: for the 180-epoch $2\times2$ checkpoint, adding CFG to IG yields no further FID improvement.

\begin{table}[H]
\caption{Guidance scales selected for Tables~\ref{tab:main-generation-results}, \ref{tab:extended-training-generation}, and~\ref{tab:full-generation-results}. For CFG alone, $s_{\mathrm{IG}}=1$; for IG alone, $s_{\mathrm{CFG}}=1$. Both scales are one for unguided sampling.}
\label{tab:selected-guidance-scales}
\centering
\small
\begin{tabular}{lrr|rr}
\toprule
& CFG & IG & \multicolumn{2}{c}{CFG + IG} \\
\cmidrule(lr){2-2}\cmidrule(lr){3-3}\cmidrule(lr){4-5}
Variant & $s_{\mathrm{CFG}}$ & $s_{\mathrm{IG}}$ & $s_{\mathrm{CFG}}$ & $s_{\mathrm{IG}}$ \\
\midrule
$1\times1$ & 3.00 & 1.75 & 2.50 & 1.78 \\
$2\times2$ & 3.00 & 1.75 & 1.75 & 1.75 \\
$2\times4$ & 3.00 & 2.00 & 1.50 & 2.00 \\
$4\times2$ & 3.00 & 2.00 & 1.50 & 2.00 \\
$4\times4$ & 3.00 & 2.75 & 1.75 & 2.25 \\
\midrule
Average $2\times2$ & -- & 2.00 & -- & -- \\
\midrule
$2\times2$ (180 epochs) & -- & 2.00 & 2.00 & 2.00 \\
$4\times4$ (300 epochs) & -- & 2.75 & -- & -- \\
\bottomrule
\end{tabular}
\end{table}

\subsection{Isolated guidance sweeps}

\paragraph{Encoder-reconstruction guidance.}
The encoder-reconstruction head provides an alternative guidance signal. We first pass its predicted dense encoder field through the frozen tokenizer and the same latent normalization:
\begin{align}
    \bar z_{\mathrm{rec}}^{E}
    =
    \frac{
        \mathcal P\!\left(R_\theta(h^{(8)})\right)-\mu_{\hat z}
    }{
        \sigma_{\hat z}
    }.
\end{align}
The guided prediction is
\begin{align}
    \bar z_{\mathrm{guided}}^{E}
    =
    \bar z_{\mathrm{full}}
    +
    s_{\mathrm{rec}}^{E}
    \left(
        \bar z_{\mathrm{full}}-\bar z_{\mathrm{rec}}^{E}
    \right),
\end{align}
where $s_{\mathrm{rec}}^{E}=0$ is neutral. We refer to this procedure as encoder-reconstruction guidance.

We test CFG, encoder-reconstruction guidance, and internal guidance in isolation. In each sweep, we vary one guidance scale and disable the other two methods, using EMA weights and the same validation conditions. CFG and internal guidance are neutral at scale one, whereas encoder-reconstruction guidance is neutral at zero.

\begin{table}[htbp]
\caption{Sampling settings for the guidance ablations. Intervals use noise-to-data time: $t=0$ is noise and $t=1$ is clean data.}
\label{tab:sampling-details}
\centering
\small
\begin{tabular}{lcccc}
\toprule
Method & Euler steps & Guidance scale & Guidance interval & Neutral value \\
\midrule
Unguided & 100 & -- & -- & CFG 1 \\
Internal guidance & 100 & Swept & $[0,0.9]$ & 1 \\
Encoder-reconstruction guidance & 100 & Swept & $[0,1]$ & 0 \\
\bottomrule
\end{tabular}
\end{table}

\input{figures/isolated_guidance_sweeps.tex}

Internal guidance gives the lowest observed isolated FID at every compression rate. CFG alone improves throughout the tested range, but becomes less effective as compression increases. Encoder-reconstruction guidance has a narrower optimum and degrades rapidly at large scales. Inception Score often continues to improve after FID reaches its minimum.

\subsection{Joint guidance searches}
\label{app:cfg-sweeps}

We jointly vary CFG with each auxiliary guidance method. The search uses irregular refinement grids, so the figures show measured configurations without interpolating unevaluated settings.

\begin{table}[htbp]
\caption{Metrics at the lowest-FID point in each joint guidance sweep for the 80-epoch models. Both auxiliary methods are combined with CFG.}
\label{tab:guidance-sweep-best}
\centering
\small
\setlength{\tabcolsep}{3.5pt}
\begin{tabular}{lrrrr}
\toprule
& \multicolumn{2}{c}{Encoder-reconstruction guidance} & \multicolumn{2}{c}{Internal guidance} \\
\cmidrule(lr){2-3}\cmidrule(lr){4-5}
Variant & FID $\downarrow$ & IS $\uparrow$ & FID $\downarrow$ & IS $\uparrow$ \\
\midrule
$2\times2$ & \textbf{1.06} & 272.83 & 1.07 & 268.50 \\
$2\times4$ & 1.17 & 279.68 & \textbf{1.16} & 273.06 \\
$4\times2$ & 1.20 & 280.14 & \textbf{1.19} & 273.98 \\
$4\times4$ & \textbf{1.35} & 295.39 & \textbf{1.35} & 282.29 \\
\bottomrule
\end{tabular}
\end{table}

After joint tuning with CFG, the two auxiliary guidance methods achieve similar FID. Encoder-reconstruction guidance gives the lowest observed value at $2\times2$, while internal guidance is stronger at $2\times4$ and $4\times2$; their $4\times4$ results are nearly identical. These comparisons concern sampling-time guidance: both auxiliary heads were trained in every model.

\input{figures/joint_guidance_sweep_data.tex}
\input{figures/joint_guidance_sweeps.tex}

\clearpage

\section{Learned pooling operator analysis}
\label{app:operator-analysis}
\label{sec:operator-analysis}

Table~\ref{tab:operator-analysis} shows that learned pooling selects a row space distinct from spatial averaging and local PCA. Its adjusted alignment with both decreases as compression increases, and it captures less held-out feature variation than the training-fitted PCA baseline. These measurements characterize the learned operator but do not establish which retained directions support reconstruction or generation. The computation is described below.

\begin{table}[t]
\caption{Learned pooling row-space analysis. Alignment scores are adjusted so that zero indicates the overlap expected between random subspaces of the same dimension and one indicates identical subspaces. Energy captured is the squared validation feature energy preserved by the learned pooling operator's row space (after centering with the training mean), expressed as a percentage of that captured by PCA with the same output dimension.}
\label{tab:operator-analysis}
\centering
\small
\setlength{\tabcolsep}{7pt}
\begin{tabular}{lrrr}
\toprule
Window & Average alignment & PCA alignment & Energy captured (\% of PCA) \\
\midrule
$2\times2$ & 0.5000 & 0.2798 & 67.60 \\
$2\times4$ & 0.3530 & 0.2114 & 52.16 \\
$4\times2$ & 0.3501 & 0.2082 & 52.05 \\
$4\times4$ & 0.2268 & 0.1524 & 38.16 \\
\bottomrule
\end{tabular}
\end{table}

Let $x\in\mathbb R^{mC}$ be a flattened pooling window containing $m=p_xp_y$ patches with $C=1024$ channels. Learned pooling applies
\begin{equation}
    y=Ax+b, \qquad A\in\mathbb R^{C\times mC}.
\end{equation}
The input directions retained by the pooled token form the row space of $A$; the bias does not affect this space. We represent it with an orthonormal basis $Q_A\in\mathbb R^{mC\times C}$ obtained from the right singular vectors of $A$. All analyzed operators have numerical rank $C$.

Average pooling retains the spatially constant directions, with basis
\begin{equation}
    Q_{\mathrm{avg}}=\frac{1}{\sqrt m}
    \begin{bmatrix}I_C & I_C & \cdots & I_C\end{bmatrix}^{\!\top}.
\end{equation}
For PCA, $Q_{\mathrm{PCA}}$ contains the leading $C$ components fitted on centered training blocks using a randomized approximation. Because it is fitted on training data, it is not guaranteed to strictly maximize variance on the validation set. We measure alignment with either reference basis $Q_B$ as
\begin{equation}
    O(A,B)=\frac{1}{C}\left\|Q_A^\top Q_B\right\|_F^2
    =\frac{1}{C}\sum_{i=1}^{C}\cos^2\theta_i,
\end{equation}
where $\theta_i$ are their principal angles. The score ranges from zero for orthogonal subspaces to one for identical subspaces.

Rate-matched random subspaces have expected overlap $1/m$. This expectation holds for independent, uniformly oriented $C$-dimensional subspaces of $\mathbb R^{mC}$. To compare window sizes, we report the adjusted score
\begin{equation}
    O_{\mathrm{adj}}=\frac{O-1/m}{1-1/m}.
\end{equation}
This maps the random expectation to zero and identical subspaces to one. The adjusted score provides a normalized geometric comparison across pooling dimensions, not a statistical significance test, and can take negative values. Table~\ref{tab:operator-analysis} reports adjusted alignment, while Table~\ref{tab:operator-analysis-full} includes both raw and adjusted values.

\begin{table}[htbp]
\caption{Complete learned pooling row-space results. ``Random'' is the expected raw overlap of random subspaces of the same dimension. Energy captured is reported as a percentage of rate-matched PCA.}
\label{tab:operator-analysis-full}
\centering
\small
\setlength{\tabcolsep}{3.5pt}
\begin{tabular}{lrrrrrr}
\toprule
& & \multicolumn{2}{c}{Average overlap} & \multicolumn{2}{c}{PCA overlap} & \\
\cmidrule(lr){3-4}\cmidrule(lr){5-6}
Window & Random & Raw & Adjusted & Raw & Adjusted & Energy (\% PCA) \\
\midrule
$2\times2$ & 0.2500 & 0.6250 & 0.5000 & 0.4598 & 0.2798 & 67.60 \\
$2\times4$ & 0.1250 & 0.4339 & 0.3530 & 0.3100 & 0.2114 & 52.16 \\
$4\times2$ & 0.1250 & 0.4313 & 0.3501 & 0.3072 & 0.2082 & 52.05 \\
$4\times4$ & 0.0625 & 0.2751 & 0.2268 & 0.2054 & 0.1524 & 38.16 \\
\bottomrule
\end{tabular}
\end{table}

The average-pooling subspace is also the zero-frequency, or DC, component of a two-dimensional discrete cosine transform. Thus, $O(A,\mathrm{avg})$ measures alignment with the spatial average. The complement $1-O(A,\mathrm{avg})$ measures the fraction of orthonormalized row-space energy outside the spatial-average subspace, not the fraction of input variance retained.

The analysis uses the clean EMA tokenizer at decoder step 40032. For each window geometry, we sample 4096 blocks from one ImageNet training image per class and 4096 blocks from one disjoint validation image per class. A randomized $C$-component PCA is fitted only on the training blocks. The analysis uses one checkpoint per geometry and a finite sample of blocks; small differences should therefore be interpreted cautiously.

After centering validation blocks $X$ with the training mean, we measure the fraction of squared feature energy captured by an orthonormal basis $Q$:
\begin{equation}
    V(Q)=\frac{\lVert XQ\rVert_F^2}{\lVert X\rVert_F^2}.
\end{equation}
We report $100\,V(Q_A)/V(Q_{\mathrm{PCA}})$, comparing learned pooling with the rate-matched PCA baseline. This projection-based quantity is invariant to kernel scaling. Raw-kernel DCT energy is not invariant to invertible output-channel mixing and is therefore excluded as an intrinsic retained-subspace statistic.

\input{figures/generated_samples.tex}

\end{document}

%% file: math_commands.tex
\usepackage{amsmath,amsfonts,bm}

\def\eqref#1{equation~\ref{#1}}
\def\1{\bm{1}}

\DeclareMathAlphabet{\mathsfit}{\encodingdefault}{\sfdefault}{m}{sl}
\SetMathAlphabet{\mathsfit}{bold}{\encodingdefault}{\sfdefault}{bx}{n}

%% file: pooldino_style.tex
\definecolor{poolRed}{HTML}{FF0000}
\definecolor{poolTeal}{HTML}{00A08A}
\definecolor{poolGold}{HTML}{F2AD00}
\definecolor{poolPurple}{HTML}{884EA0}
\definecolor{poolCyan}{HTML}{5BBCD6}

\newif\ifpoolShowCFGIG
\poolShowCFGIGfalse

\newcommand{\poolSpeedupAtRow}[4]{%
  \pgfplotstablegetelem{#1}{batch}\of{#2}%
  \let\poolSpeedupBatch\pgfplotsretval
  \pgfplotstablegetelem{#1}{mean}\of{#2}%
  \let\poolSpeedupBase\pgfplotsretval
  \pgfplotstablegetelem{#1}{mean}\of{#3}%
  \let\poolSpeedupPooled\pgfplotsretval
  \pgfmathsetmacro{\poolSpeedupRatio}{\poolSpeedupPooled/\poolSpeedupBase}%
  \coordinate (#4-base) at (axis cs:\poolSpeedupBatch,\poolSpeedupBase);
  \coordinate (#4-pooled) at (axis cs:\poolSpeedupBatch,\poolSpeedupPooled);
  \expandafter\xdef\csname poolSpeedupRatio#4\endcsname{\poolSpeedupRatio}%
}
\newcommand{\poolDrawSpeedup}[1]{%
  \draw[{Latex[length=1.5mm,width=1.1mm]}-{Latex[length=1.5mm,width=1.1mm]},
    draw=black!85, line width=0.8pt, shorten <=3pt, shorten >=3pt,
    preaction={draw=white,line width=2.8pt}]
    (#1-base) --
    node[midway, anchor=east, xshift=-4pt, font=\scriptsize,
      text=black!85, fill=white, inner sep=1.5pt]
      {$\pgfmathprintnumber[fixed,precision=2,zerofill]{\csname poolSpeedupRatio#1\endcsname}\times$}
    (#1-pooled);
}

\tikzset{
  pool arrow/.style={-{Latex[length=2.2mm,width=1.5mm]}, draw=black!80, line width=0.8pt},
  pool model/.style={draw=black!80, fill=white, rounded corners=1.5pt,
    line width=0.8pt, minimum height=1.15cm, align=center, font=\footnotesize},
  pool stage label/.style={align=center, font=\footnotesize, text=black!80},
  pool operation label/.style={align=center, font=\footnotesize, text=black!85,
    fill=white, inner sep=1.5pt},
  pool process/.style={draw=black!70, fill=poolCyan!7, rounded corners=1.5pt,
    line width=0.8pt, minimum height=0.88cm, align=center, font=\footnotesize},
  pool head/.style={draw=black!65, fill=poolGold!7, rounded corners=1.5pt,
    line width=0.7pt, minimum height=0.68cm, align=center, font=\scriptsize},
  pool loss/.style={draw=poolPurple!75!black, fill=poolPurple!6,
    rounded corners=1.5pt, line width=0.75pt, minimum height=0.68cm,
    align=center, font=\footnotesize},
  pool auxiliary arrow/.style={-{Latex[length=1.8mm,width=1.25mm]},
    draw=black!55, line width=0.65pt},
  pool target line/.style={draw=black!42, line width=0.55pt},
  pool target arrow/.style={pool target line,
    -{Latex[length=1.6mm,width=1.1mm]}},
  pool patch/.style={draw=white, line width=0.45pt},
  pool token/.style={draw=black!55, line width=0.5pt},
  pool group/.style={draw=black!50, line width=0.55pt},
  pics/pool image/.style={code={
    \path[fill=poolCyan!18, rounded corners=1pt] (-0.72,-0.72) rectangle (0.72,0.72);
    \begin{scope}
      \clip[rounded corners=1pt] (-0.72,-0.72) rectangle (0.72,0.72);
      \fill[poolGold!95] (0.43,0.40) circle (0.14);
      \fill[poolPurple!72] (-0.82,-0.28) -- (-0.24,0.38) -- (0.19,-0.22) --
        (0.48,0.10) -- (0.82,-0.38) -- (0.82,-0.82) -- (-0.82,-0.82) -- cycle;
      \fill[poolTeal!88] (-0.82,-0.42) .. controls (-0.35,-0.29) and (0.20,-0.58) ..
        (0.82,-0.34) -- (0.82,-0.82) -- (-0.82,-0.82) -- cycle;
    \end{scope}
    \draw[black!70, rounded corners=1pt, line width=0.65pt]
      (-0.72,-0.72) rectangle (0.72,0.72);
  }}
}

%% file: figures/quality_throughput.tex
\begin{figure}[H]
\centering
% The previous Figure 1 with larger labels and leader lines is still available:
% figures/archive/quality_throughput_large_labels.tex (editable source) and .pdf.
% To restore it, replace the input below with the archived .tex file.
\input{figures/quality_throughput_sampling_plot.tex}
\caption{Generation quality versus latent-sampling throughput on an NVIDIA H100 NVL at batch size 128. PoolDINO and the RAEv2 reference use Internal Guidance (IG) with 100 Euler steps. Red points use 80 training epochs; gold points use extended training (180 epochs for $2\times2$ and 300 for $4\times4$). Diamonds pair published FIDs with measured sampling rates. Appendix~\ref{app:throughput} details the measurement protocols and shows that reducing to 50 sampling steps doubles throughput while maintaining comparable generation quality.}
\label{fig:quality-throughput}
\end{figure}

%% file: figures/quality_throughput_sampling_plot.tex
% Full-sampling measurements use corrected pooling-specific time shifts.
% Provenance: figures/quality_throughput_measurements.json.
% Extended-training FIDs: logs/REPEATCONV2X2_180_RESULTS.tex and logs/REPEATCONV4X4_300_RESULTS.tex.
% PoolDINO uses IG alone. Literature FIDs are published annotations.
\begingroup
\begin{tikzpicture}
\begin{axis}[
    scale only axis,
    width=0.87\textwidth, height=0.34\textwidth,
    % Left annotation padding keeps baseline labels inside the plotting area.
    xmin=-12, xmax=65, xtick={0,10,20,30,40,50,60},
    ymin=0.70, ymax=2.70, ytick={1.0,1.3,1.6,1.9,2.2,2.5},
    xlabel={Latent samples/s $\rightarrow$}, ylabel={FID $\downarrow$},
    scaled ticks=false,
    tick label style={font=\scriptsize},
    label style={font=\footnotesize},
    ylabel style={rotate=-90},
    axis x line*=bottom, axis y line*=left,
    axis line style={black!45}, tick style={black!45},
    tick align=outside, xtick pos=bottom, ytick pos=left,
    xmajorgrids=false, ymajorgrids=true, grid style={black!8},
    legend style={at={(1,1)},anchor=north east,outer sep=0pt,
        draw=black!45,line width=0.4pt,fill=white,
        inner sep=3pt,font=\footnotesize,cells={anchor=west}},
    clip=false,
]
\addlegendimage{poolRed,line width=1pt,mark=*,mark size=2.3pt}
\addlegendentry{PoolDINO: 80 epochs}
\addlegendimage{poolGold,line width=1pt,mark=*,mark size=2.3pt}
\addlegendentry{PoolDINO: extended training}

% A guide connects the discrete pooling variants; no interpolated frontier.
\addplot[poolRed,line width=1pt,forget plot] coordinates {
    (6.5162377754015415,1.083134)
    (24.15883440694129,1.088975)
    (40.10041655873189,1.19126)
    (58.457215615361896,1.444216)
};
% All PoolDINO and unpooled-reference points use 100 steps.
\addplot[poolGold,line width=1pt,forget plot] coordinates {
    (6.5162377754015415,1.083134)
    (24.15883440694129,1.0525499986095497)
    (58.457215615361896,1.28819)
};
\addplot[only marks,black!80,mark=*,mark size=2.3pt,
    forget plot,error bars/.cd,x dir=both,x explicit]
    coordinates {(6.5162377754015415,1.083134) +- (0.008849058916579925,0)};
\node[anchor=north,font=\scriptsize,text=black!80,
    fill=white,inner sep=1pt,yshift=-5pt]
    at (axis cs:6.5162377754015415,1.083134) {RAEv2};

\addplot[only marks,poolRed,mark=*,mark size=2.3pt,
    forget plot,error bars/.cd,x dir=both,x explicit]
    coordinates {(24.15883440694129,1.088975) +- (0.05262807260879476,0)};
\node[anchor=south,font=\scriptsize,text=poolRed,xshift=-3pt,yshift=5pt]
    at (axis cs:24.15883440694129,1.088975) {$4\times$};
\addplot[only marks,poolRed,mark=*,mark size=2.3pt,
    forget plot,error bars/.cd,x dir=both,x explicit]
    coordinates {(40.10041655873189,1.19126) +- (0.0720525749532469,0)};
\node[anchor=south,font=\scriptsize,text=poolRed,yshift=5pt]
    at (axis cs:40.10041655873189,1.19126) {$8\times$};
\addplot[only marks,poolRed,mark=*,mark size=2.3pt,
    forget plot,error bars/.cd,x dir=both,x explicit]
    coordinates {(58.457215615361896,1.444216) +- (0.19728694433753302,0)};
\node[anchor=south,font=\scriptsize,text=poolRed,xshift=-6pt,yshift=5pt]
    at (axis cs:58.457215615361896,1.444216) {$16\times$};

% Architecture-level 100-step timings are shared across training durations.
\addplot[only marks,poolGold,mark=*,mark size=2.3pt,
    forget plot,error bars/.cd,x dir=both,x explicit] coordinates {
    (24.15883440694129,1.0525499986095497) +- (0.05262807260879476,0)
};
\node[anchor=north,font=\scriptsize,text=poolGold,yshift=-5pt,align=center]
    at (axis cs:24.15883440694129,1.0525499986095497) {$4\times$\\[-0.5ex]180 epochs};

\input{figures/quality_throughput_literature_points.tex}
\addplot[only marks,poolGold,mark=*,mark size=2.3pt,
    forget plot,error bars/.cd,x dir=both,x explicit]
    coordinates {(58.457215615361896,1.28819) +- (0.19728694433753302,0)};
\node[anchor=north,font=\scriptsize,text=poolGold,
    yshift=-5pt,align=center]
    at (axis cs:58.457215615361896,1.28819) {$16\times$\\[-0.5ex]300 epochs};
\end{axis}
\end{tikzpicture}
\endgroup

%% file: figures/pooled_decoder_training.tex
\begin{tikzpicture}[x=1cm,y=1cm]
  \def\poolcell{0.42}

  % Invisible stage bounds make every inter-stage gap and arrow consistent,
  % independent of the rendered width of the model labels.
  \node[inner sep=0pt, minimum width=1.44cm, minimum height=1.44cm] (input) {};
  \node[pool model, fill=poolRed!4, minimum width=1.45cm, right=0.52cm of input]
    (dino) {DINOv3-L/16};
  \node[inner sep=0pt, minimum width=1.68cm, minimum height=1.68cm,
    right=0.52cm of dino] (dense) {};
  \node[inner sep=0pt, minimum width=0.84cm, minimum height=0.84cm,
    right=0.72cm of dense] (pooled) {};
  \node[inner sep=0pt, minimum width=1.68cm, minimum height=1.68cm,
    right=0.72cm of pooled] (repeated) {};
  \node[pool model, fill=poolGold!7, minimum width=1.45cm,
    right=0.52cm of repeated] (decoder) {ViT-XL\\decoder};
  \node[inner sep=0pt, minimum width=1.44cm, minimum height=1.44cm,
    right=0.52cm of decoder] (output) {};

  % Compact status markers: the encoder is frozen and the decoder is trained.
  \node[anchor=south east, inner sep=0pt]
    at ([xshift=-0.08cm,yshift=0.08cm]dino.south east)
    {\twemoji[height=0.22cm]{snowflake}};
  \node[anchor=south east, inner sep=0pt]
    at ([xshift=-0.08cm,yshift=0.08cm]decoder.south east)
    {\twemoji[height=0.22cm]{fire}};

  % Input and reconstruction thumbnails.
  \pic at (input.center) {pool image};
  \pic at (output.center) {pool image};
  \node[pool stage label, below=0.12cm of input] {$x$};
  \node[pool stage label, below=0.12cm of output] {$\hat x$};

  % Dense 4x4 schematic. Each 2x2 window shares a hue, while every source
  % patch has a distinct shade.
  \begin{scope}[shift={(dense.south west)}]
    \foreach \gx/\gy/\base in {0/1/poolCyan,1/1/poolTeal,0/0/poolGold,1/0/poolPurple} {
      \foreach \lx/\ly/\tone in {0/0/42,1/0/55,0/1/68,1/1/82} {
        \pgfmathsetmacro{\px}{(2*\gx+\lx)*\poolcell}
        \pgfmathsetmacro{\py}{(2*\gy+\ly)*\poolcell}
        \path[pool patch, fill=\base!\tone!white]
          (\px,\py) rectangle ++(\poolcell,\poolcell);
      }
      \pgfmathsetmacro{\groupx}{2*\gx*\poolcell}
      \pgfmathsetmacro{\groupy}{2*\gy*\poolcell}
      \draw[pool group] (\groupx,\groupy) rectangle ++(0.84,0.84);
    }
  \end{scope}
  \node[pool stage label, below=0.12cm of dense] {$z$\\$16\!\times\!16$ tokens};

  % One learned projection per non-overlapping 2x2 window.
  \begin{scope}[shift={(pooled.south west)}]
    \foreach \gx/\gy/\base in {0/1/poolCyan,1/1/poolTeal,0/0/poolGold,1/0/poolPurple} {
      \pgfmathsetmacro{\px}{\gx*\poolcell}
      \pgfmathsetmacro{\py}{\gy*\poolcell}
      \path[pool token, fill=\base!72!white]
        (\px,\py) rectangle ++(\poolcell,\poolcell);
    }
  \end{scope}
  \node[pool stage label, below=0.12cm of pooled] {$\hat z$};

  % Repeat each pooled token back over its source window.
  \begin{scope}[shift={(repeated.south west)}]
    \foreach \gx/\gy/\base in {0/1/poolCyan,1/1/poolTeal,0/0/poolGold,1/0/poolPurple} {
      \foreach \lx/\ly in {0/0,1/0,0/1,1/1} {
        \pgfmathsetmacro{\px}{(2*\gx+\lx)*\poolcell}
        \pgfmathsetmacro{\py}{(2*\gy+\ly)*\poolcell}
        \path[pool patch, fill=\base!72!white]
          (\px,\py) rectangle ++(\poolcell,\poolcell);
      }
      \pgfmathsetmacro{\groupx}{2*\gx*\poolcell}
      \pgfmathsetmacro{\groupy}{2*\gy*\poolcell}
      \draw[pool group] (\groupx,\groupy) rectangle ++(0.84,0.84);
    }
  \end{scope}
  \node[pool stage label, below=0.12cm of repeated]
    {decoder input\\$16\!\times\!16$ tokens};

  % Identical arrow clearance on both sides of every stage.
  \draw[pool arrow, shorten <=3pt, shorten >=3pt] (input.east) -- (dino.west);
  \draw[pool arrow, shorten <=3pt, shorten >=3pt] (dino.east) -- (dense.west);
  \draw[pool arrow, shorten <=3pt, shorten >=3pt]
    (dense.east) -- node[pool operation label, above=0.11cm] {$\mathcal P$} (pooled.west);
  \draw[pool arrow, shorten <=3pt, shorten >=3pt]
    (pooled.east) -- node[pool operation label, font=\scriptsize, above=0.11cm] {repeat} (repeated.west);
  \draw[pool arrow, shorten <=3pt, shorten >=3pt] (repeated.east) -- (decoder.west);
  \draw[pool arrow, shorten <=3pt, shorten >=3pt] (decoder.east) -- (output.west);
\end{tikzpicture}

%% file: figures/curated_compression_samples.tex
% Rows use cherry_pick.py displayed IDs (shuffle enabled, seed 42), in user order.
\begin{figure}[tbp]
\centering
\begingroup
\small
\setlength{\tabcolsep}{0pt}
\renewcommand{\arraystretch}{0}
\newcommand{\poolCuratedImage}[2]{\includegraphics[width=\dimexpr0.25\textwidth-2.25pt\relax]{figures/samples/curated/#1/#2.jpg}}
\begin{tabular}{@{}c@{\hspace{3pt}}c@{\hspace{3pt}}c@{\hspace{3pt}}c@{}}
\shortstack{$1\times1$\\\scriptsize Unpooled} &
\shortstack{$2\times2$\\\scriptsize $4\times$ compression} &
\shortstack{$2\times4$\\\scriptsize $8\times$ compression} &
\shortstack{$4\times4$\\\scriptsize $16\times$ compression} \\[5pt]
\poolCuratedImage{pool1x1-ep80}{0000} & \poolCuratedImage{repeatconv2x2-ep80}{0694} & \poolCuratedImage{repeatconv2x4-ep80}{0907} & \poolCuratedImage{repeatconv4x4-ep80}{0032} \\[3pt]
\poolCuratedImage{pool1x1-ep80}{0035} & \poolCuratedImage{repeatconv2x2-ep80}{0696} & \poolCuratedImage{repeatconv2x4-ep80}{0640} & \poolCuratedImage{repeatconv4x4-ep80}{0003} \\[3pt]
\poolCuratedImage{pool1x1-ep80}{0098} & \poolCuratedImage{repeatconv2x2-ep80}{0585} & \poolCuratedImage{repeatconv2x4-ep80}{0516} & \poolCuratedImage{repeatconv4x4-ep80}{0154} \\[3pt]
\poolCuratedImage{pool1x1-ep80}{0452} & \poolCuratedImage{repeatconv2x2-ep80}{0306} & \poolCuratedImage{repeatconv2x4-ep80}{0584} & \poolCuratedImage{repeatconv4x4-ep80}{0095}
\end{tabular}
\endgroup
\caption{Curated samples across spatial compression levels, with all generators trained for 80 epochs. Samples are selected independently across columns. Sampling uses 100 Euler steps and each model's selected IG-only scale. Appendix~\ref{app:generated-samples} provides randomly selected samples without quality filtering.}
\label{fig:curated-compression}
\end{figure}

%% file: figures/curated_training_samples.tex
% Four user-selected rows of cherry_pick.py displayed IDs (shuffle, seed 42).
% IDs differ between the left and right pairs; filenames preserve UI IDs.
\begin{figure}[tbp]
\centering
\begingroup
\small
\setlength{\tabcolsep}{0pt}
\renewcommand{\arraystretch}{0}
\newcommand{\poolCuratedImage}[2]{\includegraphics[width=\dimexpr0.25\textwidth-4pt\relax]{figures/samples/curated/#1/#2.jpg}}
\begin{tabular}{@{}c@{\hspace{3pt}}c@{\hspace{10pt}}c@{\hspace{3pt}}c@{}}
\multicolumn{2}{c}{$2\times2$ pooling} & \multicolumn{2}{c}{$4\times4$ pooling} \\[4pt]
\cmidrule(lr){1-2}\cmidrule(lr){3-4}
80 epochs & 180 epochs & 80 epochs & 300 epochs \\[6pt]
\poolCuratedImage{repeatconv2x2-ep80}{0487} & \poolCuratedImage{repeatconv2x2-ep180}{0487} & \poolCuratedImage{repeatconv4x4-ep80}{0014} & \poolCuratedImage{repeatconv4x4-ep300}{0014} \\[3pt]
\poolCuratedImage{repeatconv2x2-ep80}{0255} & \poolCuratedImage{repeatconv2x2-ep180}{0255} & \poolCuratedImage{repeatconv4x4-ep80}{0017} & \poolCuratedImage{repeatconv4x4-ep300}{0017} \\[3pt]
\poolCuratedImage{repeatconv2x2-ep80}{0821} & \poolCuratedImage{repeatconv2x2-ep180}{0821} & \poolCuratedImage{repeatconv4x4-ep80}{0518} & \poolCuratedImage{repeatconv4x4-ep300}{0518} \\[3pt]
\poolCuratedImage{repeatconv2x2-ep80}{0864} & \poolCuratedImage{repeatconv2x2-ep180}{0864} & \poolCuratedImage{repeatconv4x4-ep80}{0551} & \poolCuratedImage{repeatconv4x4-ep300}{0551}
\end{tabular}
\endgroup
\caption{Curated comparisons of 80-epoch and extended training. Class conditions and initial noise are matched within each pair. All samples use 100 Euler steps and IG alone: scales 1.75/2.00 for the 80/180-epoch $2\times2$ models, and 2.75 for both $4\times4$ models.}
\label{fig:curated-training}
\end{figure}

%% file: figures/pooled_generator_training.tex
\begin{tikzpicture}[x=1cm,y=1cm]
  \def\poolcell{0.34}

  % Main training path.
  \node[inner sep=0pt, minimum width=1.44cm, minimum height=1.44cm]
    (input) at (0,0.85) {};
  \node[pool model, fill=poolRed!4, minimum width=1.52cm]
    (dino) at (2.35,0.85) {DINOv3-L/16};
  \node[pool process, fill=poolTeal!6, minimum width=0.78cm]
    (pooling) at (4.35,0.85) {$\mathcal P$};
  \node[inner sep=0pt, minimum width=0.68cm, minimum height=0.68cm]
    (noisy) at (6.00,0.85) {};
  \node[pool model, fill=poolGold!7, minimum width=1.72cm]
    (generator) at (8.40,0.85) {DDT generator\\$G_\theta$};
  \node[pool loss, minimum width=1.04cm]
    (lfm) at (11.80,0.85) {$\mathcal L_{\mathrm{FM}}$};

  \begin{scope}[shift={(input.center)}]
    \pic {pool image};
  \end{scope}
  \node[pool stage label, below=0.10cm of input] {$x$};

  % A compact pooled grid represents the noised latent presented to the model.
  \begin{scope}[shift={(noisy.south west)}]
    \foreach \gx/\gy/\base in {0/1/poolCyan,1/1/poolTeal,0/0/poolGold,1/0/poolPurple} {
      \pgfmathsetmacro{\px}{\gx*\poolcell}
      \pgfmathsetmacro{\py}{\gy*\poolcell}
      \path[pool token, fill=\base!55!white]
        (\px,\py) rectangle ++(\poolcell,\poolcell);
    }
  \end{scope}
  \node[pool stage label, below=0.10cm of noisy] {$z_t$};

  \node[pool process, fill=white, minimum width=0.78cm, minimum height=0.55cm]
    (noise) at (6.00,2.45) {$t,\ \epsilon$};
  \node[pool process, fill=white, minimum width=0.56cm, minimum height=0.55cm]
    (condition) at (8.40,2.45) {$y$};

  \node[anchor=south east, inner sep=0pt]
    at ([xshift=-0.08cm,yshift=0.08cm]dino.south east)
    {\twemoji[height=0.21cm]{snowflake}};
  \node[anchor=south east, inner sep=0pt]
    at ([xshift=-0.06cm,yshift=0.06cm]pooling.south east)
    {\twemoji[height=0.18cm]{snowflake}};
  \node[anchor=south east, inner sep=0pt]
    at ([xshift=-0.08cm,yshift=0.08cm]generator.south east)
    {\twemoji[height=0.21cm]{fire}};

  \coordinate (input out) at ([xshift=0.12cm]input.east);
  \coordinate (dino in) at ([xshift=-0.12cm]dino.west);
  \coordinate (dino out) at ([xshift=0.12cm]dino.east);
  \coordinate (pooling in) at ([xshift=-0.12cm]pooling.west);
  \coordinate (pooling out) at ([xshift=0.12cm]pooling.east);
  \coordinate (noisy in) at ([xshift=-0.12cm]noisy.west);
  \coordinate (noisy out) at ([xshift=0.12cm]noisy.east);
  \coordinate (generator in) at ([xshift=-0.12cm]generator.west);
  \coordinate (generator out) at ([xshift=0.12cm]generator.east);
  \coordinate (lfm in) at ([xshift=-0.12cm]lfm.west);
  \coordinate (noise out) at ([yshift=-0.12cm]noise.south);
  \coordinate (noisy top) at ([yshift=0.12cm]noisy.north);
  \coordinate (condition out) at ([yshift=-0.12cm]condition.south);
  \coordinate (generator top) at ([yshift=0.12cm]generator.north);
  \draw[pool arrow] (input out) -- (dino in);
  \draw[pool arrow] (dino out) -- (pooling in);
  \draw[pool arrow] (pooling out) -- (noisy in);
  \draw[pool arrow] (noise out) -- (noisy top);
  \draw[pool arrow] (noisy out) -- (generator in);
  \draw[pool arrow] (condition out) -- (generator top);
  \draw[pool arrow] (generator out) -- (lfm in);

  % Auxiliary losses form one column beneath the primary objective.
  \node[pool head, fill=poolGold!7, minimum width=0.88cm]
    (ighead) at (10.15,-0.23) {$H_\theta$};
  \node[pool head, fill=poolPurple!6, minimum width=0.88cm]
    (rechead) at (10.15,-1.31) {$R_\theta$};
  \node[pool loss, minimum width=0.96cm]
    (lig) at (11.80,-0.23) {$\mathcal L_{\mathrm{IG}}$};
  \node[pool loss, minimum width=1.13cm]
    (lrec) at (11.80,-1.31) {$\mathcal L_{\mathrm{rec}}^{E}$};
  \coordinate (hidden top) at ([yshift=-0.18cm]generator.south);
  \coordinate (ig branch) at (hidden top |- ighead.west);
  \coordinate (rec branch) at (hidden top |- rechead.west);

  \draw[draw=black!55, line width=0.65pt]
    (hidden top) -- node[pool operation label, font=\scriptsize, right=0.02cm]
    {$h^{(8)}$} (rec branch);
  \draw[pool auxiliary arrow, shorten >=5pt]
    (ig branch) -- (ighead.west);
  \draw[pool auxiliary arrow, shorten >=5pt]
    (rec branch) -- (rechead.west);
  \coordinate (ighead out) at ([xshift=0.12cm]ighead.east);
  \coordinate (lig in) at ([xshift=-0.12cm]lig.west);
  \coordinate (rechead out) at ([xshift=0.12cm]rechead.east);
  \coordinate (lrec in) at ([xshift=-0.12cm]lrec.west);
  \draw[pool auxiliary arrow] (ighead out) -- (lig in);
  \draw[pool auxiliary arrow] (rechead out) -- (lrec in);

  % The clean pooled latent is the shared target of both flow objectives.
  \coordinate (clean start) at ([xshift=0.22cm]pooling.east);
  \coordinate (clean upper left) at ([yshift=2.02cm]clean start);
  \coordinate (clean bus anchor) at ([xshift=0.68cm]lfm.east);
  \coordinate (clean bus top) at (clean bus anchor |- clean upper left);
  \coordinate (clean bus fm) at (clean bus anchor |- lfm.east);
  \coordinate (clean bus ig) at (clean bus anchor |- lig.east);
  \coordinate (lfm target in) at ([xshift=0.12cm]lfm.east);
  \coordinate (lig target in) at ([xshift=0.12cm]lig.east);
  \draw[pool target line]
    (clean start) -- (clean upper left) -- (clean bus top) -- (clean bus ig);
  \draw[pool target arrow] (clean bus fm) -- (lfm target in);
  \draw[pool target arrow] (clean bus ig) -- (lig target in);

  % The dense encoder field branches from the signal just after DINOv3.
  \coordinate (dense start) at ([xshift=0.22cm]dino.east);
  \coordinate (dense lower left) at ([yshift=-3.15cm]dense start);
  \coordinate (dense lower right) at (lrec.south |- dense lower left);
  \draw[pool target arrow, shorten >=5pt]
    (dense start) -- (dense lower left) -- (dense lower right) -- (lrec.south);
\end{tikzpicture}

%% file: figures/training_duration.tex
\subsection{Effect of training duration}
\label{app:training-duration}

We compare the extended-training checkpoints with the approximately
compute-matched endpoints from Appendix~\ref{app:training-budget}, sweeping
the IG scale at each duration (Tables~\ref{tab:training-duration-two-two}
and~\ref{tab:training-duration-four-four}). All evaluations use 50,000 samples,
100 Euler steps, and IG alone, with the remaining evaluation settings held
fixed across training durations. Increasing training from 180 to
222 epochs leaves the best observed FID nearly unchanged for $2\times2$
pooling (1.0462 versus 1.0476). For $4\times4$, increasing training from 300
to 390 epochs slightly worsens the best observed FID (1.2880 versus 1.3160).
These results indicate diminishing returns under the tested training recipe;
repeated evaluations would be needed to assess statistical significance.

IS generally increases with training duration at fixed guidance scales,
so the FID and IS trends differ. The main generation tables and throughput figures
retain their reported 180/300-epoch operating points (IG scales 2.00 and 2.75,
respectively); the finer sweeps below also evaluate intermediate scales.

\input{figures/training_duration_data.tex}

\begin{table}[H]
\caption{IG sweeps for $2\times2$ learned pooling at 180 and 222 epochs.
Bold marks the lowest observed FID at each duration.}
\label{tab:training-duration-two-two}
\centering
\begin{tabular}{@{}r|rr|rr@{}}
\toprule
& \multicolumn{2}{c|}{180 epochs} & \multicolumn{2}{c}{222 epochs} \\
IG scale & FID $\downarrow$ & IS $\uparrow$ & FID $\downarrow$ & IS $\uparrow$ \\
\midrule
\poolTrainingDurationTwoTwoRows
\bottomrule
\end{tabular}
\end{table}

\begin{table}[H]
\caption{IG sweeps for $4\times4$ learned pooling at 300 and 390 epochs.
Bold marks the lowest observed FID at each duration.}
\label{tab:training-duration-four-four}
\centering
\begin{tabular}{@{}r|rr|rr@{}}
\toprule
& \multicolumn{2}{c|}{300 epochs} & \multicolumn{2}{c}{390 epochs} \\
IG scale & FID $\downarrow$ & IS $\uparrow$ & FID $\downarrow$ & IS $\uparrow$ \\
\midrule
\poolTrainingDurationFourFourRows
\bottomrule
\end{tabular}
\end{table}

%% file: figures/training_duration_data.tex
% User-provided results, 2026-10-01; 390-epoch sweep completed 2026-10-03.
% Shared by the manuscript appendix and research log. Preserve supplied precision.
\newcommand{\poolTrainingDurationTwoTwoRows}{%
1.000 & 2.7805 & 192.37 & 2.7682 & 193.79 \\
1.250 & 1.7069 & 218.79 & 1.7269 & 220.06 \\
1.500 & 1.2237 & 240.43 & 1.2330 & 242.09 \\
1.750 & 1.0603 & 257.87 & 1.0645 & 259.57 \\
1.875 & \textbf{1.0462} & 265.43 & \textbf{1.0476} & 266.27 \\
2.000 & 1.0525 & 272.00 & 1.0542 & 272.29 \\
2.125 & 1.0962 & 277.19 & 1.0810 & 277.62 \\
2.250 & 1.1340 & 280.11 & 1.1161 & 282.42 \\
2.500 & 1.2429 & 287.69 & 1.2058 & 288.80 \\
}
\newcommand{\poolTrainingDurationFourFourRows}{%
1.000 & 7.1942 & 146.13 & 6.7782 & 150.30 \\
1.250 & 4.7407 & 171.29 & 4.4667 & 176.07 \\
1.500 & 3.1969 & 194.26 & 3.0450 & 198.28 \\
1.750 & 2.2363 & 214.17 & 2.1802 & 218.03 \\
2.000 & 1.7026 & 229.57 & 1.6962 & 233.37 \\
2.250 & 1.4350 & 241.78 & 1.4425 & 245.59 \\
2.500 & 1.3068 & 251.22 & 1.3433 & 255.35 \\
2.625 & \textbf{1.2880} & 255.05 & 1.3277 & 258.97 \\
2.750 & 1.2882 & 258.04 & 1.3178 & 262.48 \\
2.770 & 1.2902 & 258.74 & 1.3173 & 262.61 \\
2.875 & 1.3018 & 260.34 & \textbf{1.3160} & 264.73 \\
3.000 & 1.3193 & 262.32 & 1.3362 & 266.73 \\
3.125 & 1.3605 & 263.72 & 1.3572 & 267.37 \\
3.250 & 1.4002 & 265.53 & 1.3924 & 269.10 \\
3.500 & 1.4867 & 266.76 & 1.4531 & 268.97 \\
}

%% file: figures/quality_throughput_guidance.tex
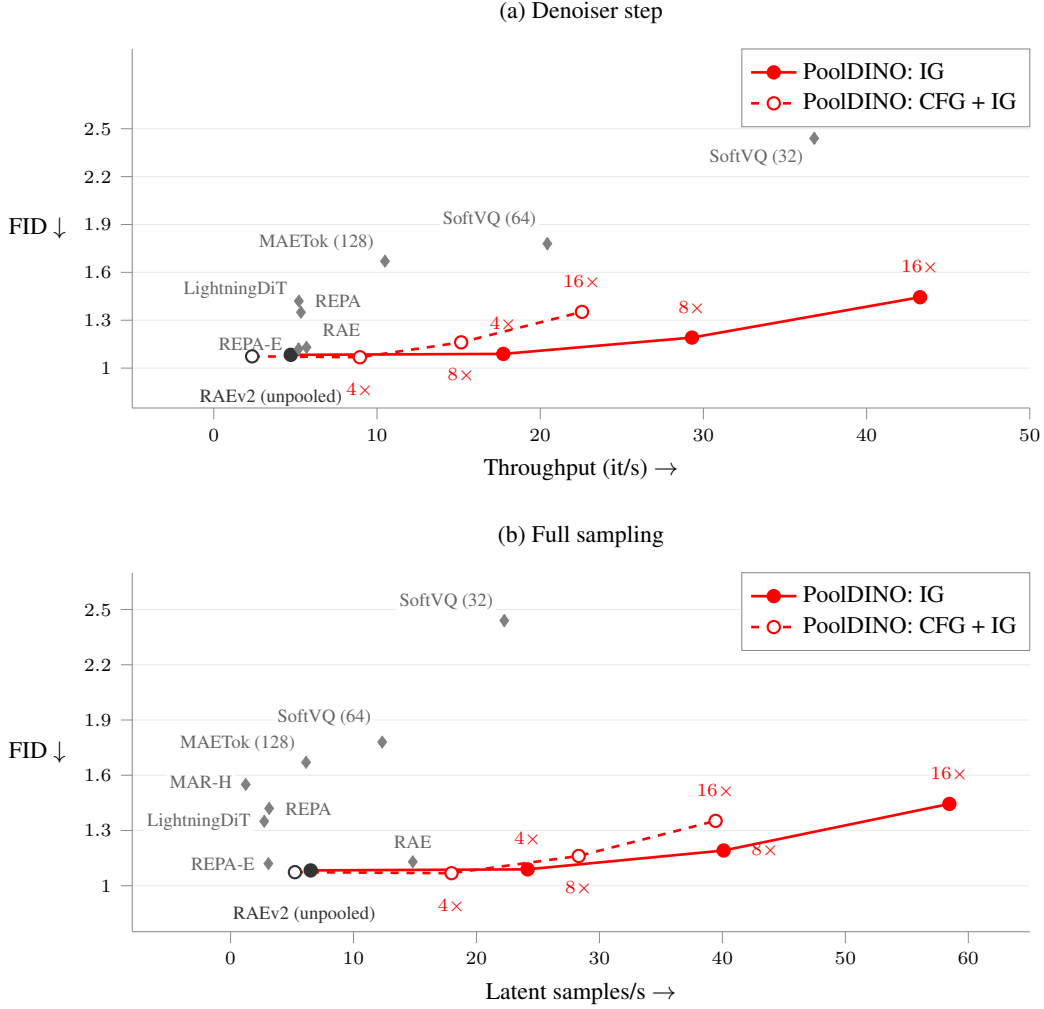
\begin{figure}[t]
\centering
\begingroup
\poolShowCFGIGtrue
\input{figures/quality_throughput_plot.tex}
\endgroup
\caption{Quality--throughput comparison with both guidance settings on an H100 NVL at batch size 128. Our 80-epoch models use IG alone (solid, filled) or CFG+IG (dashed, hollow), with FID evaluated at 100 Euler steps. Point labels give token compression; black circles denote the unpooled RAEv2 reference. Literature baselines and timing conventions match Figure~\ref{fig:quality-throughput}; full-sampling throughput excludes image decoding.}
\label{fig:quality-throughput-guidance}
\end{figure}

%% file: figures/quality_throughput_plot.tex
% H100 NVL, batch 128. Measurements and error bars are unchanged.
% Provenance: figures/quality_throughput_measurements.json.
% Match Figure 1: direct labels, gray baselines, red 80-epoch models,
% circular markers, horizontal grid, and open top/right borders.
\begingroup
\begin{tikzpicture}
\begin{groupplot}[
    group style={group size=1 by 2, vertical sep=62pt},
    scale only axis,
    width=0.85\textwidth, height=0.34\textwidth,
    ymin=0.75, ymax=2.70, ytick={1.0,1.3,1.6,1.9,2.2,2.5},
    ylabel={FID $\downarrow$}, ylabel style={rotate=-90},
    scaled ticks=false,
    tick label style={font=\scriptsize},
    label style={font=\footnotesize}, title style={font=\footnotesize},
    axis x line*=bottom, axis y line*=left,
    axis line style={black!45}, tick style={black!45},
    tick align=outside, xtick pos=bottom, ytick pos=left,
    xmajorgrids=false, ymajorgrids=true, grid style={black!8},
    legend style={at={(1,1)},anchor=north east,outer sep=0pt,
        draw=black!45,line width=0.4pt,fill=white,
        inner sep=3pt,font=\footnotesize,cells={anchor=west}},
    clip=false,
]
\nextgroupplot[title={(a) Denoiser step}, xlabel={Throughput (it/s) $\rightarrow$},
    ymax=3.00, xmin=-5, xmax=50, xtick={0,10,20,30,40,50}]
\addlegendimage{poolRed,solid,line width=1pt,mark=*,mark size=2.3pt,
    mark options={solid,fill=poolRed,line width=0.8pt}}
\addlegendentry{PoolDINO: IG}
\ifpoolShowCFGIG
\addlegendimage{poolRed,dashed,line width=1pt,mark=*,mark size=2.3pt,
    mark options={solid,fill=white,line width=0.8pt}}
\addlegendentry{PoolDINO: CFG + IG}
\fi
\addplot[poolRed,solid,line width=1pt,forget plot] coordinates {
(4.712748488400175,1.083134)
(17.73794739341444,1.088975)
(29.2985491526518,1.19126)
(43.2746789027505,1.444216)
};
\addplot[only marks,black!80,mark=*,mark size=2.3pt,
    mark options={solid,fill=black!80,line width=0.8pt},
    forget plot,error bars/.cd,x dir=both,x explicit]
    coordinates {(4.712748488400175,1.083134) +- (0.013762746665959868,0)};
\addplot[only marks,poolRed,mark=*,mark size=2.3pt,
    mark options={solid,fill=poolRed,line width=0.8pt},
    forget plot,error bars/.cd,x dir=both,x explicit]
    coordinates {(17.73794739341444,1.088975) +- (0.05252443308956509,0)};
\node[anchor=south,font=\scriptsize,text=poolRed,yshift=5pt]
    at (axis cs:17.73794739341444,1.088975) {$4\times$};
\addplot[only marks,poolRed,mark=*,mark size=2.3pt,
    mark options={solid,fill=poolRed,line width=0.8pt},
    forget plot,error bars/.cd,x dir=both,x explicit]
    coordinates {(29.2985491526518,1.19126) +- (0.08885486409816346,0)};
\node[anchor=south,font=\scriptsize,text=poolRed,yshift=5pt]
    at (axis cs:29.2985491526518,1.19126) {$8\times$};
\addplot[only marks,poolRed,mark=*,mark size=2.3pt,
    mark options={solid,fill=poolRed,line width=0.8pt},
    forget plot,error bars/.cd,x dir=both,x explicit]
    coordinates {(43.2746789027505,1.444216) +- (0.08173328831628239,0)};
\node[anchor=south,font=\scriptsize,text=poolRed,yshift=5pt]
    at (axis cs:43.2746789027505,1.444216) {$16\times$};
\ifpoolShowCFGIG
\addplot[poolRed,dashed,line width=1pt,forget plot] coordinates {
(2.347490032462688,1.07359)
(8.952391322592309,1.068004)
(15.1566196798851,1.1615)
(22.558194323123068,1.352229)
};
\addplot[only marks,black!80,mark=*,mark size=2.3pt,
    mark options={solid,fill=white,line width=0.8pt},
    forget plot,error bars/.cd,x dir=both,x explicit]
    coordinates {(2.347490032462688,1.07359) +- (0.007203360889996675,0)};
\addplot[only marks,poolRed,mark=*,mark size=2.3pt,
    mark options={solid,fill=white,line width=0.8pt},
    forget plot,error bars/.cd,x dir=both,x explicit]
    coordinates {(8.952391322592309,1.068004) +- (0.028280913279967717,0)};
\node[anchor=north,font=\scriptsize,text=poolRed,yshift=-6pt]
    at (axis cs:8.952391322592309,1.068004) {$4\times$};
\addplot[only marks,poolRed,mark=*,mark size=2.3pt,
    mark options={solid,fill=white,line width=0.8pt},
    forget plot,error bars/.cd,x dir=both,x explicit]
    coordinates {(15.1566196798851,1.1615) +- (0.03675040487760712,0)};
\node[anchor=north,font=\scriptsize,text=poolRed,yshift=-6pt]
    at (axis cs:15.1566196798851,1.1615) {$8\times$};
\addplot[only marks,poolRed,mark=*,mark size=2.3pt,
    mark options={solid,fill=white,line width=0.8pt},
    forget plot,error bars/.cd,x dir=both,x explicit]
    coordinates {(22.558194323123068,1.352229) +- (0.06876688433618959,0)};
\node[anchor=south,font=\scriptsize,text=poolRed,yshift=5pt]
    at (axis cs:22.558194323123068,1.352229) {$16\times$};
\fi
\node[anchor=north,font=\scriptsize,text=black!80,yshift=-7pt]
    at (axis cs:3.5,1.04) {RAEv2 (unpooled)};
\addplot[only marks,black!50,mark=diamond*,mark size=2.2pt,
    forget plot,error bars/.cd,x dir=both,x explicit] coordinates {
(10.488728111996585,1.67) +- (0.024476392770924043,0)
(20.436129570034304,1.78) +- (0.05480720970453702,0)
(36.79115463287783,2.44) +- (0.04432809559762806,0)
(5.211216904973409,1.42) +- (0.02265814579078974,0)
(5.680593954077519,1.13) +- (0.022429045098401507,0)
(5.34150010029639,1.35) +- (0.016389764518946876,0)
(5.19487677095369,1.12) +- (0.05364609174650594,0)
};
\begin{scope}[every node/.style={font=\scriptsize,
    text=black!60,fill=white,inner sep=1pt}]
\node[anchor=south east,xshift=-3pt,yshift=3pt] at (axis cs:10.488728111996585,1.67) {MAETok (128)};
\node[anchor=south east,xshift=-3pt,yshift=5pt] at (axis cs:20.436129570034304,1.78) {SoftVQ (64)};
\node[anchor=north east,xshift=-3pt,yshift=-3pt] at (axis cs:36.79115463287783,2.44) {SoftVQ (32)};
\node[anchor=west,xshift=5pt] at (axis cs:5.211216904973409,1.42) {REPA};
\node[anchor=south west,xshift=5pt,yshift=3pt] at (axis cs:5.680593954077519,1.13) {RAE};
\node[anchor=south east,xshift=-4pt,yshift=4pt] at (axis cs:5.34150010029639,1.35) {LightningDiT};
\node[anchor=east,xshift=-5pt,yshift=2pt] at (axis cs:5.19487677095369,1.12) {REPA-E};
\end{scope}
\nextgroupplot[title={(b) Full sampling}, xlabel={Latent samples/s $\rightarrow$},
    xmin=-8, xmax=65, xtick={0,10,20,30,40,50,60}]
\addlegendimage{poolRed,solid,line width=1pt,mark=*,mark size=2.3pt,
    mark options={solid,fill=poolRed,line width=0.8pt}}
\addlegendentry{PoolDINO: IG}
\ifpoolShowCFGIG
\addlegendimage{poolRed,dashed,line width=1pt,mark=*,mark size=2.3pt,
    mark options={solid,fill=white,line width=0.8pt}}
\addlegendentry{PoolDINO: CFG + IG}
\fi
\addplot[poolRed,solid,line width=1pt,forget plot] coordinates {
(6.5162377754015415,1.083134)
(24.15883440694129,1.088975)
(40.10041655873189,1.19126)
(58.457215615361896,1.444216)
};
\addplot[only marks,black!80,mark=*,mark size=2.3pt,
    mark options={solid,fill=black!80,line width=0.8pt},
    forget plot,error bars/.cd,x dir=both,x explicit]
    coordinates {(6.5162377754015415,1.083134) +- (0.008849058916579925,0)};
\addplot[only marks,poolRed,mark=*,mark size=2.3pt,
    mark options={solid,fill=poolRed,line width=0.8pt},
    forget plot,error bars/.cd,x dir=both,x explicit]
    coordinates {(24.15883440694129,1.088975) +- (0.05262807260879476,0)};
\node[anchor=south,font=\scriptsize,text=poolRed,yshift=5pt]
    at (axis cs:24.15883440694129,1.088975) {$4\times$};
\addplot[only marks,poolRed,mark=*,mark size=2.3pt,
    mark options={solid,fill=poolRed,line width=0.8pt},
    forget plot,error bars/.cd,x dir=both,x explicit]
    coordinates {(40.10041655873189,1.19126) +- (0.0720525749532469,0)};
\node[anchor=west,font=\scriptsize,text=poolRed,xshift=7pt]
    at (axis cs:40.10041655873189,1.19126) {$8\times$};
\addplot[only marks,poolRed,mark=*,mark size=2.3pt,
    mark options={solid,fill=poolRed,line width=0.8pt},
    forget plot,error bars/.cd,x dir=both,x explicit]
    coordinates {(58.457215615361896,1.444216) +- (0.19728694433753302,0)};
\node[anchor=south,font=\scriptsize,text=poolRed,yshift=5pt]
    at (axis cs:58.457215615361896,1.444216) {$16\times$};
\ifpoolShowCFGIG
\addplot[poolRed,dashed,line width=1pt,forget plot] coordinates {
(5.235062367213122,1.07359)
(17.968399807441802,1.068004)
(28.31324140504651,1.1615)
(39.44474204620997,1.352229)
};
\addplot[only marks,black!80,mark=*,mark size=2.3pt,
    mark options={solid,fill=white,line width=0.8pt},
    forget plot,error bars/.cd,x dir=both,x explicit]
    coordinates {(5.235062367213122,1.07359) +- (0.005367735346055888,0)};
\addplot[only marks,poolRed,mark=*,mark size=2.3pt,
    mark options={solid,fill=white,line width=0.8pt},
    forget plot,error bars/.cd,x dir=both,x explicit]
    coordinates {(17.968399807441802,1.068004) +- (0.042631259407653846,0)};
\node[anchor=north,font=\scriptsize,text=poolRed,yshift=-6pt]
    at (axis cs:17.968399807441802,1.068004) {$4\times$};
\addplot[only marks,poolRed,mark=*,mark size=2.3pt,
    mark options={solid,fill=white,line width=0.8pt},
    forget plot,error bars/.cd,x dir=both,x explicit]
    coordinates {(28.31324140504651,1.1615) +- (0.0491575526730079,0)};
\node[anchor=north,font=\scriptsize,text=poolRed,yshift=-6pt]
    at (axis cs:28.31324140504651,1.1615) {$8\times$};
\addplot[only marks,poolRed,mark=*,mark size=2.3pt,
    mark options={solid,fill=white,line width=0.8pt},
    forget plot,error bars/.cd,x dir=both,x explicit]
    coordinates {(39.44474204620997,1.352229) +- (0.11118511274130775,0)};
\node[anchor=south,font=\scriptsize,text=poolRed,yshift=5pt]
    at (axis cs:39.44474204620997,1.352229) {$16\times$};
\fi
\node[anchor=north,font=\scriptsize,text=black!80,yshift=-7pt]
    at (axis cs:6,1.04) {RAEv2 (unpooled)};
\addplot[only marks,black!50,mark=diamond*,mark size=2.2pt,
    forget plot,error bars/.cd,x dir=both,x explicit] coordinates {
(6.140202012224861,1.67) +- (0.0022831176485612306,0)
(12.333923401332111,1.78) +- (0.019504329331953216,0)
(22.25987542683997,2.44) +- (0.07407767770274822,0)
(3.1383672813042254,1.42) +- (0.000657529677442022,0)
(1.2310240113665198,1.55) +- (0.004902102944190099,0)
(14.820128554110196,1.13) +- (0.017648222970774795,0)
(2.7346668802603142,1.35) +- (0.0008613857960049152,0)
(3.0745529762099024,1.12) +- (0.0008952060899096071,0)
};
\begin{scope}[every node/.style={font=\scriptsize,
    text=black!60,fill=white,inner sep=1pt}]
\node[anchor=south east,xshift=-3pt,yshift=3pt] at (axis cs:6.140202012224861,1.67) {MAETok (128)};
\node[anchor=south east,xshift=-3pt,yshift=5pt] at (axis cs:12.333923401332111,1.78) {SoftVQ (64)};
\node[anchor=south east,xshift=-3pt,yshift=3pt] at (axis cs:22.25987542683997,2.44) {SoftVQ (32)};
\node[anchor=west,xshift=5pt] at (axis cs:3.1383672813042254,1.42) {REPA};
\node[anchor=east,xshift=-4pt,yshift=1pt] at (axis cs:1.2310240113665198,1.55) {MAR-H};
\node[anchor=south,yshift=4pt] at (axis cs:14.820128554110196,1.13) {RAE};
\node[anchor=east,xshift=-4pt] at (axis cs:2.7346668802603142,1.35) {LightningDiT};
\node[anchor=east,xshift=-4pt] at (axis cs:3.0745529762099024,1.12) {REPA-E};
\end{scope}
\end{groupplot}
\end{tikzpicture}
\endgroup

%% file: figures/inception_throughput.tex
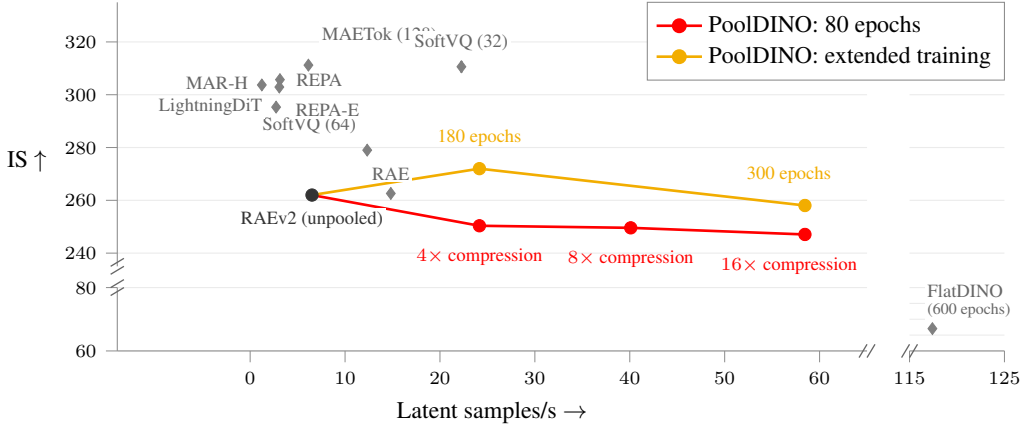
\begin{figure}[!htbp]
\centering
\input{figures/inception_throughput_plot.tex}
\caption{Inception Score versus latent-sampling throughput on an H100 NVL at batch size 128. PoolDINO and the unpooled RAEv2 reference use IG alone at 100 Euler steps, including the extended-training checkpoints. IS is taken at the FID-selected guidance settings, not maximized separately.}
\label{fig:inception-throughput}
\end{figure}

%% file: figures/inception_throughput_plot.tex
% Full-sampling measurements use corrected pooling-specific time shifts.
% Provenance: figures/quality_throughput_measurements.json.
% IS sources and matching FIDs: figures/inception_throughput_sources.json.
% Same checkpoints and FID-selected guidance settings as Figure 1; not best-IS settings.
\begingroup
\begin{tikzpicture}

% -------- TOP MAIN --------
\begin{axis}[
    scale only axis,
    name=throughputMainTop,
    width=0.71\textwidth, height=0.25\textwidth,
    xmin=-14, xmax=65, xticklabels=\empty,
    ymin=235, ymax=335, ytick={240,260,280,300,320},
    ylabel={IS $\uparrow$},
    scaled ticks=false,
    tick label style={font=\scriptsize},
    label style={font=\footnotesize},
    ylabel style={rotate=-90, yshift=-10pt},
    axis x line*=top, axis x line=none, axis y line*=left,
    axis line style={black!45}, tick style={black!45},
    tick align=outside, ytick pos=left,
    xmajorgrids=false, ymajorgrids=true, grid style={black!8},
    clip=false,
]

\addplot[poolRed,line width=1pt,forget plot] coordinates {
    (6.5162377754015415,261.997809)
    (24.15883440694129,250.362663)
    (40.10041655873189,249.577541)
    (58.457215615361896,247.073224)
};
\addplot[poolGold,line width=1pt,forget plot] coordinates {
    (6.5162377754015415,261.997809)
    (24.15883440694129,271.9952880859375)
    (58.457215615361896,258.04)
};
\addplot[only marks,black!80,mark=*,mark size=2.3pt,
    forget plot,error bars/.cd,x dir=both,x explicit]
    coordinates {(6.5162377754015415,261.997809) +- (0.008849058916579925,0)};
\node[anchor=north,font=\scriptsize,text=black!80,
    fill=white,inner sep=1pt,yshift=-5pt]
    at (axis cs:6.5162377754015415,261.997809) {RAEv2 (unpooled)};

\addplot[only marks,poolRed,mark=*,mark size=2.3pt,
    forget plot,error bars/.cd,x dir=both,x explicit]
    coordinates {(24.15883440694129,250.362663) +- (0.05262807260879476,0)};
\node[anchor=north,font=\scriptsize,text=poolRed,yshift=-5pt]
    at (axis cs:24.15883440694129,250.362663) {$4\times$ compression};
\addplot[only marks,poolRed,mark=*,mark size=2.3pt,
    forget plot,error bars/.cd,x dir=both,x explicit]
    coordinates {(40.10041655873189,249.577541) +- (0.0720525749532469,0)};
\node[anchor=north,font=\scriptsize,text=poolRed,yshift=-5pt]
    at (axis cs:40.10041655873189,249.577541) {$8\times$ compression};
\addplot[only marks,poolRed,mark=*,mark size=2.3pt,
    forget plot,error bars/.cd,x dir=both,x explicit]
    coordinates {(58.457215615361896,247.073224) +- (0.19728694433753302,0)};
\node[anchor=north,font=\scriptsize,text=poolRed,xshift=-6pt,yshift=-5pt]
    at (axis cs:58.457215615361896,247.073224) {$16\times$ compression};

\addplot[only marks,poolGold,mark=*,mark size=2.3pt,
    forget plot,error bars/.cd,x dir=both,x explicit] coordinates {
    (24.15883440694129,271.9952880859375) +- (0.05262807260879476,0)
    (58.457215615361896,258.04) +- (0.19728694433753302,0)
};
\node[anchor=south,font=\scriptsize,text=poolGold,yshift=5pt]
    at (axis cs:24.15883440694129,271.9952880859375) {180 epochs};
\node[anchor=south,font=\scriptsize,text=poolGold,xshift=-6pt,yshift=5pt]
    at (axis cs:58.457215615361896,258.04) {300 epochs};

\addplot[only marks,black!50,mark=diamond*,mark size=2.2pt,
    forget plot,error bars/.cd,x dir=both,x explicit] coordinates {
    (6.140202012224861,311.2) +- (0.0022831176485612306,0)
    (12.333923401332111,279.0) +- (0.019504329331953216,0)
    (22.25987542683997,310.6) +- (0.07407767770274822,0)
    (3.1383672813042254,305.7) +- (0.000657529677442022,0)
    (1.2310240113665198,303.7) +- (0.004902102944190099,0)
    (14.820128554110196,262.6) +- (0.017648222970774795,0)
    (2.7346668802603142,295.3) +- (0.0008613857960049152,0)
    (3.0745529762099024,302.9) +- (0.0008952060899096071,0)
};
\begin{scope}[every node/.style={font=\scriptsize,
    text=black!60,fill=white,inner sep=1pt}]
\node[anchor=south west,xshift=4pt,yshift=7pt]
    at (axis cs:6.140202012224861,311.2) {MAETok (128)};
\node[anchor=south east,xshift=-3pt,yshift=5pt]
    at (axis cs:12.333923401332111,279.0) {SoftVQ (64)};
\node[anchor=south,yshift=5pt]
    at (axis cs:22.25987542683997,310.6) {SoftVQ (32)};
\node[anchor=west,xshift=5pt]
    at (axis cs:3.1383672813042254,305.7) {REPA};
\node[anchor=east,xshift=-4pt,yshift=1pt]
    at (axis cs:1.2310240113665198,303.7) {MAR-H};
\node[anchor=south,yshift=4pt]
    at (axis cs:14.820128554110196,262.6) {RAE};
\node[anchor=east,xshift=-4pt]
    at (axis cs:2.7346668802603142,295.3) {LightningDiT};
\node[anchor=north west,xshift=5pt,yshift=-5pt]
    at (axis cs:3.0745529762099024,302.9) {REPA-E};
\end{scope}
\end{axis}

% -------- BOTTOM MAIN --------
\begin{axis}[
    scale only axis,
    name=throughputMainBottom,
    at={(throughputMainTop.south west)}, anchor=north west, yshift=-8pt,
    width=0.71\textwidth, height=0.06\textwidth,
    xmin=-14, xmax=65, xtick={0,10,20,30,40,50,60},
    ymin=60, ymax=80, ytick={60, 80},
    xlabel={Latent samples/s $\rightarrow$},
    scaled ticks=false,
    tick label style={font=\scriptsize},
    label style={font=\footnotesize},
    axis x line*=bottom, axis y line*=left,
    axis line style={black!45}, tick style={black!45},
    tick align=outside, xtick pos=bottom, ytick pos=left,
    xmajorgrids=false, ymajorgrids=true, grid style={black!8},
    clip=false,
]
\end{axis}

% -------- TOP TAIL --------
\begin{axis}[
    name=throughputTailTop,
    at={(throughputMainTop.south east)}, anchor=south west, xshift=16pt,
    scale only axis,
    width={0.71\textwidth*10/79}, height=0.25\textwidth,
    xmin=115, xmax=125, xticklabels=\empty,
    ymin=235, ymax=335, yticklabels=\empty,
    scaled ticks=false, tick label style={font=\scriptsize},
    axis x line=none, axis y line*=left,
    y axis line style={draw=none}, ytick style={draw=none},
    axis line style={black!45}, tick style={black!45},
    tick align=outside,
    xmajorgrids=false, ymajorgrids=true, grid style={black!8},
    legend style={at={(1,1)},anchor=north east,outer sep=0pt,
        draw=black!45,line width=0.4pt,fill=white,
        inner sep=3pt,font=\footnotesize,cells={anchor=west}},
    clip=false,
]
\addlegendimage{poolRed,line width=1pt,mark=*,mark size=2.3pt}
\addlegendentry{PoolDINO: 80 epochs}
\addlegendimage{poolGold,line width=1pt,mark=*,mark size=2.3pt}
\addlegendentry{PoolDINO: extended training}
\end{axis}

% -------- BOTTOM TAIL --------
\begin{axis}[
    name=throughputTailBottom,
    at={(throughputMainBottom.south east)}, anchor=south west, xshift=16pt,
    scale only axis,
    width={0.71\textwidth*10/79}, height=0.06\textwidth,
    xmin=115, xmax=125, xtick={115,125},
    ymin=60, ymax=80, yticklabels=\empty,
    scaled ticks=false, tick label style={font=\scriptsize},
    axis x line*=bottom, axis y line*=left,
    y axis line style={draw=none}, ytick style={draw=none},
    axis line style={black!45}, tick style={black!45},
    tick align=outside, xtick pos=bottom,
    xmajorgrids=false, ymajorgrids=true, grid style={black!8},
    clip=false,
]
\addplot[only marks,black!50,mark=diamond*,mark size=2.2pt,
    forget plot]
    coordinates {(117.41,67.05)};
\node[anchor=south west,font=\scriptsize,text=black!60,
    fill=white,inner sep=1pt,xshift=-3pt,yshift=3pt,align=left]
    at (axis cs:117.41,67.05) {FlatDINO \\[-0.5ex] \tiny (600 epochs)};
\end{axis}

% Explicit break marks
% X-axis break
\foreach \edge in {throughputMainBottom.south east,throughputTailBottom.south west} {
    \draw[black!60,line width=0.5pt]
        ([xshift=-3pt,yshift=-2.5pt]\edge) -- ([xshift=0pt,yshift=2.5pt]\edge)
        ([xshift=0pt,yshift=-2.5pt]\edge) -- ([xshift=3pt,yshift=2.5pt]\edge);
}
% Y-axis break (Main)
\foreach \edge in {throughputMainTop.south west,throughputMainBottom.north west} {
    \draw[black!60,line width=0.5pt]
        ([yshift=-3pt,xshift=-2.5pt]\edge) -- ([yshift=0pt,xshift=2.5pt]\edge)
        ([yshift=0pt,xshift=-2.5pt]\edge) -- ([yshift=3pt,xshift=2.5pt]\edge);
}
\end{tikzpicture}
\endgroup

%% file: figures/sampling_steps_results.tex
\subsection{Sampling-step ablation}
\label{app:sampling-steps}

We compare 50 and 100 Euler steps while keeping each checkpoint's selected
IG scale fixed. Across all evaluated pooling geometries and training durations,
halving the sampling steps doubles latent-sampling throughput with little
change in generation quality (Table~\ref{tab:sampling-step-comparison}).
The absolute FID difference remains below 0.04 in every configuration.

\begin{table}[!htbp]
\caption{IG-only generation with 100 versus 50 Euler steps. Throughput is latent samples/s on an H100 NVL at batch size 128; $\pm$ reports timing standard deviation.}
\label{tab:sampling-step-comparison}
\centering
\small
\setlength{\tabcolsep}{5pt}
\begin{tabular}{@{}lrr|rr|rr@{}}
\toprule
& & & \multicolumn{2}{c|}{FID $\downarrow$} & \multicolumn{2}{c}{Latent samples/s $\uparrow$} \\
\cmidrule(lr){4-5}\cmidrule(lr){6-7}
Pooling & Epochs & $s_{\mathrm{IG}}$ & 100 steps & 50 steps & 100 steps & 50 steps \\
\midrule
$2\times2$ & 80  & 1.75 & 1.0890 & 1.1164 & $24.16\pm0.05$ & $48.44\pm0.17$ \\
$2\times2$ & 180 & 2.00 & 1.0525 & 1.0629 & $24.16\pm0.05$ & $48.44\pm0.17$ \\
\midrule
$2\times4$ & 80  & 2.00 & 1.1913 & 1.2283 & $40.10\pm0.07$ & $80.42\pm0.33$ \\
\midrule
$4\times4$ & 80  & 2.75 & 1.4442 & 1.4429 & $58.46\pm0.20$ & $117.13\pm0.38$ \\
$4\times4$ & 300 & 2.75 & 1.2882 & 1.2981 & $58.46\pm0.20$ & $117.13\pm0.38$ \\
\bottomrule
\end{tabular}
\end{table}

For the 300-epoch $4\times4$ generator, we further vary the step count
from 10 to 100
(Figure~\ref{fig:sampling-steps}).
Figure~\ref{fig:quality-throughput-50steps}
shows all five measured 50-step PoolDINO configurations alongside the
literature baselines, including FlatDINO. Figure~\ref{fig:quality-throughput}
and the main generation tables retain 100 steps for our models to isolate
the effect of training duration.

\input{figures/quality_throughput_50steps.tex}

\input{figures/sampling_steps.tex}

%% file: figures/quality_throughput_50steps.tex
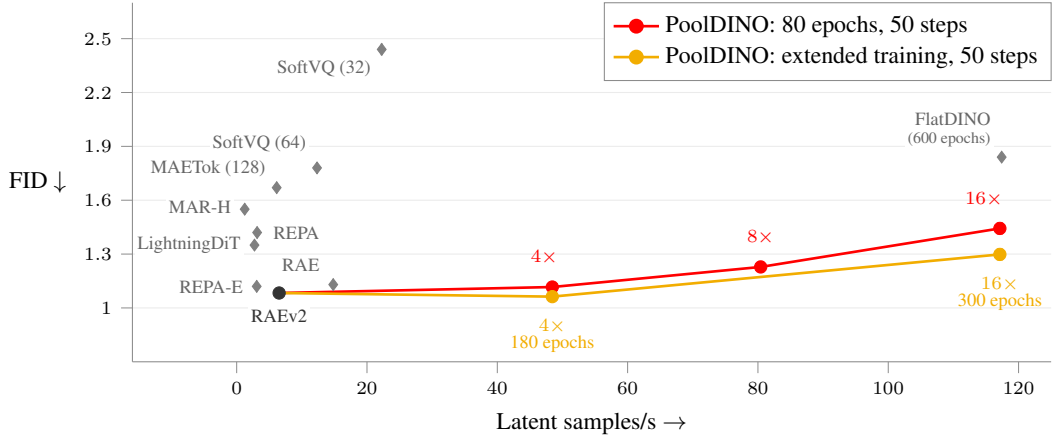
\begin{figure}[!htbp]
\centering
\input{figures/quality_throughput_50steps_plot.tex}
\caption{Generation quality versus latent-sampling throughput with 50-step sampling for PoolDINO, using IG alone on an H100 NVL at batch size 128. Red points show the 80-epoch models; gold points show extended training. The RAEv2 reference uses IG with 100 steps. Literature points, including FlatDINO, retain their published FIDs and measured throughput under their respective sampling protocols.}
\label{fig:quality-throughput-50steps}
\end{figure}

%% file: figures/quality_throughput_50steps_plot.tex
% Full-sampling measurements use corrected pooling-specific time shifts.
% Provenance: figures/quality_throughput_measurements.json.
% All five 50-step PoolDINO points: logs/FIFTY_STEP_GENERATION_RESULTS.tex.
% PoolDINO uses IG alone. Literature FIDs are published annotations.
\begingroup
\begin{tikzpicture}
\begin{axis}[
    scale only axis,
    width=0.87\textwidth, height=0.34\textwidth,
    % Left annotation padding keeps baseline labels inside the plotting area.
    xmin=-16, xmax=125, xtick={0,20,40,60,80,100,120},
    ymin=0.70, ymax=2.70, ytick={1.0,1.3,1.6,1.9,2.2,2.5},
    xlabel={Latent samples/s $\rightarrow$}, ylabel={FID $\downarrow$},
    scaled ticks=false,
    tick label style={font=\scriptsize},
    label style={font=\footnotesize},
    ylabel style={rotate=-90},
    axis x line*=bottom, axis y line*=left,
    axis line style={black!45}, tick style={black!45},
    tick align=outside, xtick pos=bottom, ytick pos=left,
    xmajorgrids=false, ymajorgrids=true, grid style={black!8},
    legend style={at={(1,1)},anchor=north east,outer sep=0pt,
        draw=black!45,line width=0.4pt,fill=white,
        inner sep=3pt,font=\footnotesize,cells={anchor=west}},
    clip=false,
]
\addlegendimage{poolRed,line width=1pt,mark=*,mark size=2.3pt}
\addlegendentry{PoolDINO: 80 epochs, 50 steps}
\addlegendimage{poolGold,line width=1pt,mark=*,mark size=2.3pt}
\addlegendentry{PoolDINO: extended training, 50 steps}

% A guide connects the discrete pooling variants; no interpolated frontier.
\addplot[poolRed,line width=1pt,forget plot] coordinates {
    (6.5162377754015415,1.083134)
    (48.44,1.1164)
    (80.42,1.2283)
    (117.13,1.4429)
};
% PoolDINO uses measured 50-step results; RAEv2 is a separate 100-step reference.
\addplot[poolGold,line width=1pt,forget plot] coordinates {
    (6.5162377754015415,1.083134)
    (48.44,1.0629)
    (117.13,1.2981)
};
\addplot[only marks,poolRed,mark=*,mark size=2.3pt,
    forget plot,error bars/.cd,x dir=both,x explicit]
    coordinates {(48.44,1.1164) +- (0.17,0)};
\node[anchor=south,font=\scriptsize,text=poolRed,xshift=-3pt,yshift=5pt]
    at (axis cs:48.44,1.1164) {$4\times$};
\addplot[only marks,poolRed,mark=*,mark size=2.3pt,
    forget plot,error bars/.cd,x dir=both,x explicit]
    coordinates {(80.42,1.2283) +- (0.33,0)};
\node[anchor=south,font=\scriptsize,text=poolRed,yshift=5pt]
    at (axis cs:80.42,1.2283) {$8\times$};
\addplot[only marks,poolRed,mark=*,mark size=2.3pt,
    forget plot,error bars/.cd,x dir=both,x explicit]
    coordinates {(117.13,1.4429) +- (0.38,0)};
\node[anchor=south,font=\scriptsize,text=poolRed,xshift=-6pt,yshift=5pt]
    at (axis cs:117.13,1.4429) {$16\times$};

% New 50-step timings are measured per architecture with random weights.
% They are reused across training epochs, not extrapolated from 100 steps.
\addplot[only marks,poolGold,mark=*,mark size=2.3pt,
    forget plot,error bars/.cd,x dir=both,x explicit] coordinates {
    (48.44,1.0629) +- (0.17,0)
};
\node[anchor=north,font=\scriptsize,text=poolGold,yshift=-5pt,align=center]
    at (axis cs:48.44,1.0629) {$4\times$\\[-0.5ex]180 epochs};

\input{figures/quality_throughput_literature_points.tex}
\addplot[only marks,black!80,mark=*,mark size=2.3pt,
    forget plot,error bars/.cd,x dir=both,x explicit]
    coordinates {(6.5162377754015415,1.083134) +- (0.008849058916579925,0)};
\node[anchor=north,font=\scriptsize,text=black!80,
    fill=white,inner sep=1pt,yshift=-5pt]
    at (axis cs:6.5162377754015415,1.083134) {RAEv2};
\addplot[only marks,poolGold,mark=*,mark size=2.3pt,
    forget plot,error bars/.cd,x dir=both,x explicit]
    coordinates {(117.13,1.2981) +- (0.38,0)};
\node[anchor=north,font=\scriptsize,text=poolGold,
    yshift=-5pt,align=center]
    at (axis cs:117.13,1.2981) {$16\times$\\[-0.5ex]300 epochs};
% FlatDINO: user-supplied throughput and reported paper FID (preview).
% Confirm the matched sampling protocol before final submission.
% Throughput uncertainty is +/- 0.33 samples/s; omitted visually for clarity.
\addplot[only marks,black!50,mark=diamond*,mark size=2.2pt,
    forget plot]
    coordinates {(117.41,1.84)};
\node[anchor=south east,font=\scriptsize,text=black!60,
    fill=white,inner sep=1pt,xshift=-3pt,yshift=3pt,align=right]
    at (axis cs:117.41,1.84) {FlatDINO \\[-0.5ex] \tiny (600 epochs)};
\end{axis}
\end{tikzpicture}
\endgroup

%% file: figures/sampling_steps.tex
\begin{figure}[!htbp]
\centering
\begingroup
\input{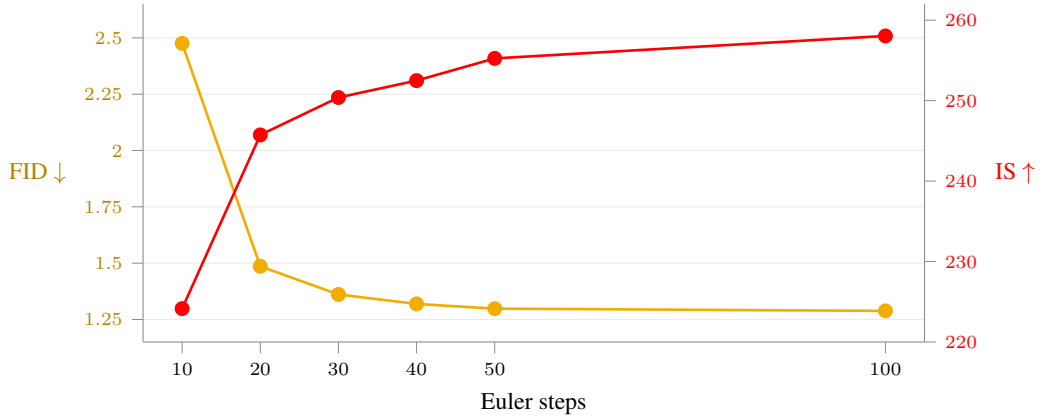}
\begin{tikzpicture}
\begin{axis}[
    name=samplingFID,
    scale only axis,
    width=0.74\textwidth, height=0.32\textwidth,
    xmin=5, xmax=105, xtick={10,20,30,40,50,100},
    ymin=1.15, ymax=2.65, ytick={1.25,1.5,1.75,2,2.25,2.5},
    xlabel={Euler steps}, ylabel={FID $\downarrow$},
    scaled ticks=false, tick label style={font=\scriptsize},
    label style={font=\footnotesize},
    ylabel style={rotate=-90,text=poolGold!75!black},
    yticklabel style={text=poolGold!75!black},
    axis x line*=bottom, axis y line*=left,
    axis line style={black!45}, tick style={black!45},
    tick align=outside, xtick pos=bottom, ytick pos=left,
    xmajorgrids=false, ymajorgrids=true, grid style={black!8},
]
\addplot[poolGold,line width=1pt,mark=*,mark size=2.3pt]
    table[x=steps,y=fid] {\poolSamplingStepsData};
\end{axis}
\begin{axis}[
    at={(samplingFID.south west)}, anchor=south west,
    scale only axis,
    width=0.74\textwidth, height=0.32\textwidth,
    xmin=5, xmax=105, xtick=\empty,
    ymin=220, ymax=262, ytick={220,230,240,250,260},
    ylabel={IS $\uparrow$},
    scaled ticks=false, tick label style={font=\scriptsize},
    label style={font=\footnotesize},
    ylabel style={rotate=-90,text=poolRed},
    yticklabel style={text=poolRed},
    axis x line=none, axis y line*=right,
    axis line style={black!45}, tick style={black!45},
    tick align=outside, ytick pos=right,
    xmajorgrids=false, ymajorgrids=false,
]
\addplot[poolRed,line width=1pt,mark=*,mark size=2.3pt]
    table[x=steps,y=is] {\poolSamplingStepsData};
\end{axis}
\end{tikzpicture}
\endgroup
\caption{Sampling-step ablation for the $4\times4$ generator trained for 300 epochs. Gold shows FID (left axis); red shows IS (right axis). The IG scale is fixed at 2.75 rather than retuned for each step count.}
\label{fig:sampling-steps}
\end{figure}

%% file: figures/speed.tex
\begin{figure}[t]
\centering
\input{figures/speed_plot.tex}
\caption{Denoiser forward-pass speed on a single GPU, using the flow-model architecture in Table~\ref{tab:arch-hparams}. Legend entries give pooling windows and token compression. Each point is the mean of 30 measurements averaging 20 forward calls; error bars show one standard deviation and are mostly smaller than the markers. Benchmarks use BF16 computation and include host dispatch and synchronization, but not end-to-end image generation.}
\label{fig:inference-speed}
\end{figure}
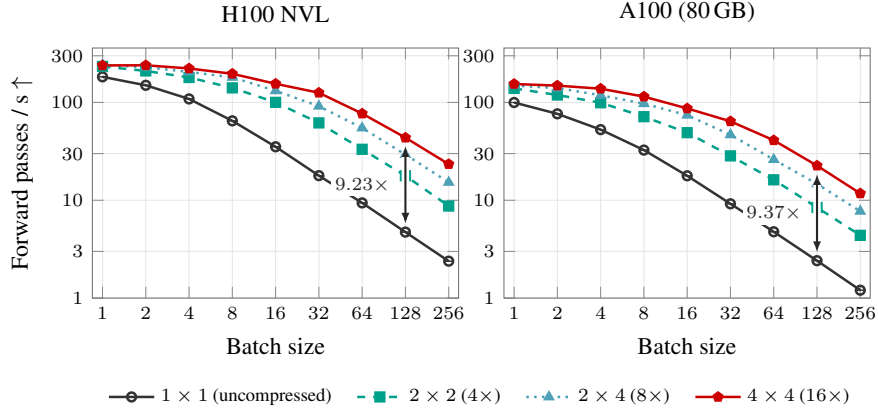

%% file: figures/speed_plot.tex
% Requires pgfplots/groupplots and the shared pooldino_style.tex palette.
% Kept separate from the float so it can also be compiled as a standalone PDF.
\begingroup
\input{figures/speed_data.tex}
\pgfplotsset{
    speed axis/.style={
        width=0.46\textwidth, height=0.35\textwidth,
        xmode=log, log basis x=2, ymode=log,
        log ticks with fixed point,
        xmin=0.85, xmax=300, ymin=1, ymax=360,
        xtick={1,2,4,8,16,32,64,128,256},
        ytick={1,3,10,30,100,300},
        minor tick num=0,
        xlabel={Batch size},
        tick label style={font=\scriptsize},
        label style={font=\small}, title style={font=\small},
        axis line style={black!55}, tick style={black!45},
        grid=major, grid style={black!10},
        every axis plot/.append style={line width=0.95pt, mark size=1.65pt},
        error bars/y dir=both, error bars/y explicit,
        error bars/error bar style={line width=0.5pt},
        error bars/error mark options={rotate=90, mark size=1.4pt},
    },
    speed uncompressed/.style={black!80, mark=o, mark options={solid, fill=white}},
    speed two/.style={poolTeal, dashed, mark=square*, mark options={solid}},
    speed eight/.style={poolCyan!85!black, dotted, mark=triangle*, mark options={solid}},
    speed four/.style={poolRed!80!black, mark=pentagon*, mark options={solid}},
}
\begin{tikzpicture}
\begin{groupplot}[
    speed axis,
    group style={group size=2 by 1, horizontal sep=0.6cm},
]
\nextgroupplot[
    title={H100 NVL},
    ylabel={Forward passes / s $\uparrow$},
    legend to name=poolspeedlegend,
    legend style={
        draw=none, font=\scriptsize, legend columns=4,
        /tikz/every even column/.append style={column sep=0.25cm},
    },
]
\addplot+[speed uncompressed] table[x=batch,y=mean,y error=std] {\speedHOne};
\addlegendentry{$1\times1$ (uncompressed)}
\addplot+[speed two] table[x=batch,y=mean,y error=std] {\speedHTwo};
\addlegendentry{$2\times2$ ($4\times$)}
\addplot+[speed eight] table[x=batch,y=mean,y error=std] {\speedHEight};
\addlegendentry{$2\times4$ ($8\times$)}
\addplot+[speed four] table[x=batch,y=mean,y error=std] {\speedHFour};
\addlegendentry{$4\times4$ ($16\times$)}
% Row 7 is batch size 128 in both source tables.
\poolSpeedupAtRow{7}{\speedHOne}{\speedHFour}{speedH}

\nextgroupplot[title={A100 (80\,GB)}]
\addplot+[speed uncompressed] table[x=batch,y=mean,y error=std] {\speedAOne};
\addplot+[speed two] table[x=batch,y=mean,y error=std] {\speedATwo};
\addplot+[speed eight] table[x=batch,y=mean,y error=std] {\speedAEight};
\addplot+[speed four] table[x=batch,y=mean,y error=std] {\speedAFour};
\poolSpeedupAtRow{7}{\speedAOne}{\speedAFour}{speedA}
\end{groupplot}
\poolDrawSpeedup{speedH}
\poolDrawSpeedup{speedA}
\path (group c1r1.south west) -- coordinate (speedbottom) (group c2r1.south east);
\node[anchor=north, inner sep=0pt] at ([yshift=-1.05cm]speedbottom)
    {\pgfplotslegendfromname{poolspeedlegend}};
\end{tikzpicture}
\endgroup

%% file: figures/speed_data.tex
% Extracted from fwd_measurements.json: forward, status=ok.
% Corrected model: trunk width 1440 / 20 heads; head width 2048 / 16 heads.
% Rates are batched iterations per second, not end-to-end generated images/s.
% SD is across 30 averages of 20 iterations; no rounding applied here.
\pgfplotstableread{
batch mean std
1 182.72760882661908 0.27036021549841216
2 149.72546039781085 0.09557328735887884
4 108.59688355908295 0.5198457417237784
8 64.8404387296805 0.33043188188927936
16 35.22938513360999 0.17662978246441322
32 17.852131717228104 0.06404435728200553
64 9.397682018706515 0.028576913563117798
128 4.726147980755713 0.012323200768866931
256 2.3844471622865804 0.006306229048182618
}\speedHOne

\pgfplotstableread{
batch mean std
1 233.13302457070418 0.12329583754108972
2 209.9961994605691 0.08393957862228185
4 179.96366014764214 0.13818830393188855
8 140.89768939837614 0.11962284988107597
16 100.14322466263717 0.8667446009173235
32 61.78344410140229 0.26013558190141095
64 33.2015156115808 0.12874161137459944
128 17.797022074641486 0.055197464403504606
256 8.7307812633132 0.03670461383206777
}\speedHTwo

\pgfplotstableread{
batch mean std
1 242.01250071934103 0.14122322719494618
2 228.9407450839676 0.1012646171071865
4 205.1455008929035 2.888477886216819
8 180.55276829863058 0.08999796555548874
16 131.55010021941058 0.07118713221475188
32 91.56423006417305 0.48084142164227506
64 54.83589566666757 0.34823764476161484
128 29.322956608012706 0.10134674607151228
256 15.235474530058788 0.03369338055364425
}\speedHEight

\pgfplotstableread{
batch mean std
1 239.10403207259807 0.0950041260172339
2 239.74532560611803 0.13997819895937585
4 222.56967343785234 0.1756681108438191
8 196.29743463322072 0.23226895218616964
16 155.165402913031 0.12483124844952886
32 125.5995164323638 0.5007358771514275
64 76.93556169175905 0.15485577474403175
128 43.6245243365758 0.14642990823051627
256 23.413744013397093 0.0811654589339671
}\speedHFour

\pgfplotstableread{
batch mean std
1 99.81289556205971 0.12760244058243617
2 76.49811988667663 0.08442636483803184
4 52.74411233558868 0.1538144005413504
8 32.58272711403607 0.04933348489026639
16 17.787440154252188 0.0473046331411705
32 9.200642682681229 0.0384325044733852
64 4.761887960153132 0.01597622700220833
128 2.411070852221283 0.00802243470377441
256 1.2033154177264027 0.0029965596986951336
}\speedAOne

\pgfplotstableread{
batch mean std
1 139.1594057166154 0.2064000732306605
2 119.14165483618638 0.17022471426896313
4 99.28093325347012 0.12489048822626746
8 71.52485440899896 0.097080513335979
16 49.08149361070462 0.2246024653522096
32 28.42776794507605 0.07584723515646352
64 16.189066382230365 0.03137333674292512
128 8.436855507826696 0.031223953713396046
256 4.378603694582013 0.0200642264017834
}\speedATwo

\pgfplotstableread{
batch mean std
1 148.53020347605732 0.23135068945183582
2 140.37065445422098 0.19142079897476214
4 118.39413264751295 0.2180327926284999
8 97.18720140327338 1.129914504862595
16 74.1159985145025 0.1701862487020309
32 46.689535415414 0.11703369917422844
64 26.05784481560238 0.05008751991454664
128 14.698797176750885 0.04399940953569515
256 7.7153988013571 0.030741524299600396
}\speedAEight

\pgfplotstableread{
batch mean std
1 154.3991683361102 0.22089959195645162
2 148.63897389799027 0.20240891757435256
4 138.14522435491872 0.2775034720486328
8 114.64380538601425 0.20575696000866192
16 86.8459415093857 0.17299713902127167
32 64.07931071270991 0.10761467431499756
64 41.02520458120065 0.10792193843971819
128 22.586490934191804 0.05436183895812986
256 11.741996367687728 0.03428679700869795
}\speedAFour

%% file: figures/train_speed.tex
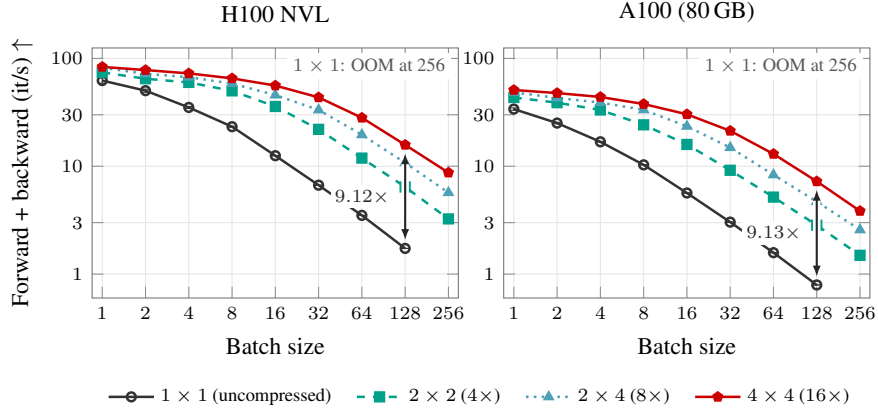
\begin{figure}[t]
\centering
\input{figures/train_speed_plot.tex}
\caption{Generator forward--backward throughput on a single GPU. Points show mean iterations/s across 30 measurements of 20 iterations. Timing includes the forward pass, main-prediction MSE loss, backward pass, and dispatch/synchronization; it excludes optimizer updates, auxiliary losses, and the encoder/decoder. The uncompressed model runs out of memory at batch size 256 on both GPUs.}
\label{fig:training-speed}
\end{figure}

%% file: figures/train_speed_plot.tex
% Requires pgfplots/groupplots and the shared pooldino_style.tex palette.
% Kept separate from the float so it can also be compiled as a standalone PDF.
\begingroup
\input{figures/train_speed_data.tex}
\pgfplotsset{
    train speed axis/.style={
        width=0.46\textwidth, height=0.35\textwidth,
        xmode=log, log basis x=2, ymode=log,
        log ticks with fixed point,
        xmin=0.85, xmax=300, ymin=0.6, ymax=125,
        xtick={1,2,4,8,16,32,64,128,256},
        ytick={1,3,10,30,100},
        minor tick num=0,
        xlabel={Batch size},
        tick label style={font=\scriptsize},
        label style={font=\small}, title style={font=\small},
        axis line style={black!55}, tick style={black!45},
        grid=major, grid style={black!10},
        every axis plot/.append style={line width=0.95pt, mark size=1.65pt},
        error bars/y dir=both, error bars/y explicit,
        error bars/error bar style={line width=0.5pt},
        error bars/error mark options={rotate=90, mark size=1.4pt},
    },
    train speed uncompressed/.style={black!80, mark=o, mark options={solid, fill=white}},
    train speed two/.style={poolTeal, dashed, mark=square*, mark options={solid}},
    train speed eight/.style={poolCyan!85!black, dotted, mark=triangle*, mark options={solid}},
    train speed four/.style={poolRed!80!black, mark=pentagon*, mark options={solid}},
}
\begin{tikzpicture}
\begin{groupplot}[
    train speed axis,
    group style={group size=2 by 1, horizontal sep=0.6cm},
]
\nextgroupplot[
    title={H100 NVL},
    ylabel={Forward + backward (it/s) $\uparrow$},
    legend to name=pooltrainspeedlegend,
    legend style={
        draw=none, font=\scriptsize, legend columns=4,
        /tikz/every even column/.append style={column sep=0.25cm},
    },
]
\addplot+[train speed uncompressed] table[x=batch,y=mean,y error=std] {\trainSpeedHOne};
\addlegendentry{$1\times1$ (uncompressed)}
\addplot+[train speed two] table[x=batch,y=mean,y error=std] {\trainSpeedHTwo};
\addlegendentry{$2\times2$ ($4\times$)}
\addplot+[train speed eight] table[x=batch,y=mean,y error=std] {\trainSpeedHEight};
\addlegendentry{$2\times4$ ($8\times$)}
\addplot+[train speed four] table[x=batch,y=mean,y error=std] {\trainSpeedHFour};
\addlegendentry{$4\times4$ ($16\times$)}
% Row 7 is batch size 128 in both source tables.
\poolSpeedupAtRow{7}{\trainSpeedHOne}{\trainSpeedHFour}{trainSpeedH}
\node[anchor=north east, font=\scriptsize, text=black!65, fill=white, inner sep=2pt]
    at (rel axis cs:0.98,0.98) {$1\times1$: OOM at 256};

\nextgroupplot[title={A100 (80\,GB)}]
\addplot+[train speed uncompressed] table[x=batch,y=mean,y error=std] {\trainSpeedAOne};
\addplot+[train speed two] table[x=batch,y=mean,y error=std] {\trainSpeedATwo};
\addplot+[train speed eight] table[x=batch,y=mean,y error=std] {\trainSpeedAEight};
\addplot+[train speed four] table[x=batch,y=mean,y error=std] {\trainSpeedAFour};
\poolSpeedupAtRow{7}{\trainSpeedAOne}{\trainSpeedAFour}{trainSpeedA}
\node[anchor=north east, font=\scriptsize, text=black!65, fill=white, inner sep=2pt]
    at (rel axis cs:0.98,0.98) {$1\times1$: OOM at 256};
\end{groupplot}
\poolDrawSpeedup{trainSpeedH}
\poolDrawSpeedup{trainSpeedA}
\path (group c1r1.south west) -- coordinate (trainspeedbottom) (group c2r1.south east);
\node[anchor=north, inner sep=0pt] at ([yshift=-1.05cm]trainspeedbottom)
    {\pgfplotslegendfromname{pooltrainspeedlegend}};
\end{tikzpicture}
\endgroup

%% file: figures/train_speed_data.tex
% Extracted from fwd_bwd_measurements.json: forward_backward, status=ok.
% Corrected model: trunk width 1440 / 20 heads; head width 2048 / 16 heads.
% Rates are batched iterations per second, not end-to-end generated images/s.
% SD is across 30 averages of 20 iterations; no rounding applied here.
% Both 1x1/b256 configurations are OOM, not zero-throughput observations.
\pgfplotstableread{
batch mean std
1 62.32266712812237 3.461302758187643
2 50.21288812163783 1.6648369692850666
4 35.09985267355631 0.7316593031083455
8 23.26676461800486 0.28475843237872245
16 12.524748364588937 0.04855936562604261
32 6.671519306939505 0.02338487210322735
64 3.503132238720803 0.014331910382217112
128 1.730092479868648 0.002976689656796567
}\trainSpeedHOne

\pgfplotstableread{
batch mean std
1 74.22323632653114 4.875761223352631
2 64.75896022471878 3.8014678683453025
4 59.90773065620526 3.202882836563244
8 50.01947807999181 1.5622207258298637
16 35.82655784683305 0.7900530638471276
32 21.991986640077968 0.24112742042237553
64 11.897046645945958 0.055369168649799544
128 6.4274900394230565 0.019408810726205355
256 3.2489602517435405 0.01921741793941907
}\trainSpeedHTwo

\pgfplotstableread{
batch mean std
1 82.06379256600262 4.137602703495575
2 71.9152359091309 4.668429786790525
4 67.01324728748392 4.05077293395212
8 58.40410947330228 2.182640118812897
16 45.70035098038307 1.371691481140611
32 33.28464923702592 0.6874873268236946
64 19.583895231241698 0.19356520092681612
128 10.76464620554332 0.03165814624840154
256 5.6855591408581505 0.014311895073094235
}\trainSpeedHEight

\pgfplotstableread{
batch mean std
1 83.40104252539066 4.2590902544901095
2 77.9619815559563 3.8033885756505477
4 72.60763326944681 3.2874918273741307
8 65.28665711260956 3.836406391952955
16 55.9879417044683 1.9577617579391047
32 43.4973054518789 1.215711764853718
64 28.158028564248593 0.44739154987091584
128 15.782353954447329 0.0994722000495899
256 8.727571147891874 0.02203739974330587
}\trainSpeedHFour

\pgfplotstableread{
batch mean std
1 33.902519935150224 0.8370176545382295
2 25.135258257286804 0.41702531081391014
4 16.85058002476927 0.14788980512684569
8 10.301882590186096 0.031176295565806283
16 5.648138950386823 0.01955462480370252
32 3.033317124754861 0.015565486652837343
64 1.5794156426154258 0.008139776836344663
128 0.7928358309239157 0.002065548743789152
}\trainSpeedAOne

\pgfplotstableread{
batch mean std
1 43.41593856790707 1.4261311201706814
2 38.88217991006558 1.0842669065496366
4 33.24220269635211 0.7881470825731919
8 24.215953012303736 0.37207417794670017
16 15.931893987353785 0.1779374250872942
32 9.175852963025248 0.04254032597320781
64 5.174573634808949 0.021004101106491077
128 2.840250078664469 0.012790206481598586
256 1.4919162583268877 0.0057566962788190325
}\trainSpeedATwo

\pgfplotstableread{
batch mean std
1 48.41762900847518 1.7123823505335318
2 42.986321980733706 1.3518820454312912
4 38.940256804058535 1.096937050342179
8 33.321425207672156 0.767924899127247
16 23.46710286734641 0.34825492398383395
32 14.88434197731429 0.15421474006282948
64 8.319948814378757 0.024333514838044648
128 4.702749032190764 0.01824114544690661
256 2.5702840553895823 0.014779178859215363
}\trainSpeedAEight

\pgfplotstableread{
batch mean std
1 50.91827615512681 2.136063743710017
2 47.70918979313895 1.6480890025567843
4 43.88545287861998 1.391023168020737
8 37.75740672921994 0.9943045244024817
16 30.320168759556516 0.626276228193888
32 21.29516788269127 0.27469215077221437
64 12.997017903886805 0.09034768949530962
128 7.241374087640133 0.023754223203077407
256 3.845537893699195 0.017000191322008668
}\trainSpeedAFour

%% file: figures/isolated_guidance_sweeps.tex
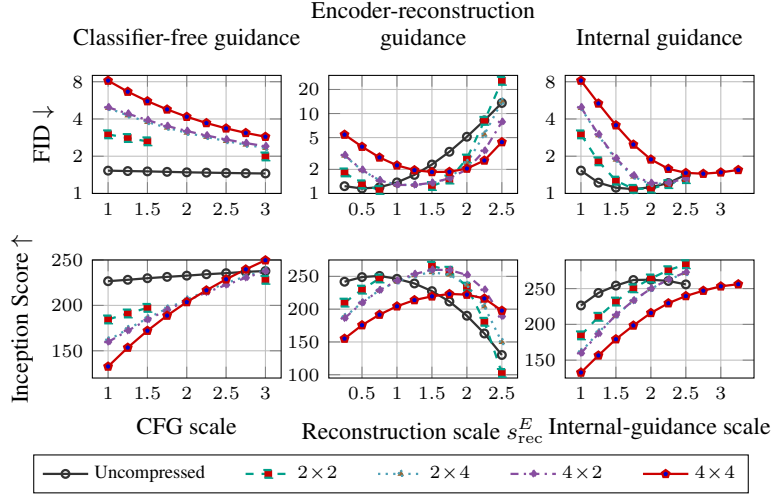
\begin{figure}[H]
\centering
\pgfplotsset{
    isolated guidance axis/.style={
        width=0.292\textwidth,
        height=0.225\textwidth,
        grid=major,
        unbounded coords=jump,
        tick label style={font=\scriptsize},
        label style={font=\small},
        title style={font=\small},
        every axis plot/.append style={line width=0.85pt, mark size=1.45pt},
    },
    uncompressed curve/.style={black!80, mark=o},
    twobytwo curve/.style={poolTeal, dashed, mark=square*},
    twobyfour curve/.style={poolCyan!85!black, dotted, mark=triangle*},
    fourbytwo curve/.style={poolPurple, dashdotted, mark=diamond*},
    fourbyfour curve/.style={poolRed!80!black, mark=pentagon*},
}
\begin{tikzpicture}
\begin{groupplot}[
    isolated guidance axis,
    group style={group size=3 by 2, horizontal sep=0.63cm, vertical sep=0.88cm},
]

\nextgroupplot[
    title={Classifier-free guidance},
    ylabel={FID $\downarrow$},
    ymode=log,
    log ticks with fixed point,
    ymin=1, ymax=9,
    ytick={1,2,4,8},
    xtick={1,1.5,2,2.5,3},
    legend to name=isolatedguidancelegend,
    legend style={font=\scriptsize, legend columns=5, /tikz/every even column/.append style={column sep=0.45cm}},
]
\addplot+[uncompressed curve] coordinates {
    (1,1.528786) (1.25,1.517083) (1.5,1.507897) (1.75,1.495018)
    (2,1.482275) (2.25,1.473433) (2.5,1.464692) (2.75,1.456355) (3,1.450217)
};
\addlegendentry{Uncompressed}
\addplot+[twobytwo curve] coordinates {
    (1,3.004695) (1.25,2.805519) (1.5,2.637863)
    (1.75,nan) (2,nan) (2.25,nan) (2.5,nan) (2.75,nan) (3,1.991862)
};
\addlegendentry{$2\!\times\!2$}
\addplot+[twobyfour curve] coordinates {
    (1,4.906273) (1.25,4.268915) (1.5,3.777737) (1.75,3.397641)
    (2,3.090340) (2.25,2.860638) (2.5,2.650288) (2.75,2.476977) (3,2.329361)
};
\addlegendentry{$2\!\times\!4$}
\addplot+[fourbytwo curve] coordinates {
    (1,4.969498) (1.25,4.416938) (1.5,3.913689) (1.75,3.508656)
    (2,3.197997) (2.25,2.942109) (2.5,2.729170) (2.75,2.554116) (3,2.407647)
};
\addlegendentry{$4\!\times\!2$}
\addplot+[fourbyfour curve] coordinates {
    (1,8.164129) (1.25,6.649615) (1.5,5.571557) (1.75,4.778291)
    (2,4.169128) (2.25,3.709889) (2.5,3.358222) (2.75,3.086567) (3,2.874497)
};
\addlegendentry{$4\!\times\!4$}

\nextgroupplot[
    title={Encoder-reconstruction\\guidance}, title style={align=center},
    ymode=log,
    log ticks with fixed point,
    ymin=1, ymax=30,
    ytick={1,2,5,10,20},
    xtick={0.5,1,1.5,2,2.5},
]
\addplot+[uncompressed curve] coordinates {
    (0.25,1.233671) (0.5,1.153618) (0.75,1.194157) (1,1.375094)
    (1.25,1.707394) (1.5,2.302052) (1.75,3.324902) (2,5.124962)
    (2.25,8.212867) (2.5,13.554090)
};
\addplot+[twobytwo curve] coordinates {
    (0.25,1.832613) (0.5,1.299976) (0.75,1.105845)
    (1,nan) (1.25,nan)
    (1.5,1.238814) (1.75,1.492849) (2,2.733353) (2.25,8.097199) (2.5,25.626920)
};
\addplot+[twobyfour curve] coordinates {
    (0.25,2.962753) (0.5,1.949179) (0.75,1.466105) (1,1.283642)
    (1.25,1.276877) (1.5,1.340912) (1.75,1.545247) (2,2.404185)
    (2.25,5.476777) (2.5,14.112163)
};
\addplot+[fourbytwo curve] coordinates {
    (0.25,3.017840) (0.5,1.974285) (0.75,1.454566) (1,1.278203)
    (1.25,1.276419) (1.5,1.359711) (1.75,1.545388) (2,1.998178)
    (2.25,3.425963) (2.5,7.854625)
};
\addplot+[fourbyfour curve] coordinates {
    (0.25,5.445989) (0.5,3.845442) (0.75,2.822361) (1,2.246253)
    (1.25,1.957371) (1.5,1.857149) (1.75,1.865732) (2,2.048500)
    (2.25,2.584892) (2.5,4.392789)
};

\nextgroupplot[
    title={Internal guidance},
    ymode=log,
    log ticks with fixed point,
    ymin=1, ymax=9,
    ytick={1,2,4,8},
    xtick={1,1.5,2,2.5,3},
]
\addplot+[uncompressed curve] coordinates {
    (1,1.531265) (1.25,1.222275) (1.5,1.113251) (1.75,1.083134)
    (2,1.122641) (2.25,1.213580) (2.5,1.417889)
};
\addplot+[twobytwo curve] coordinates {
    (1,3.001905) (1.25,1.818002) (1.5,1.269792) (1.75,1.088975)
    (2,1.093832) (2.25,1.189809) (2.5,1.305514)
};
\addplot+[twobyfour curve] coordinates {
    (1,4.903648) (1.25,2.946225) (1.5,1.872620) (1.75,1.366922)
    (2,1.191260) (2.25,1.206589) (2.5,1.298969)
};
\addplot+[fourbytwo curve] coordinates {
    (1,4.971094) (1.25,3.003647) (1.5,1.919844) (1.75,1.398181)
    (2,1.213788) (2.25,1.219843) (2.5,1.322639)
};
\addplot+[fourbyfour curve] coordinates {
    (1,8.169008) (1.25,5.338555) (1.5,3.559785) (1.75,2.486142)
    (2,1.883799) (2.25,1.583331) (2.5,1.467013) (2.75,1.444216)
    (3,1.480201) (3.25,1.549359)
};

\nextgroupplot[
    xlabel={CFG scale},
    ylabel={Inception Score $\uparrow$},
    ymin=120, ymax=250,
    xtick={1,1.5,2,2.5,3},
]
\addplot+[uncompressed curve] coordinates {
    (1,226.632043) (1.25,228.177435) (1.5,229.900618) (1.75,231.387651)
    (2,232.731741) (2.25,234.272157) (2.5,235.464822) (2.75,236.952200) (3,237.799852)
};
\addplot+[twobytwo curve] coordinates {
    (1,184.492383) (1.25,191.071475) (1.5,197.250835)
    (1.75,nan) (2,nan) (2.25,nan) (2.5,nan) (2.75,nan) (3,228.100533)
};
\addplot+[twobyfour curve] coordinates {
    (1,161.423326) (1.25,173.873729) (1.5,185.783794) (1.75,196.526980)
    (2,206.433046) (2.25,215.940202) (2.5,224.543538) (2.75,231.968558) (3,238.804895)
};
\addplot+[fourbytwo curve] coordinates {
    (1,159.671309) (1.25,172.068678) (1.5,184.077769) (1.75,195.073792)
    (2,205.031273) (2.25,214.222771) (2.5,222.496083) (2.75,230.358301) (3,237.632010)
};
\addplot+[fourbyfour curve] coordinates {
    (1,132.672199) (1.25,153.497641) (1.5,172.107581) (1.75,188.754810)
    (2,203.667041) (2.25,216.600856) (2.5,228.737341) (2.75,239.660403) (3,249.720816)
};

\nextgroupplot[
    xlabel={Reconstruction scale $s_{\mathrm{rec}}^{E}$},
    ymin=95, ymax=275,
    xtick={0.5,1,1.5,2,2.5},
]
\addplot+[uncompressed curve] coordinates {
    (0.25,241.824937) (0.5,248.920952) (0.75,250.691595) (1,246.262241)
    (1.25,239.173260) (1.5,227.481204) (1.75,211.677072) (2,189.958333)
    (2.25,162.909264) (2.5,130.104253)
};
\addplot+[twobytwo curve] coordinates {
    (0.25,209.950487) (0.5,229.757443) (0.75,246.795239)
    (1,nan) (1.25,nan)
    (1.5,266.172797) (1.75,258.549203) (2,234.258583) (2.25,180.756306) (2.5,102.714334)
};
\addplot+[twobyfour curve] coordinates {
    (0.25,187.871918) (0.5,210.077997) (0.75,228.998314) (1,242.335274)
    (1.25,251.521300) (1.5,256.314565) (1.75,252.466948) (2,236.774895)
    (2.25,204.163498) (2.5,149.760181)
};
\addplot+[fourbytwo curve] coordinates {
    (0.25,185.973390) (0.5,209.665268) (0.75,228.779451) (1,243.299591)
    (1.25,253.504202) (1.5,260.010198) (1.75,259.725894) (2,252.136589)
    (2.25,229.596243) (2.5,188.887071)
};
\addplot+[fourbyfour curve] coordinates {
    (0.25,155.076117) (0.5,175.605716) (0.75,191.650444) (1,204.378342)
    (1.25,213.856459) (1.5,219.530664) (1.75,223.200310) (2,222.161534)
    (2.25,216.240768) (2.5,197.745572)
};

\nextgroupplot[
    xlabel={Internal-guidance scale},
    ymin=125, ymax=290,
    xtick={1,1.5,2,2.5,3},
]
\addplot+[uncompressed curve] coordinates {
    (1,226.410039) (1.25,243.940845) (1.5,254.208043) (1.75,261.997809)
    (2,262.399236) (2.25,260.686333) (2.5,255.859261)
};
\addplot+[twobytwo curve] coordinates {
    (1,184.615826) (1.25,210.576254) (1.5,231.887701) (1.75,250.362663)
    (2,264.160016) (2.25,275.671985) (2.5,283.993359)
};
\addplot+[twobyfour curve] coordinates {
    (1,161.293666) (1.25,188.802184) (1.5,212.468024) (1.75,233.080403)
    (2,249.577541) (2.25,262.601457) (2.5,271.770956)
};
\addplot+[fourbytwo curve] coordinates {
    (1,159.626884) (1.25,186.890512) (1.5,213.231194) (1.75,233.194673)
    (2,249.963042) (2.25,262.651419) (2.5,272.928967)
};
\addplot+[fourbyfour curve] coordinates {
    (1,132.634711) (1.25,156.584772) (1.5,179.289087) (1.75,198.432138)
    (2,216.004582) (2.25,229.574306) (2.5,239.625264) (2.75,247.073224)
    (3,253.170390) (3.25,256.155474)
};

\end{groupplot}
\end{tikzpicture}

\pgfplotslegendfromname{isolatedguidancelegend}
\caption{Isolated guidance-scale sweeps on ImageNet-256. Top: FID on a logarithmic scale. Bottom: Inception Score. Missing points in the corrected $2\times2$ sweeps are left blank.}
\label{fig:isolated-guidance-sweeps}
\end{figure}

%% file: figures/joint_guidance_sweep_data.tex
% Joint 100-step guidance sweeps from W&B.
% Columns are: internal/DINO-patch-guidance scale, CFG scale, FID, Inception Score.
% Add newly evaluated configurations as one row in the corresponding table.

\pgfplotstableread[col sep=space]{
guide cfg fid is
0.25 2.50 1.200711 249.164621
0.25 3.00 1.194238 251.600940
0.25 3.25 1.187026 252.768216
0.25 3.50 1.188120 253.769551
0.25 3.75 1.186446 254.159030
0.50 2.50 1.135550 255.308585
0.50 3.00 1.128447 257.534737
0.50 3.25 1.121783 259.036615
0.50 3.50 1.123191 259.217639
0.50 3.75 1.122073 260.141481
0.75 2.50 1.182118 256.160300
0.75 3.00 1.181915 258.292946
0.75 3.25 1.179105 258.970786
0.75 3.50 1.180306 259.708389
0.75 3.75 1.180793 260.163853
1.00 1.25 1.340923 249.347859
1.00 1.50 1.340111 250.543491
1.00 1.75 1.365637 250.246039
1.00 2.00 1.365383 250.967082
1.00 2.25 1.366380 251.488416
1.00 2.50 1.369639 252.385556
1.00 2.75 1.366274 252.670566
1.00 3.00 1.365800 253.264809
1.00 3.25 1.362728 254.176308
1.00 3.50 1.368803 254.585866
}\jointDINOUncompressed

\pgfplotstableread[col sep=space]{
guide cfg fid is
0.50 1.50 1.201020 243.582079
0.50 1.75 1.170582 249.631329
0.50 2.00 1.147423 254.829755
0.50 2.25 1.128443 259.441223
0.50 2.50 1.116654 263.773767
0.75 1.50 1.070756 259.627158
0.75 1.75 1.063435 264.779135
0.75 2.00 1.061313 269.290085
0.75 2.25 1.060554 272.827356
0.75 2.50 1.061443 275.957129
1.00 1.25 1.082197 266.273117
1.00 1.50 1.083353 271.563898
1.00 1.75 1.091525 275.725513
1.00 2.00 1.100510 279.197452
1.00 2.25 1.103998 282.637976
1.00 2.50 1.109026 285.249045
1.00 2.75 1.112707 288.109116
1.00 3.00 1.116958 290.639331
1.25 1.50 1.148584 277.843674
1.25 2.00 1.163121 284.846158
}\jointDINOTwoTwo

\pgfplotstableread[col sep=space]{
guide cfg fid is
0.50 2.00 1.351780 255.654977
0.50 2.25 1.300880 264.245120
0.50 2.50 1.273290 271.333636
0.75 1.25 1.307666 242.858806
0.75 1.75 1.198910 264.145441
0.75 2.00 1.179979 272.362341
0.75 2.25 1.174532 279.679764
0.75 2.50 1.177614 286.606052
1.00 1.25 1.215024 256.429395
1.00 1.50 1.188099 267.424753
1.00 1.75 1.181869 276.595294
1.00 2.00 1.179445 284.268518
1.00 2.25 1.187359 290.772815
1.00 2.50 1.200080 296.539813
1.00 2.75 1.215682 301.122403
1.00 3.00 1.228728 305.289389
1.25 1.25 1.240156 265.495378
1.25 1.75 1.244273 283.899005
1.25 2.25 1.272086 295.747214
1.50 1.25 1.316846 269.567941
1.50 1.75 1.324816 285.662350
1.50 2.25 1.346988 295.713278
}\jointDINOTwoFour

\pgfplotstableread[col sep=space]{
guide cfg fid is
0.50 2.00 1.387748 255.986737
0.50 2.25 1.340088 264.245273
0.50 2.50 1.303463 272.492444
0.75 1.25 1.332223 242.284653
0.75 1.75 1.226092 264.339291
0.75 2.00 1.205700 272.837772
0.75 2.25 1.199637 280.136700
0.75 2.50 1.207029 287.423691
1.00 1.25 1.240609 255.625449
1.00 1.50 1.216554 266.483997
1.00 1.75 1.214037 275.913712
1.00 2.00 1.222905 284.100171
1.00 2.25 1.235854 290.791724
1.00 2.50 1.245932 296.247134
1.00 2.75 1.255866 301.731958
1.00 3.00 1.265591 306.047888
1.25 1.25 1.282690 265.538843
1.25 1.75 1.287704 283.582312
1.25 2.25 1.326520 296.276456
1.50 1.25 1.387443 269.068106
1.50 1.75 1.410792 286.398669
1.50 2.25 1.440125 296.732019
}\jointDINOFourTwo

\pgfplotstableread[col sep=space]{
guide cfg fid is
0.75 2.25 1.416306 279.318762
0.75 2.50 1.413556 287.700076
0.75 3.00 1.405782 304.070895
0.75 3.50 1.416821 316.380457
1.00 1.25 1.775169 228.925694
1.00 1.50 1.547770 247.103531
1.00 1.75 1.427301 262.181917
1.00 2.00 1.374222 275.305161
1.00 2.25 1.358833 285.992111
1.00 2.50 1.354372 295.385950
1.00 2.75 1.360582 302.664297
1.00 3.00 1.367749 308.919876
1.00 3.25 1.373935 313.748264
1.00 3.50 1.385289 318.845114
1.00 3.75 1.399505 323.702902
1.00 4.00 1.409070 325.677774
1.00 4.25 1.415408 328.635281
1.00 4.50 1.420817 331.299265
1.25 2.50 1.375293 298.342618
1.25 3.00 1.399085 309.458966
1.50 2.50 1.440057 298.032727
1.50 3.00 1.474096 307.273260
}\jointDINOFourFour

% Includes the fixed-IG=1.78 CFG sweep optimum recorded in
% logs/POOLDINO_PAPER_PLAN_AND_RESULTS.tex.
\pgfplotstableread[col sep=space]{
guide cfg fid is
1.50 1.75 1.106625 257.797079
1.50 2.00 1.103617 259.085022
1.50 2.25 1.099308 260.348454
1.50 2.50 1.098560 261.527515
1.50 3.00 1.097651 263.516502
1.75 1.75 1.076339 264.990683
1.75 2.00 1.073687 265.514920
1.75 2.25 1.074925 266.666199
1.75 2.50 1.074050 267.331360
1.75 3.00 1.074517 268.645758
1.78 2.50 1.073590 267.149557
2.00 1.75 1.120845 265.026910
2.00 2.00 1.118966 265.807925
2.00 2.25 1.119950 267.030008
2.00 2.50 1.120414 267.666122
2.00 3.00 1.120939 268.725961
}\jointIGUncompressed

\pgfplotstableread[col sep=space]{
guide cfg fid is
1.25 1.25 1.708121 217.570644
1.25 1.75 1.549309 230.003600
1.25 2.25 1.442694 240.223470
1.50 1.50 1.181041 245.644197
1.50 1.75 1.155078 251.808018
1.50 2.00 1.134456 257.052988
1.75 1.25 1.074372 257.245525
1.75 1.50 1.071226 263.016583
1.75 1.75 1.068004 268.501685
1.75 2.00 1.069173 272.815323
1.75 2.25 1.073889 276.371445
2.00 1.50 1.111279 275.834610
2.00 1.75 1.120443 280.712006
2.00 2.00 1.130829 284.199887
2.25 1.25 1.205022 281.336197
2.25 1.75 1.225376 290.341412
2.25 2.25 1.248179 296.599002
}\jointIGTwoTwo

\pgfplotstableread[col sep=space]{
guide cfg fid is
1.50 1.50 1.517375 238.506314
1.50 2.00 1.341890 257.103418
1.50 2.50 1.274608 271.788043
1.75 1.25 1.268528 245.648586
1.75 1.50 1.211770 257.342162
1.75 1.75 1.181502 267.541940
2.00 1.25 1.168228 262.205328
2.00 1.50 1.161500 273.056979
2.00 1.75 1.166816 281.823291
2.00 2.00 1.184459 289.019147
2.00 2.50 1.223745 302.172568
2.25 1.25 1.217565 274.490427
2.25 1.50 1.235494 284.195175
2.25 1.75 1.260713 291.956714
2.50 1.50 1.364998 292.107242
2.50 2.00 1.429521 304.524100
2.50 2.50 1.464581 313.579340
}\jointIGTwoFour

\pgfplotstableread[col sep=space]{
guide cfg fid is
1.50 1.50 1.550064 239.262491
1.50 2.00 1.392946 258.250725
1.50 2.50 1.325534 274.596503
1.75 1.25 1.296212 246.920929
1.75 1.50 1.241581 258.738492
1.75 1.75 1.218568 268.230258
2.00 1.25 1.191424 263.266351
2.00 1.50 1.188949 273.984052
2.00 1.75 1.203462 283.302661
2.00 2.00 1.220976 290.852835
2.00 2.50 1.265351 302.912204
2.25 1.25 1.238662 274.682605
2.25 1.50 1.267204 284.665976
2.25 1.75 1.295564 293.068948
2.50 1.50 1.403867 293.198712
2.50 2.00 1.470630 306.977173
2.50 2.50 1.515116 315.737061
}\jointIGFourTwo

\pgfplotstableread[col sep=space]{
guide cfg fid is
1.75 1.75 1.572434 256.339928
1.75 2.00 1.502686 269.853128
1.75 2.25 1.467363 281.659106
2.00 1.75 1.391941 270.577646
2.00 2.00 1.378806 283.051898
2.00 2.25 1.390591 293.872372
2.00 2.50 1.406885 303.603476
2.00 3.00 1.453577 317.929593
2.25 1.75 1.352229 282.293063
2.25 2.00 1.370113 293.453339
2.25 2.25 1.393647 302.757617
2.50 2.00 1.421892 299.266116
2.50 2.50 1.474500 313.614487
2.50 3.00 1.522537 324.509326
3.00 2.00 1.584015 306.152563
3.00 2.50 1.631949 317.772830
3.00 3.00 1.649717 325.318427
}\jointIGFourFour

%% file: figures/joint_guidance_sweeps.tex
% Data live in joint_guidance_sweep_data.tex. Each measured configuration is
% drawn as one square; no interpolation is used across unevaluated settings.
\begingroup
\pgfplotsset{
    colormap={pool fid}{
        rgb255(0cm)=(0,160,138);
        rgb255(0.52cm)=(242,173,0);
        rgb255(1cm)=(210,45,45)
    },
    colormap={pool is}{
        rgb255(0cm)=(210,45,45);
        rgb255(0.48cm)=(242,173,0);
        rgb255(1cm)=(0,160,138)
    },
    joint guidance axis/.style={
        scale only axis,
        width=0.39\textwidth,
        height=0.29\textwidth,
        grid=major,
        axis line style={black!50},
        tick style={black!45},
        tick align=outside,
        grid style={black!10},
        scaled ticks=false,
        enlarge x limits=0.08,
        enlarge y limits=0.14,
        xtick distance=0.5,
        ytick distance=0.25,
        tick label style={font=\footnotesize},
        label style={font=\small},
        title style={font=\small},
    },
    joint fid points/.style={
        only marks,
        scatter,
        scatter src=explicit,
        mark=square*,
        mark size=6.5pt,
        point meta min=1.05,
        point meta max=1.80,
        colormap name=pool fid,
        scatter/use mapped color={draw=black!16, fill=mapped color},
    },
    joint is points/.style={
        only marks,
        scatter,
        scatter src=explicit,
        mark=square*,
        mark size=6.5pt,
        point meta min=215,
        point meta max=335,
        colormap name=pool is,
        scatter/use mapped color={draw=black!16, fill=mapped color},
    },
    joint optimum/.style={
        only marks,
        mark=square,
        mark size=8pt,
        mark options={draw=black!90, fill=none, line width=0.85pt},
    },
}

\newcommand{\jointfidplot}[2]{%
    \addplot+[joint fid points]
        table[x=cfg, y=guide, meta=fid] {#1};
    \addplot+[joint optimum] coordinates {#2};
}
\newcommand{\jointisplot}[2]{%
    \addplot+[joint is points]
        table[x=cfg, y=guide, meta=is] {#1};
    \addplot+[joint optimum] coordinates {#2};
}

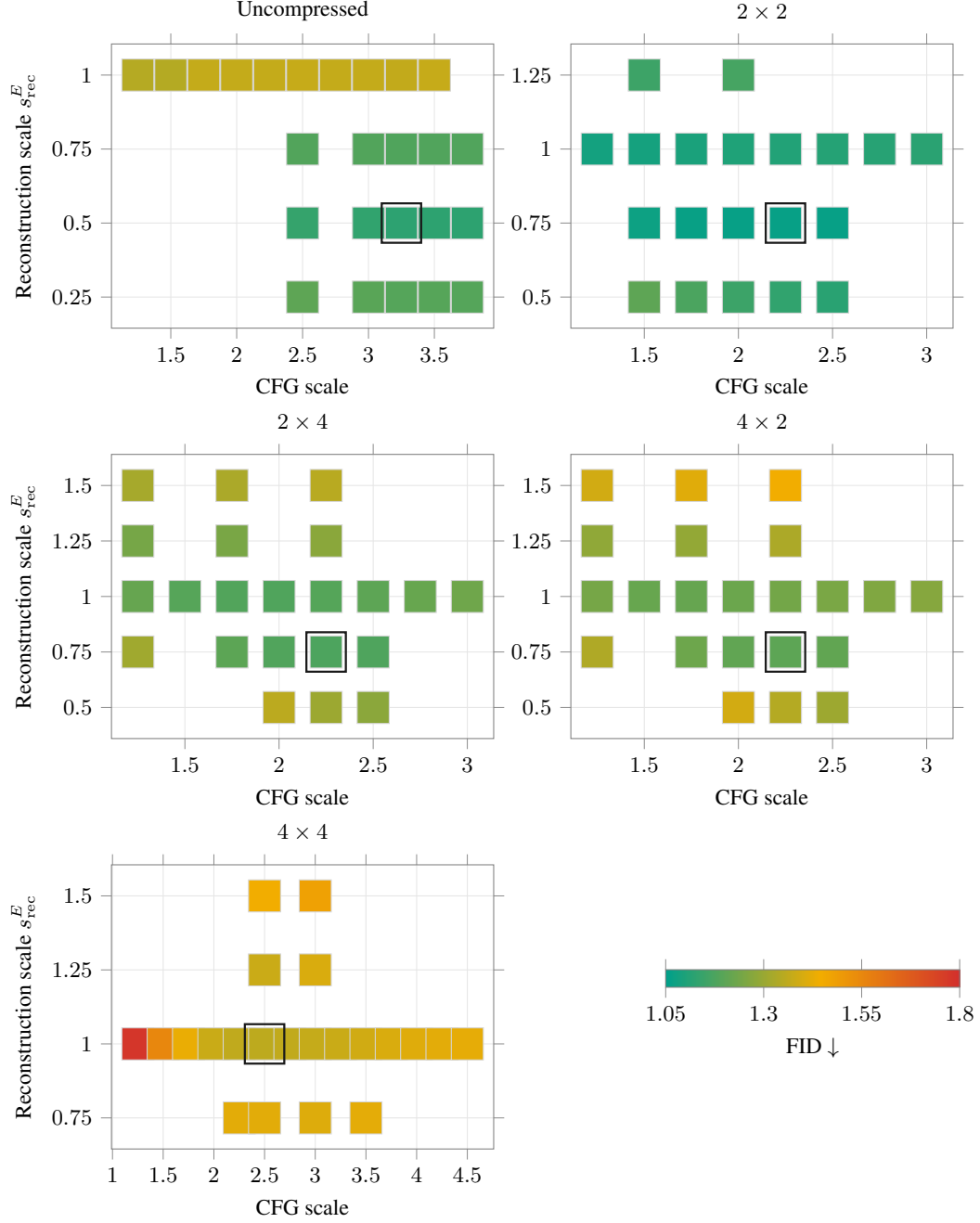
\begin{figure}[p]
\centering
\begin{tikzpicture}
\begin{groupplot}[
    joint guidance axis,
    xlabel={CFG scale},
    point meta min=1.05, point meta max=1.80,
    colormap name=pool fid,
    group style={group size=2 by 3, horizontal sep=1.10cm, vertical sep=1.80cm},
]
\nextgroupplot[title={Uncompressed}, ylabel={Reconstruction scale $s_{\mathrm{rec}}^{E}$}]
\jointfidplot{\jointDINOUncompressed}{(3.25,0.50)}

\nextgroupplot[title={$2\times2$}]
\jointfidplot{\jointDINOTwoTwo}{(2.25,0.75)}

\nextgroupplot[title={$2\times4$}, ylabel={Reconstruction scale $s_{\mathrm{rec}}^{E}$}]
\jointfidplot{\jointDINOTwoFour}{(2.25,0.75)}

\nextgroupplot[title={$4\times2$}, xlabel={CFG scale}]
\jointfidplot{\jointDINOFourTwo}{(2.25,0.75)}

\nextgroupplot[
    title={$4\times4$},
    ylabel={Reconstruction scale $s_{\mathrm{rec}}^{E}$},
    xlabel={CFG scale},
    colorbar horizontal,
    colorbar style={
        at={(1.45,0.60)}, anchor=west,
        width=0.30\textwidth,
        height=0.25cm,
        xlabel={FID $\downarrow$},
        xtick={1.05,1.3,1.55,1.8},
        tick label style={font=\footnotesize},
        label style={font=\small},
    },
]
\jointfidplot{\jointDINOFourFour}{(2.50,1.00)}
\end{groupplot}
\end{tikzpicture}
\caption{Joint CFG and encoder-reconstruction guidance search. Squares show evaluated 100-step configurations; the black outline marks the lowest FID in each panel.}
\label{fig:joint-dino-guidance-fid}
\end{figure}

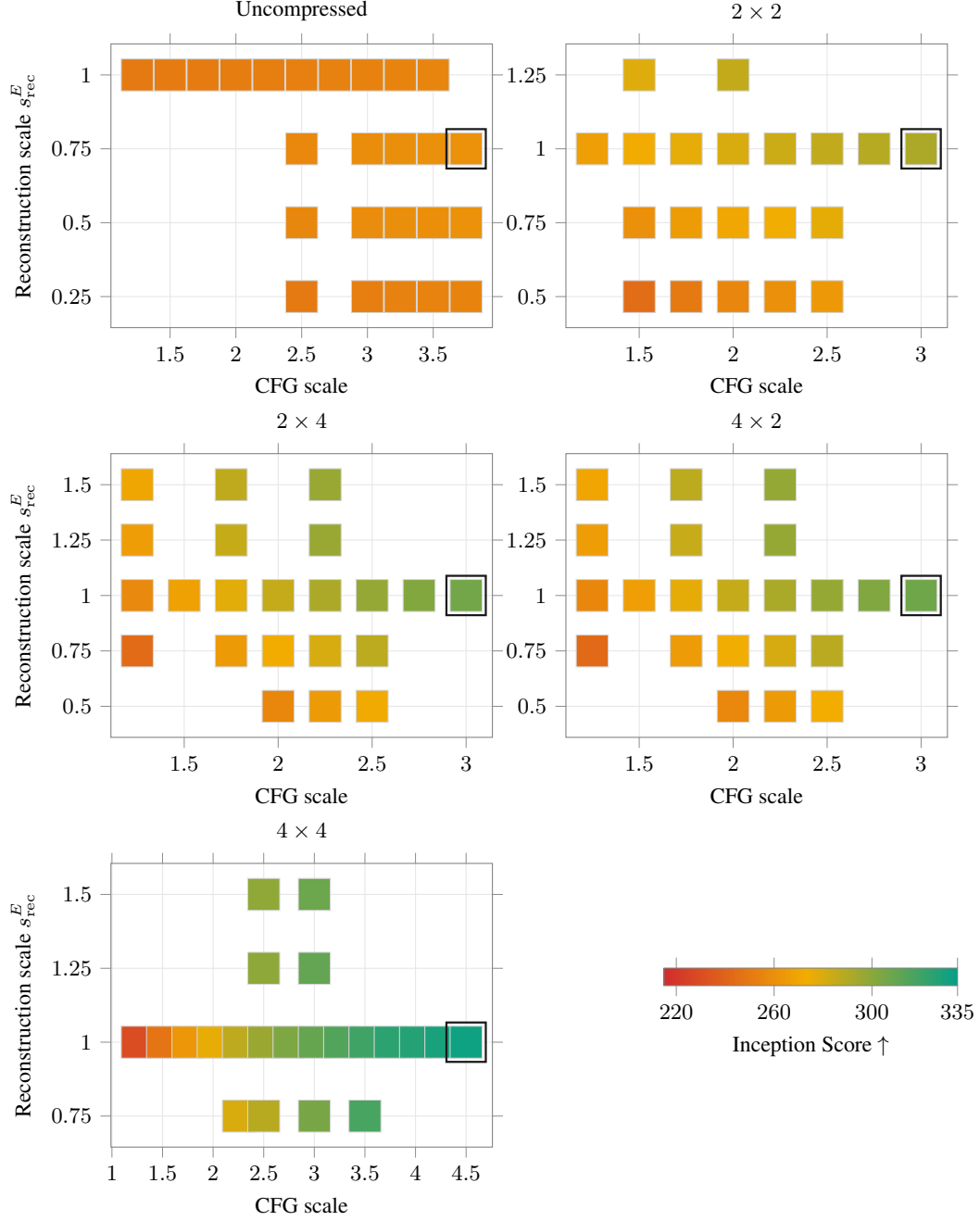
\begin{figure}[p]
\centering
\begin{tikzpicture}
\begin{groupplot}[
    joint guidance axis,
    xlabel={CFG scale},
    point meta min=215, point meta max=335,
    colormap name=pool is,
    group style={group size=2 by 3, horizontal sep=1.05cm, vertical sep=1.80cm},
]
\nextgroupplot[title={Uncompressed}, ylabel={Reconstruction scale $s_{\mathrm{rec}}^{E}$}]
\jointisplot{\jointDINOUncompressed}{(3.75,0.75)}

\nextgroupplot[title={$2\times2$}]
\jointisplot{\jointDINOTwoTwo}{(3.00,1.00)}

\nextgroupplot[title={$2\times4$}, ylabel={Reconstruction scale $s_{\mathrm{rec}}^{E}$}]
\jointisplot{\jointDINOTwoFour}{(3.00,1.00)}

\nextgroupplot[title={$4\times2$}, xlabel={CFG scale}]
\jointisplot{\jointDINOFourTwo}{(3.00,1.00)}

\nextgroupplot[
    title={$4\times4$},
    ylabel={Reconstruction scale $s_{\mathrm{rec}}^{E}$},
    xlabel={CFG scale},
    colorbar horizontal,
    colorbar style={
        at={(1.45,0.60)}, anchor=west,
        width=0.30\textwidth,
        height=0.25cm,
        xlabel={Inception Score $\uparrow$},
        xtick={220,260,300,335},
        tick label style={font=\footnotesize},
        label style={font=\small},
    },
]
\jointisplot{\jointDINOFourFour}{(4.50,1.00)}
\end{groupplot}
\end{tikzpicture}
\caption{Inception Score for the joint CFG and encoder-reconstruction guidance search. The black outline marks the highest score in each panel.}
\label{fig:joint-dino-guidance-is}
\end{figure}

\begin{figure}[p]
\centering
\begin{tikzpicture}
\begin{groupplot}[
    joint guidance axis,
    xlabel={CFG scale},
    point meta min=1.05, point meta max=1.80,
    colormap name=pool fid,
    group style={group size=2 by 3, horizontal sep=1.10cm, vertical sep=1.80cm},
]
\nextgroupplot[title={Uncompressed}, ylabel={Internal-guidance scale}]
\jointfidplot{\jointIGUncompressed}{(2.50,1.78)}

\nextgroupplot[title={$2\times2$}]
\jointfidplot{\jointIGTwoTwo}{(1.75,1.75)}

\nextgroupplot[title={$2\times4$}, ylabel={Internal-guidance scale}]
\jointfidplot{\jointIGTwoFour}{(1.50,2.00)}

\nextgroupplot[title={$4\times2$}, xlabel={CFG scale}]
\jointfidplot{\jointIGFourTwo}{(1.50,2.00)}

\nextgroupplot[
    title={$4\times4$},
    ylabel={Internal-guidance scale},
    xlabel={CFG scale},
    colorbar horizontal,
    colorbar style={
        at={(1.45,0.60)}, anchor=west,
        width=0.30\textwidth,
        height=0.25cm,
        xlabel={FID $\downarrow$},
        xtick={1.05,1.3,1.55,1.8},
        tick label style={font=\footnotesize},
        label style={font=\small},
    },
]
\jointfidplot{\jointIGFourFour}{(1.75,2.25)}
\end{groupplot}
\end{tikzpicture}
\caption{Joint CFG and internal-guidance search. Squares show evaluated 100-step configurations; the black outline marks the lowest FID in each panel.}
\label{fig:joint-ig-fid}
\end{figure}
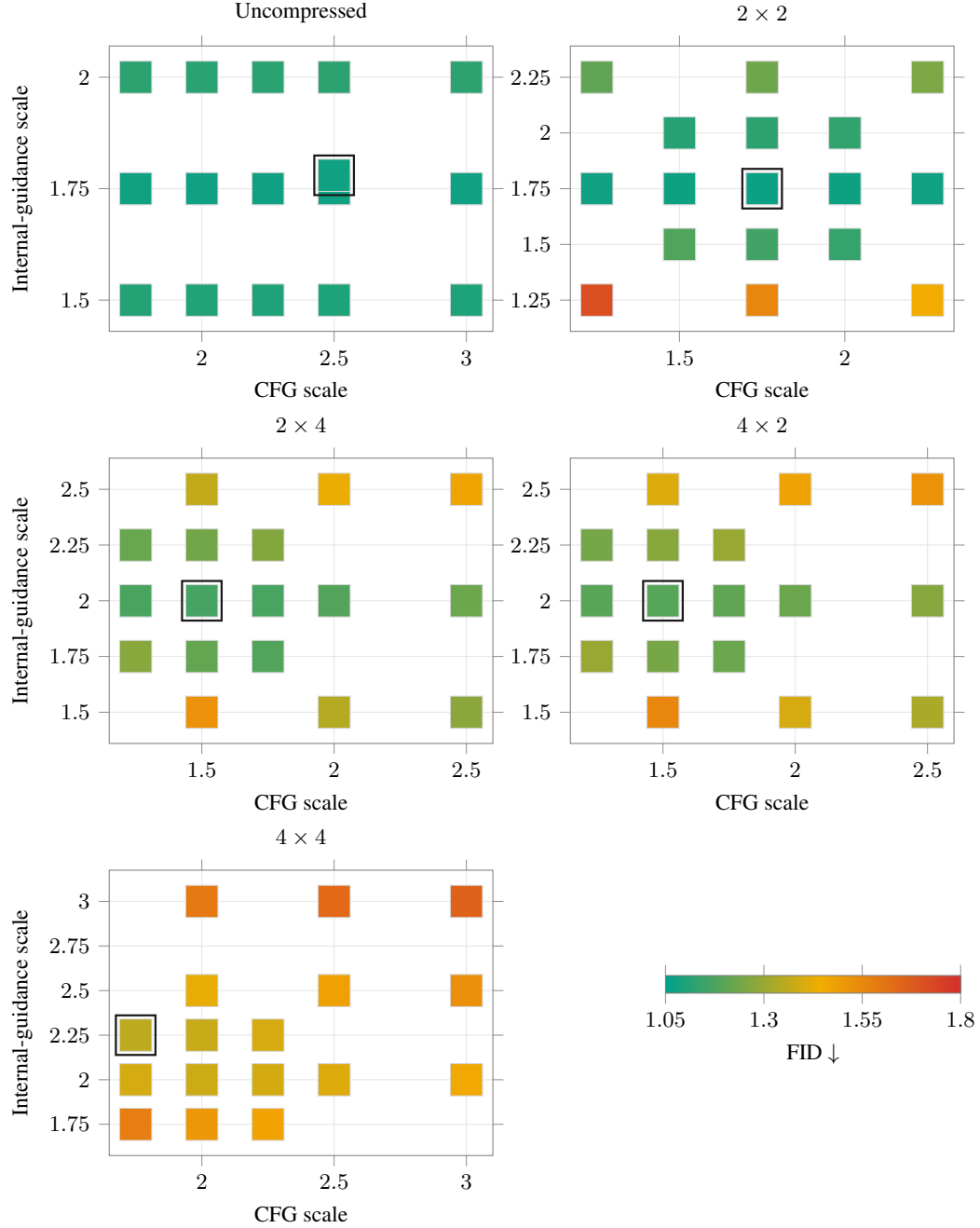

\begin{figure}[p]
\centering
\begin{tikzpicture}
\begin{groupplot}[
    joint guidance axis,
    xlabel={CFG scale},
    point meta min=215, point meta max=335,
    colormap name=pool is,
    group style={group size=2 by 3, horizontal sep=1.10cm, vertical sep=1.80cm},
]
\nextgroupplot[title={Uncompressed}, ylabel={Internal-guidance scale}]
\jointisplot{\jointIGUncompressed}{(3.00,2.00)}

\nextgroupplot[title={$2\times2$}]
\jointisplot{\jointIGTwoTwo}{(2.25,2.25)}

\nextgroupplot[title={$2\times4$}, ylabel={Internal-guidance scale}]
\jointisplot{\jointIGTwoFour}{(2.50,2.50)}

\nextgroupplot[title={$4\times2$}, xlabel={CFG scale}]
\jointisplot{\jointIGFourTwo}{(2.50,2.50)}

\nextgroupplot[
    title={$4\times4$},
    ylabel={Internal-guidance scale},
    xlabel={CFG scale},
    colorbar horizontal,
    colorbar style={
        at={(1.45,0.60)}, anchor=west,
        width=0.30\textwidth,
        height=0.25cm,
        xlabel={Inception Score $\uparrow$},
        xtick={220,260,300,335},
        tick label style={font=\footnotesize},
        label style={font=\small},
    },
]
\jointisplot{\jointIGFourFour}{(3.00,3.00)}
\end{groupplot}
\end{tikzpicture}
\caption{Inception Score for the joint CFG and internal-guidance search. The black outline marks the highest score in each panel.}
\label{fig:joint-ig-is}
\end{figure}
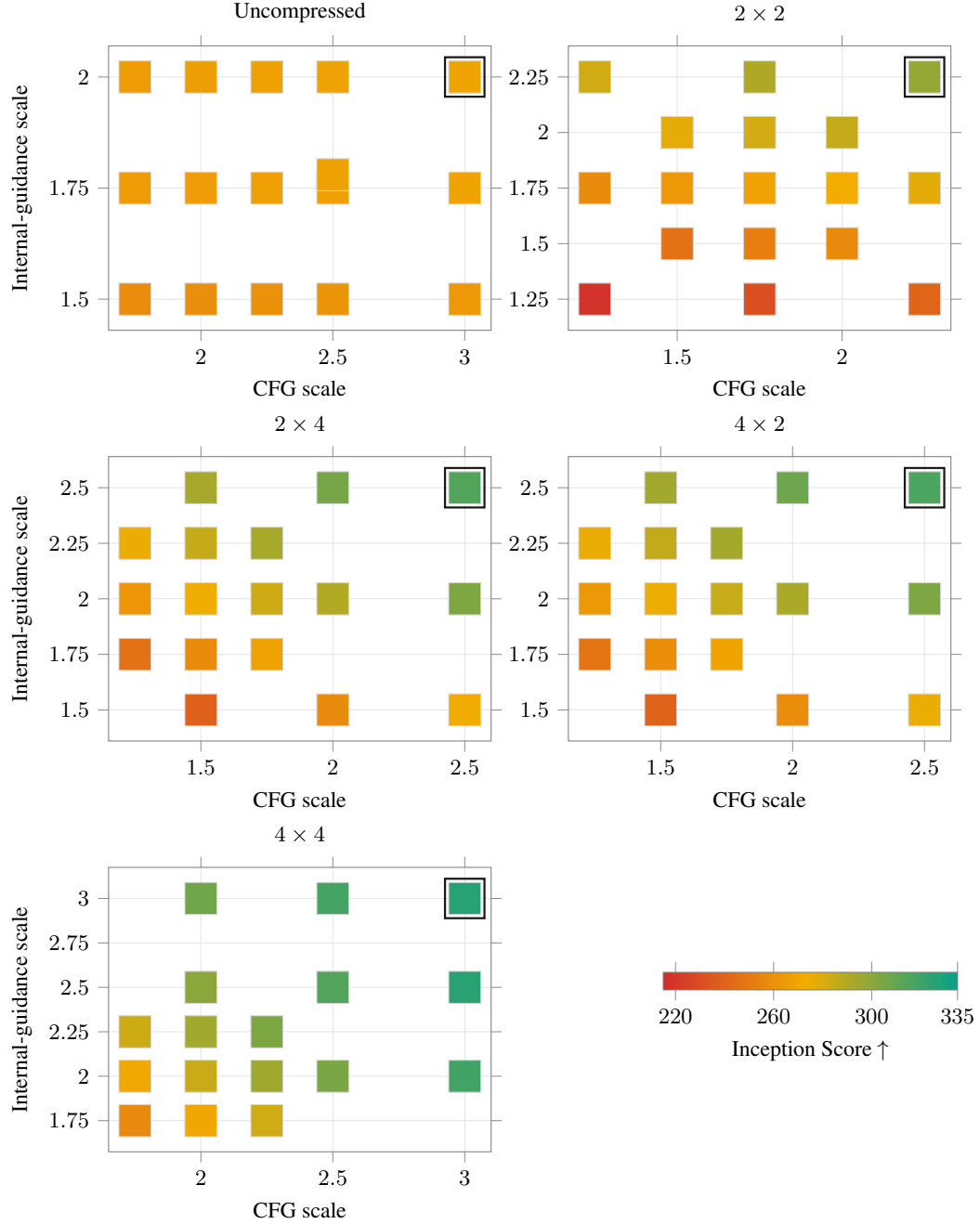
\endgroup

%% file: figures/generated_samples.tex
\clearpage
\section{Generated samples}
\label{app:generated-samples}

We show 48 randomly selected samples per model in eight-column, six-row grids.
Corresponding positions have the same class condition across all grids. Generation uses
100 Euler steps and IG alone at each model's selected scale.
\begingroup
\newcommand{\poolSampleGrid}[4]{%
\begin{figure}[H]
\centering
\includegraphics[width=\textwidth]{figures/samples/#1.jpg}
\caption{#2, trained for #3 epochs, with $s_{\mathrm{IG}}=#4$ and no CFG. Samples are randomly selected without quality filtering.}
\label{fig:samples-#1}
\end{figure}%
}

\poolSampleGrid{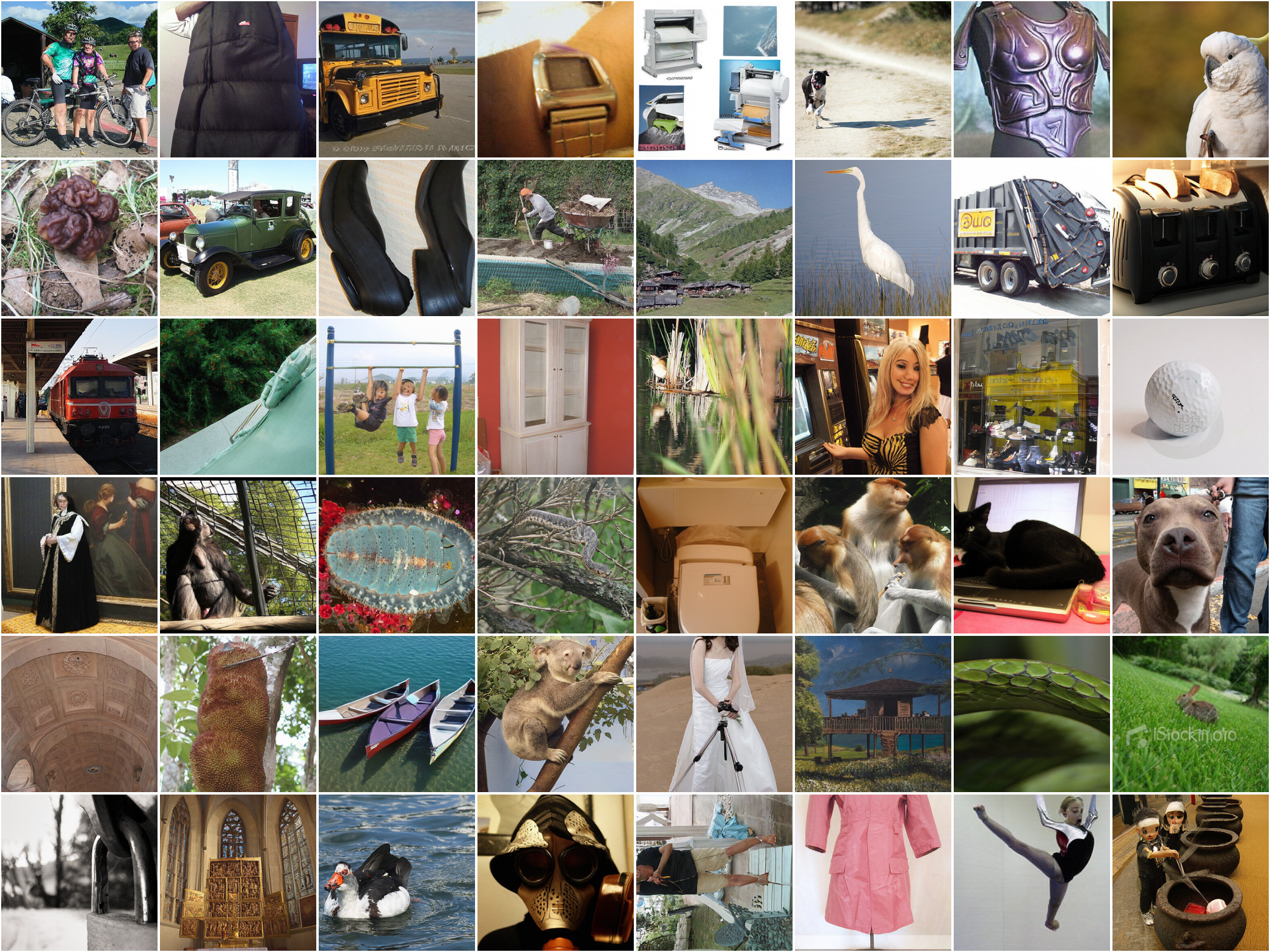}{Unpooled RAEv2 reference ($1\times1$)}{80}{1.75}
\clearpage
\poolSampleGrid{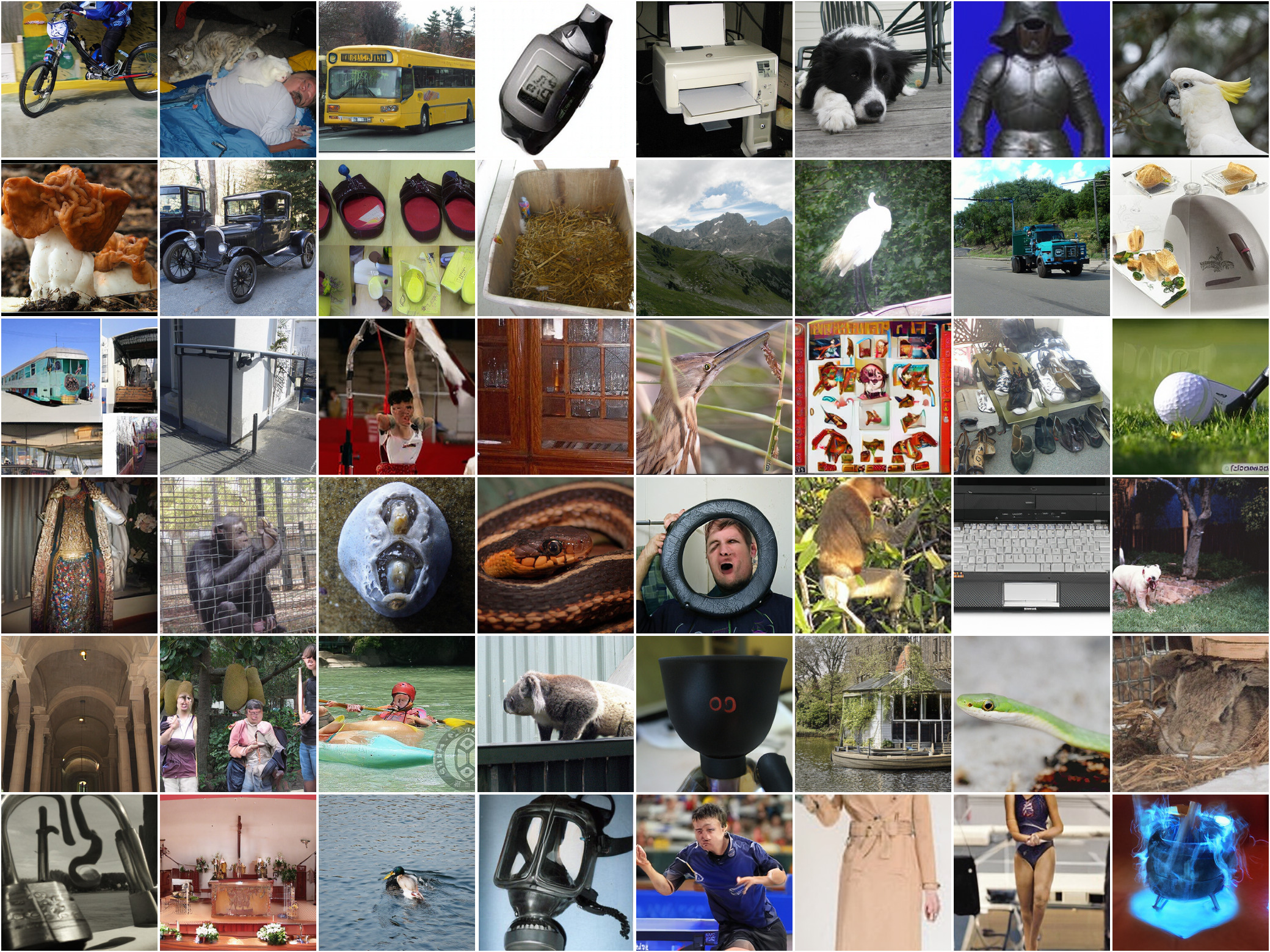}{PoolDINO with $2\times2$ learned pooling ($4\times$ token compression)}{80}{1.75}
\clearpage
\poolSampleGrid{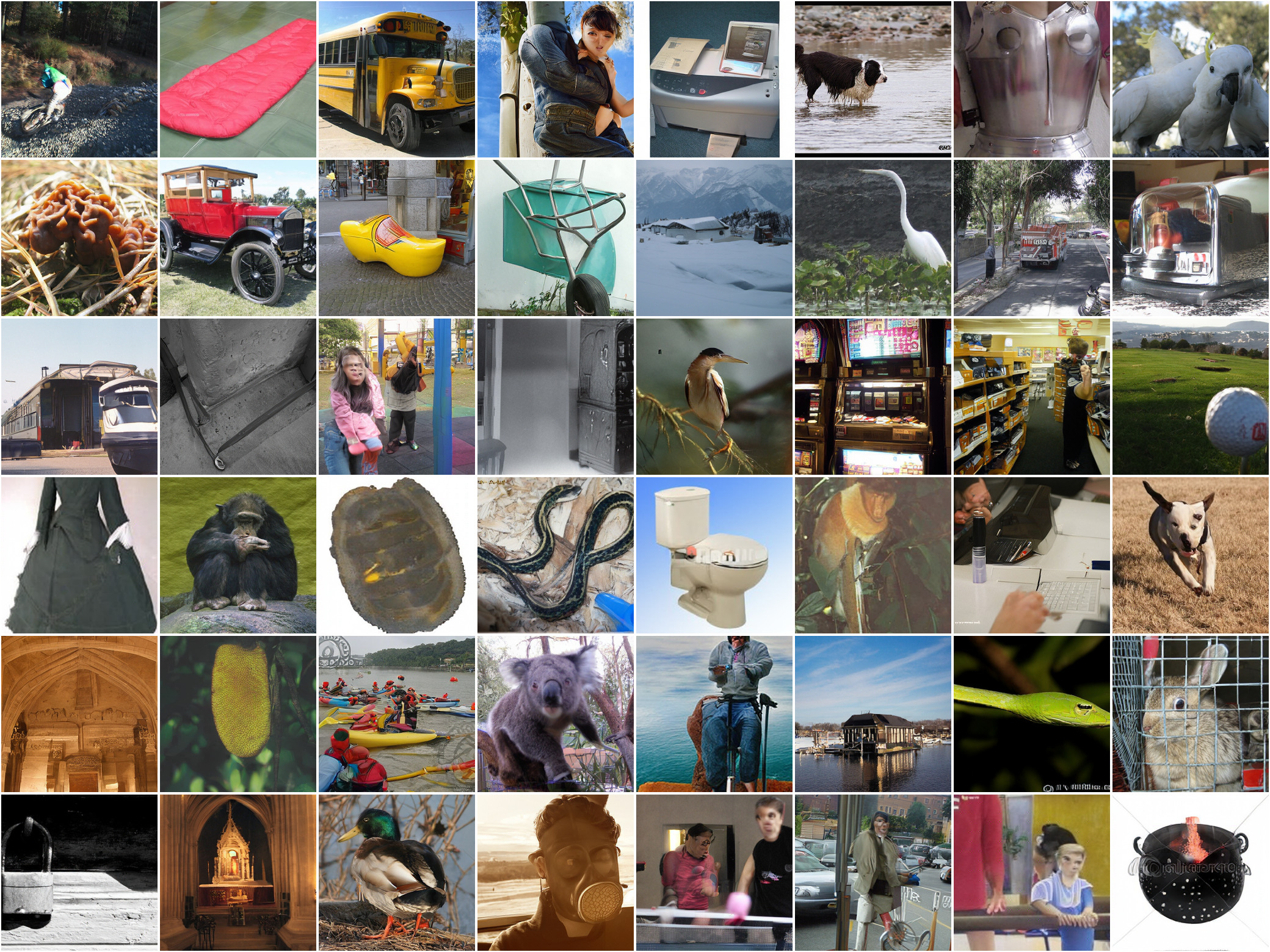}{PoolDINO with $2\times4$ learned pooling ($8\times$ token compression)}{80}{2.00}
\clearpage
\poolSampleGrid{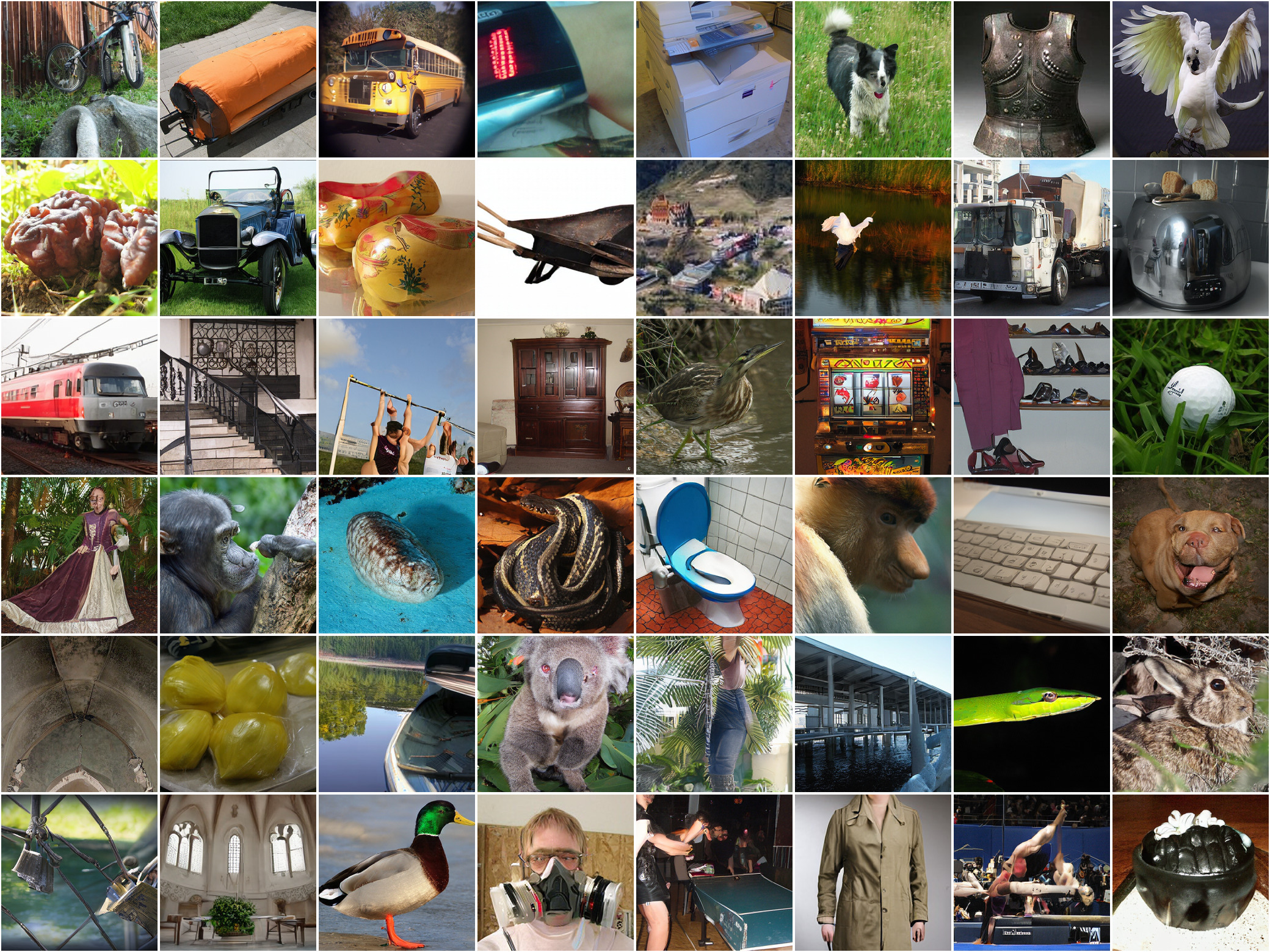}{PoolDINO with $4\times4$ learned pooling ($16\times$ token compression)}{80}{2.75}
\clearpage
\poolSampleGrid{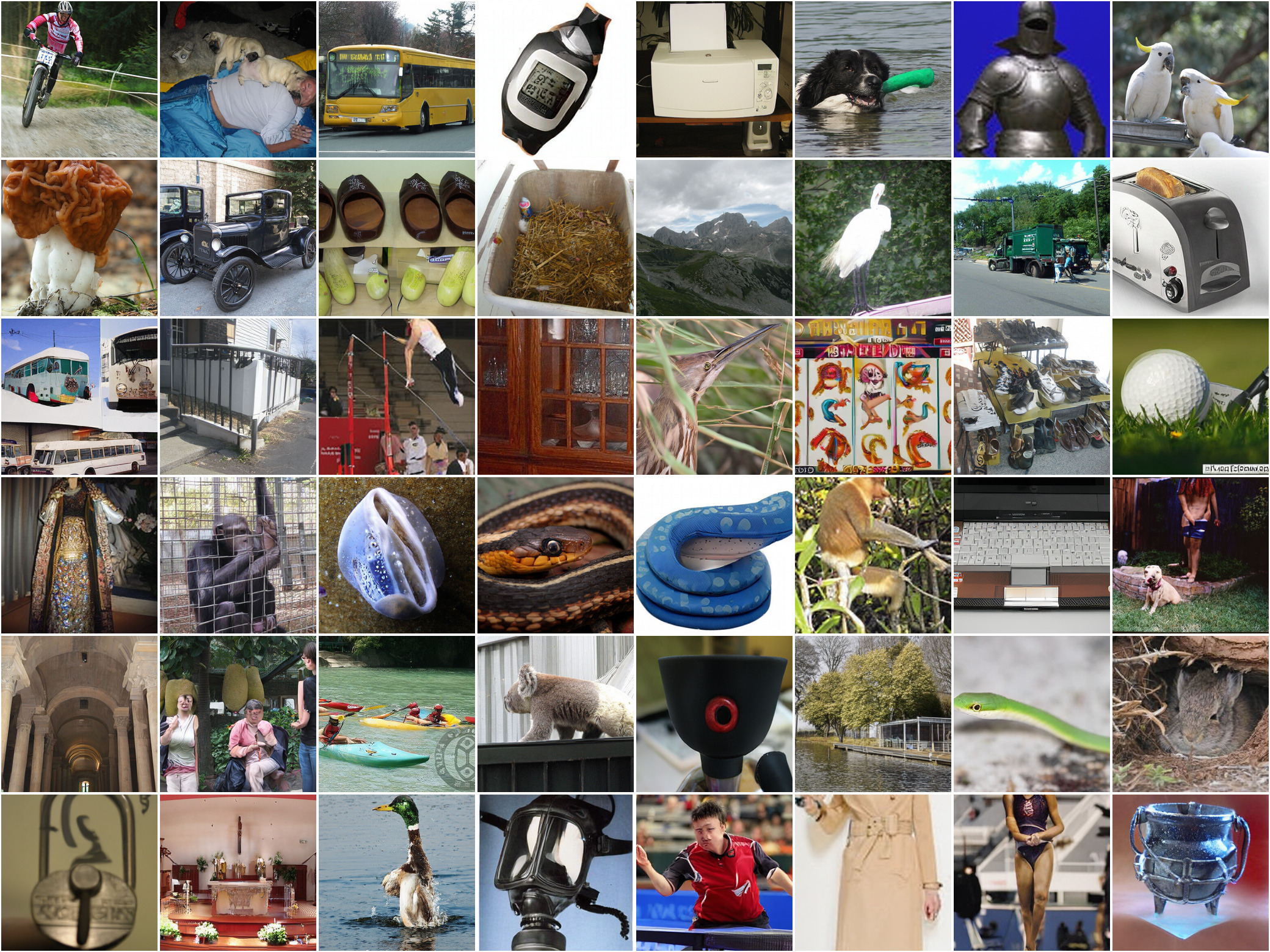}{PoolDINO with $2\times2$ learned pooling and extended training ($4\times$ token compression)}{180}{2.00}
\clearpage
\poolSampleGrid{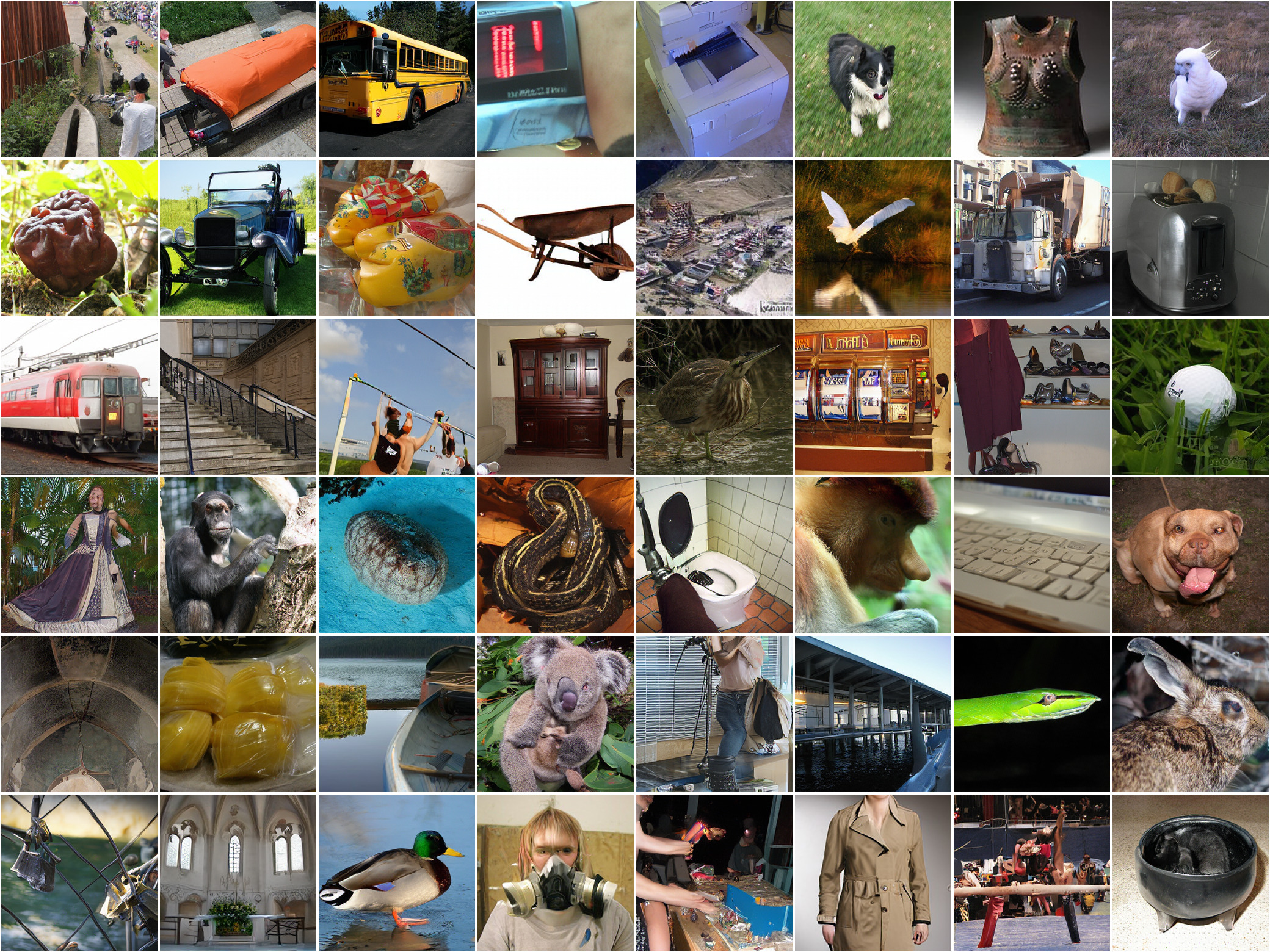}{PoolDINO with $4\times4$ learned pooling and extended training ($16\times$ token compression)}{300}{2.75}
\endgroup